\documentclass[3p,12pt]{elsarticle}

\usepackage{amsmath,amssymb}
\usepackage{latexsym}
\usepackage{bbm}
\usepackage{subfig}
\usepackage{multirow}
\usepackage{rotating}
\usepackage[section]{placeins} 

\usepackage{url}

\usepackage{color}
\usepackage{ulem}
\usepackage{algorithmic}
\usepackage{algorithm}

\journal{Information Sciences}

\newcommand{\reels}{\mathbb{R}}

\renewcommand{\Pr}{\mathbb{P}}

\newcommand{\calO}{{\cal O}}

\newcommand{\calF}{{\cal F}}
\newcommand{\calD}{{\cal D}}

\def\pih{\widehat{\pi}}

\def\bmu{\boldsymbol{\mu}}

\def\bzero{\boldsymbol{0}}

\def\bdelta{\boldsymbol{\delta}}
\def\btheta{\boldsymbol{\theta}}
\def\bSigma{\boldsymbol{\Sigma}}

\newcommand{\bthetah}{\widehat{\btheta}}

\def\bx{{\boldsymbol{x}}}

\def\bm{{\boldsymbol{m}}}
\def\bu{{\boldsymbol{u}}}
\def\bQ{{\boldsymbol{Q}}}

\def\bX{{\boldsymbol{X}}}

\def\bD{{\boldsymbol{D}}}
\def\bC{{\boldsymbol{C}}}

\def\bS{{\boldsymbol{S}}}
\def\bB{{\boldsymbol{B}}}
\def\bA{{\boldsymbol{A}}}

\def\bone{{\boldsymbol{1}}}

\newcommand{\bi}{\begin{itemize}}
\newcommand{\ei}{\end{itemize}}
\newcommand{\be}{\begin{enumerate}}
\newcommand{\ee}{\end{enumerate}}
\newcommand{\bd}{\begin{description}}
\newcommand{\ed}{\end{description}}

\newtheorem{Ex}{Example}

\begin{document}
\begin{frontmatter}

\title{Calibrated model-based evidential clustering using bootstrapping}

\author[utc,shu]{Thierry Den{\oe}ux}
\ead{Thierry.Denoeux@utc.fr}

\address[utc]{Universit\'e de Technologie de Compi\`egne, CNRS\\
UMR 7253 Heudiasyc, Compi\`egne, France}
\address[shu]{Shanghai University, UTSEUS, Shanghai, China}

\begin{abstract}
Evidential clustering is an approach to clustering in which  cluster-membership uncertainty is represented by a collection of  Dempster-Shafer mass functions forming an evidential partition. In this paper, we propose to construct these mass functions by bootstrapping finite mixture models. In the first step, we compute bootstrap percentile confidence intervals for all pairwise probabilities  (the probabilities for any two objects to belong to the same class). We then construct an evidential partition such that the pairwise belief and plausibility degrees approximate the bounds of the confidence intervals. This evidential partition is  calibrated, in the sense that the pairwise belief-plausibility intervals contain the true probabilities ``most of the time'', i.e., with a probability close to the defined confidence level. This frequentist  property is verified by simulation, and  the practical applicability of the method is demonstrated using several real datasets.  
\end{abstract}

\begin{keyword}
Belief functions; Dempster-Shafer theory; evidence theory;  resampling; unsupervised learning; mixture models.
\end{keyword}

\end{frontmatter}

\section{Introduction}

Although the first clustering algorithms  were developed more than 50 years ago (see, e.g., \cite{jain88} and references therein), cluster analysis is still a very active research topic today. One of the remaining open problems concerns the description and quantification of \emph{cluster-membership uncertainty} \cite{durso17,peters13}. Whereas classical partitional clustering algorithms such as the $c$-means procedure are fully deterministic, many of the clustering algorithms used nowadays are based on ideas from fuzzy sets \cite{bezdek81, bezdek99,durso19}, possibility theory \cite{krishnapuram93,ferone19}, rough sets \cite{lingras12,peters14} and probability theory  \cite{celeux95,mclachlan00} to represent cluster-membership uncertainty.  Recently, \emph{evidential clustering} was introduced as a very general approach to clustering that uses the Dempster-Shafer (DS) theory of belief functions \cite{dempster67a,shafer76,denoeux20b} as a model of uncertainty. At the core of the evidential clustering approach is  the notion of \emph{evidential partition} \cite{denoeux04b,masson08}. Basically, an evidential partition is a vector of $n$ mass functions $m_1,\ldots,m_n$, where $n$ is the number of objects, and $m_i$ is a DS mass function representing the uncertainty in the class-membership of object $i$ \citep{denoeux04b}. Fuzzy, probabilistic, possibilistic and rough clustering are recovered as special cases corresponding to restricted forms of the mass functions \cite{denoeux16b}. Evidential clustering has been successfully applied in various domains such as machine prognosis \cite{serir12}, medical image processing \cite{makni14,lelandais14b,lian18} and analysis of social networks \cite{zhou15}.

Different evidential clustering algorithms have been proposed to build an evidential partition of a given attribute or proximity dataset \cite{denoeux04b,masson08,denoeux16a}. The EVCLUS algorithm introduced in \cite{denoeux04b} and improved in \cite{denoeux16a} consists in searching for an evidential partition such that the degrees of conflict between pairs of mass functions $(m_i,m_j)$ match the dissimilarities $d_{ij}$ between object pairs $(i,j)$, up to an affine transformation. The Evidential $c$-Means (ECM) algorithm \cite{masson08} is an alternate optimization procedure in the  hard, fuzzy and possibilistic $c$-means family, with the difference that not only clusters, but also sets of clusters are represented by prototypes. A relational version applicable to dissimilarity data was also proposed in \cite{masson09a}.

Evidential partitions generated by EVCLUS or ECM have been shown to be more informative than hard or fuzzy partitions. In particular, they make it possible to identify objects located in a overlapping region between two or more clusters as well as outliers, and they can easily be summarized as fuzzy or rough partitions \cite{denoeux04b,masson08}. However, they are purely descriptive and unsuitable for statistical inference. In particular, if  datasets are drawn repeatedly from some probability distribution, there is no guarantee that any statements derived from the evidential partitions will be true most of time.

In this paper, we propose a new method for building an evidential partition with a well-defined \emph{frequency-calibration} property \cite{denoeux06b,denoeux18c}, which can be informally described as follows. Assume that the $n$ objects are drawn at random from some population partitioned in $c$ classes, and each object $i$ is described by an attribute vector $\bx_i$. Given any pair of mass functions $(m_i,m_j)$ representing uncertain information about two objects $i$ and $j$, we can compute a degree of belief $Bel_{ij}$ and a degree of plausibility $Pl_{ij}$ that objects $i$ and $j$ belong to the same class \cite{denoeux17a,li18}. Now, let $P_{ij}$ denote the \emph{true unknown probability} that objects $i$ and $j$ belong to the same class, given attribute vectors $\bx_i$ and $\bx_j$. We will say that an evidential partition $m_1,\ldots,m_n$ is \emph{calibrated} if, for each pair of objects $i$ and $j$, the belief-plausibility interval $[Bel_{ij},Pl_{ij}]$ is a confidence interval for the true probability $P_{ij}$, with some predefined confidence level $1-\alpha$. As a consequence, the  intervals $[Bel_{ij},Pl_{ij}]$ will contain the true probability $P_{ij}$ for a proportion at least $1-\alpha$ of object pairs $(i,j)$, on average.

Our approach to generate calibrated evidential partitions is based on \emph{bootstrapping mixture models}. Model-based clustering is a flexible approach to clustering that assumes the data to be drawn from a mixture of probability distributions \cite{banfield93,celeux95,mclachlan00}. In the case of data with continuous attributes, we typically assume a Gaussian Mixture Model (GMM), in which each of the $c$ clusters corresponds to a multivariate normal distribution \cite{yang19}. The model parameters are usually estimated by the Expectation-Maximization (EM) algorithm \cite{dempster77,mclachlan97}. The bootstrap is a resampling technique that consists in sampling $n$ observations from the dataset with replacement \cite{efron93}. By estimating the model parameters from each bootstrap sample, we will be able to compute confidence intervals $[P_{ij}^l,P_{ij}^u]$ for each pairwise probability $P_{ij}$ using the percentile method \cite{efron93}. We will then compute an evidential partition $m_1,\ldots,m_n$ such that the belief-plausibility intervals $[Bel_{ij},Pl_{ij}]$ approximate the confidence intervals $[P_{ij}^l,P_{ij}^u]$.

The rest of this paper is organized as follows. Basic definitions and results regarding evidential clustering are first recalled in Section \ref{sec:evclus}. Our method is then presented in Section \ref{sec:method}, and experimental results are reported in Section \ref{sec:results}. Finally, Section \ref{sec:concl} concludes the paper.

\section{Evidential clustering}
\label{sec:evclus}

We first briefly introduce necessary definitions and results about DS theory in Section \ref{subsec:DS}. The concept of evidential partition is then recalled in Section \ref{subsec:evclus}.

\subsection{Dempster-Shafer theory}
\label{subsec:DS}

Let $\Omega$ be a finite set. A \emph{mass function} on $\Omega$ is a mapping $m$ from the power set of $\Omega$, denoted by $2^\Omega$, to the interval $[0,1]$, such that 
\[
\sum_{A\subseteq \Omega} m(A)=1.
\]
Every subset $A$ of $\Omega$ such that $m(A)>0$ is called a \emph{focal set} of $m$. When the empty set $\emptyset$ is not a focal set, $m$ is said to be \emph{normalized}. All mass functions will be assumed to be normalized in this paper. When all focal sets are singletons, $m$ is said to be \emph{Bayesian}; it is then equivalent to a probability mass functions. A mass function with only one focal set is said to be \emph{logical}; when this focal set is a singleton, it is said to be \emph{certain}. In DS theory, $\Omega$ represents the domain of an uncertain variable $Y$, and $m$ represents evidence about $Y$. The mass $m(A)$ is then the degree with which the evidence supports  exactly  $A$ without supporting any strict subset of $A$ \citep{shafer76}. 

The \emph{belief} and \emph{plausibility} functions induced by a normalized mass function $m$ are defined, respectively, as
\[
Bel(A):=\sum_{B\subseteq A} m(B) \quad \textrm{and} \quad Pl(A):=\sum_{B\cap A\neq\emptyset} m(B),
\]
for all $A \subseteq \Omega$. The following equalities hold: $Bel(\emptyset)=Pl(\emptyset)=0$, $Bel(\Omega)=Pl(\Omega)=1$, and $Pl(A)=1-Bel(\overline{A})$ for all $A\subseteq \Omega$, where $\overline{A}$ denotes the complement of $A$. The quantity $Bel(A)$ measures the total support in $A$, while $Pl(A)$ measures the lack of support in the complement of $A$. Clearly, $Bel(A)\le Pl(A)$ for all $A\subseteq \Omega$. The three functions $m$, $Bel$ and $Pl$ are three different representations of the same information, as knowing any of them allows us to recover  the other two \citep{shafer76}.

\subsection{Evidential partitions}
\label{subsec:evclus}

Let $\calO$ be a set of $n$ objects. Each object is assumed to belong to one and only one group in $\Omega=\{\omega_1,\ldots,\omega_c\}$. An \emph{evidential (or credal) partition} \citep{denoeux04b} is  a collection $M=(m_1,\ldots,m_n)$ of $n$ mass functions on $\Omega$, in which $m_i$ represents evidence about the group membership of object $i$. An evidential partition thus represents uncertainty about the clustering of objects in $\calO$. The notion of evidential encompasses several classical clustering structures \citep{denoeux16b}:
\bi
\item When  mass functions $m_i$ are certain, then $M$ is equivalent to a hard partition; this case corresponds to full certainty about the group of each object.
\item When mass functions are Bayesian, then $M$ boils down to a fuzzy partition, where the degree of membership $u_{ik}$ of object $i$ to group $k$ is  $u_{ik}=Bel_i(\{\omega_k\})=Pl_i(\{\omega_k\}) \in[0,1]$.
\item When each mass function $m_i$ is logical with focal set $A_i\subseteq \Omega$, $m_i$ is equivalent to a rough partition \citep{peters15a}. The lower and upper  approximations of cluster $\omega_k$ are then defined, respectively, as the set of objects that \emph{surely} belong to group $\omega_k$, and the set of objects that \emph{possibly} belong to group $\omega_k$; they are formally given by
\begin{equation}
\label{eq:lower_upper}
\omega^l_k:=\{i \in \calO| A_i=\{\omega_k\}\} \quad \textrm{and} \quad \omega^u_k:=\{i \in \calO| \omega_k \in A_i\}.
\end{equation}
We then have $Bel_i(\{\omega_k\})=I[i \in \omega_k^l]$ and $Pl_i(\{\omega_k\})=I[i \in \omega_k^u]$, where $I[\cdot]$ denotes the indicator function. 
\ei


\begin{Ex}
\label{ex:butterfly}
Consider the {\tt Butterfly} data displayed in Figure \ref{fig:butterfly}, consisting in 11 objects described by two attributes. Figure \ref{fig:butterfly_credalpart} shows a normalized evidential partition of these data obtained by ECM, with $c=2$ clusters. (An evidential partition is said to be normalized if it is composed of normalized mass functions). We can see, for instance, that object 9, which is situated in the center of the rightmost cluster $\omega_1$, has a mass function $m_9$ such that $m_9(\{\omega_1\})\approx 1$, while object 6, which is located between clusters $\omega_1$ and $\omega_2$, is assigned a mass function $m_6$ verifying   $m_6(\Omega)\approx 1$ with $\Omega=\{\omega_1,\omega_2\}$.

\begin{figure}
\centering  
\subfloat[\label{fig:butterfly}]{\includegraphics[width=0.5\textwidth]{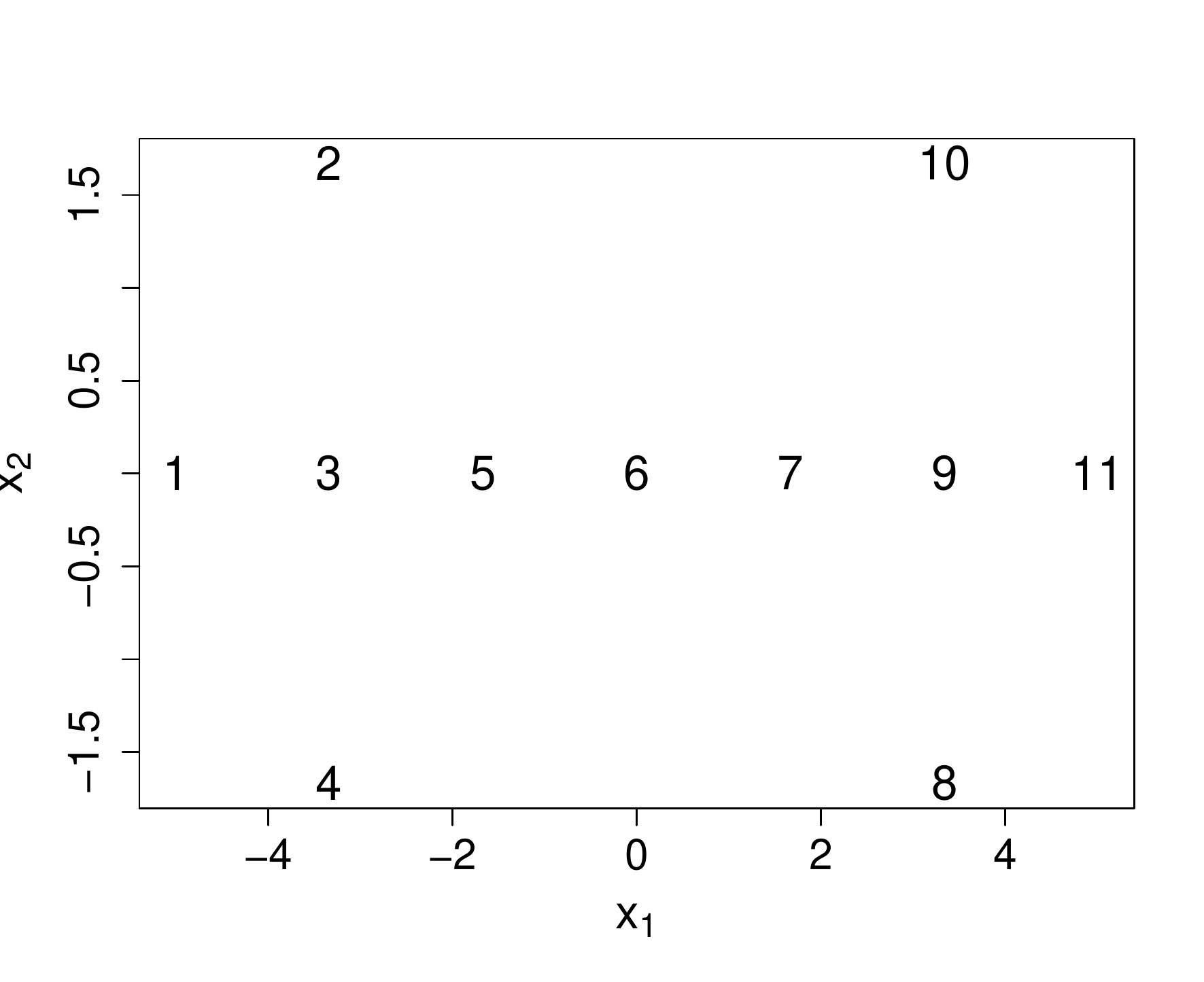}}
\subfloat[\label{fig:butterfly_credalpart}]{\includegraphics[width=0.5\textwidth]{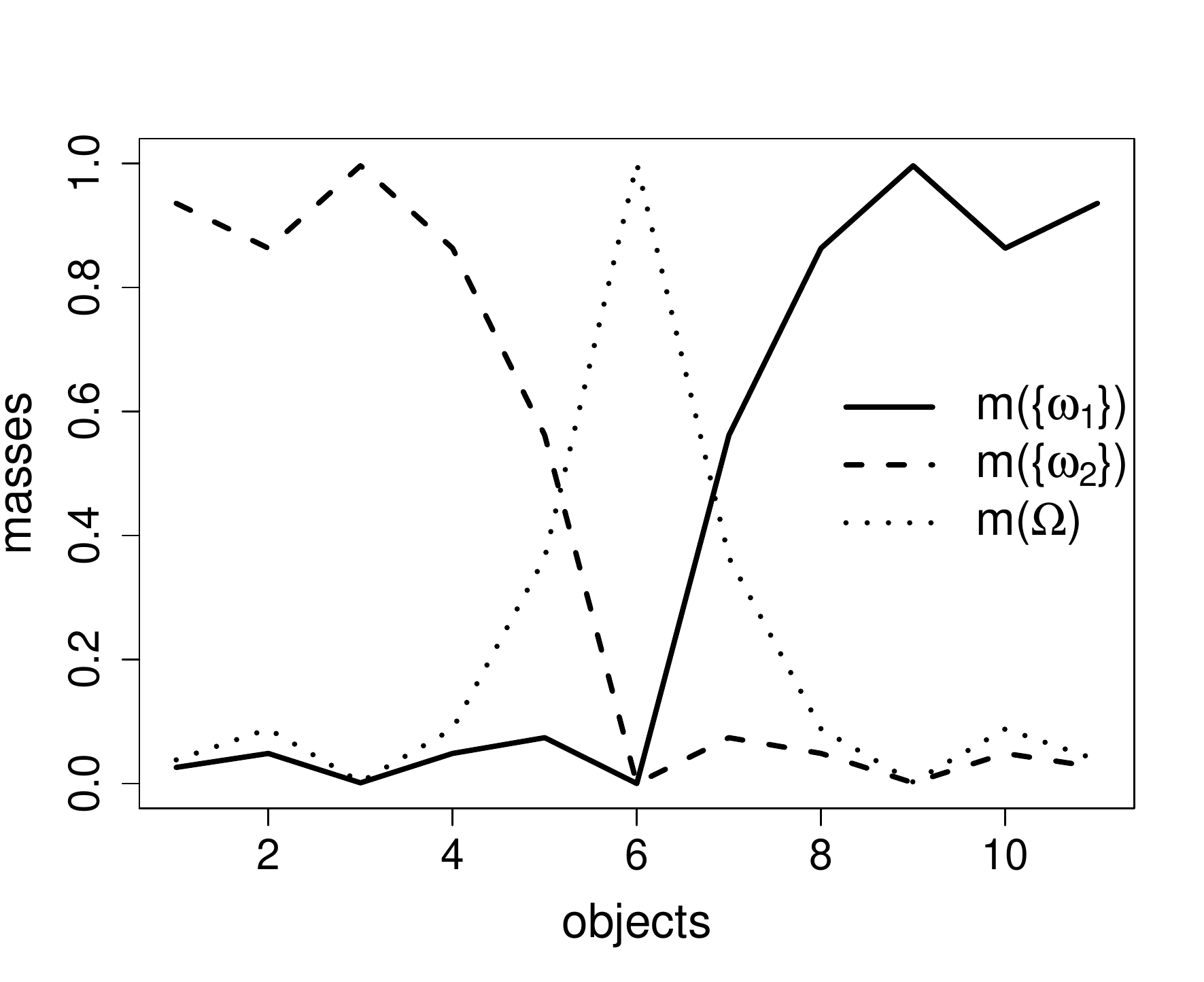}}
\caption{Butterfly dataset (a) and evidential partition with $c=2$ obtained by ECM (b).   \label{fig:ex1}}
\end{figure}
\end{Ex}

Given two distinct objects $i$ and $j$ with corresponding normalized mass functions $m_i$ and $m_j$, we may consider the set $\Theta_{ij}=\{s_{ij},\neg s_{ij}\}$, where $s_{ij}$  denotes the proposition ``Objects $i$ and $j$ belong to the same cluster'' and $\neg s_{ij}$ is the negation of $s$. As shown in \cite{denoeux17a}, the normalized mass function $m_{ij}$ on $\Theta_{ij}$ derived from $m_i$ and $m_j$ has the following expression:
\begin{subequations}
\label{eq:mij}
\begin{align}
m_{ij}(\{s_{ij}\}) & =\sum_{k=1}^c m_i(\{\omega_k\}) m_j(\{\omega_k\}) \\
m_{ij}(\{\neg s_{ij}\}) & =\sum_{A \cap B=\emptyset} m_i(A)m_j(B)\\
m_{ij}(\Theta_{ij}) & =\sum_{A \cap B\neq\emptyset} m_i(A)m_j(B)-\sum_{k=1}^c m_i(\{\omega_k\}) m_j(\{\omega_k\}).
\end{align}
\end{subequations}
Thus, the belief and plausibility that objects $i$ and $j$ belong to the same class are given, respectively, by
\begin{subequations}
\label{eq:BelPlij}
\begin{equation}
Bel_{ij}(\{s_{ij}\})  = m_{ij}(\{s_{ij}\})=\sum_{k=1}^c m_i(\{\omega_k\}) m_j(\{\omega_k\}) 
\end{equation}
and
\begin{equation}
Pl_{ij}(\{s_{ij}\})  =m_{ij}(\{s_{ij}\})+m_{ij}(\Theta_{ij})=\sum_{A \cap B\neq\emptyset} m_i(A)m_j(B).
\end{equation}
\end{subequations}

Given an evidential partition $M=(m_1,\ldots,m_2)$,  the tuple $R=(m_{ij})_{1\le i< j \le n}$ is  called the \emph{relational representation} of  $M$ \citep{denoeux17a}.

\begin{Ex}
Consider objects 4 and 5  the  Example \ref{ex:butterfly} (see Figure \ref{fig:ex1}). We have
\[
m_4(\{\omega_1\})=0.049, \quad m_4(\{\omega_2\})=0.863, \quad m_4(\Omega)=0.088
\]
and
\[
m_5(\{\omega_1\})=0.074, \quad m_5(\{\omega_2\})=0.558, \quad m_5(\Omega)=0.368.
\]
Consequently, we  have
\begin{align*}
m_{45}(\{s_{45}\})&=0.049\times 0.074 + 0.863 \times 0.558 \approx 0.485\\
m_{45}(\{\neg s_{45}\})&=0.049\times 0.558 + 0.863 \times 0.074 \approx 0.0912\\
m_{45}(\Theta_{45})&\approx 1-0.485-0.0912=0.423.
\end{align*}
The degree of belief that objects 4 and 5 belong to the same class is 0.485, and the degree of plausibility is $0.485+0.423=0.908$. Figure~\ref{fig:ex2} displays the complete relational representation of the evidential partition of the {\tt Butterfly} data shown in Figure \ref{fig:butterfly_credalpart}.  The matrices containing $m_{ij}(\{s_{ij}\})$, $m_{ij}(\{\neg s_{ij}\})$ and $m_{ij}(\Theta_{ij})$ are represented graphically in Figures \ref{fig:ms}, \ref{fig:mns} and \ref{fig:mo}, respectively. The pairwise plausibilities $Pl_{ij}(\{s_{ij}\})$ are represented in Figure \ref{fig:pl}. 

\begin{figure}
\centering  
\subfloat[\label{fig:ms}]{\includegraphics[width=0.4\textwidth]{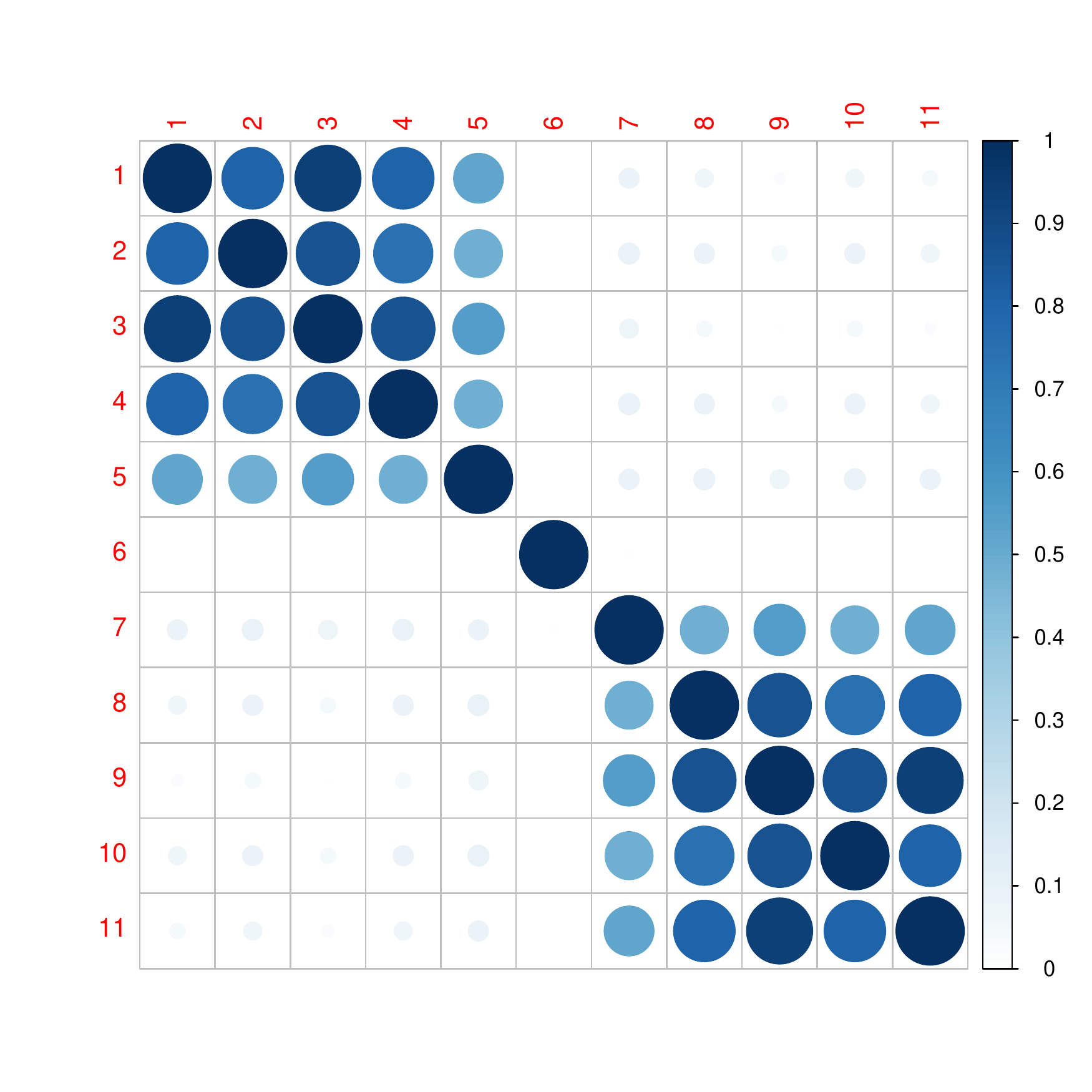}}
\subfloat[\label{fig:mns}]{\includegraphics[width=0.4\textwidth]{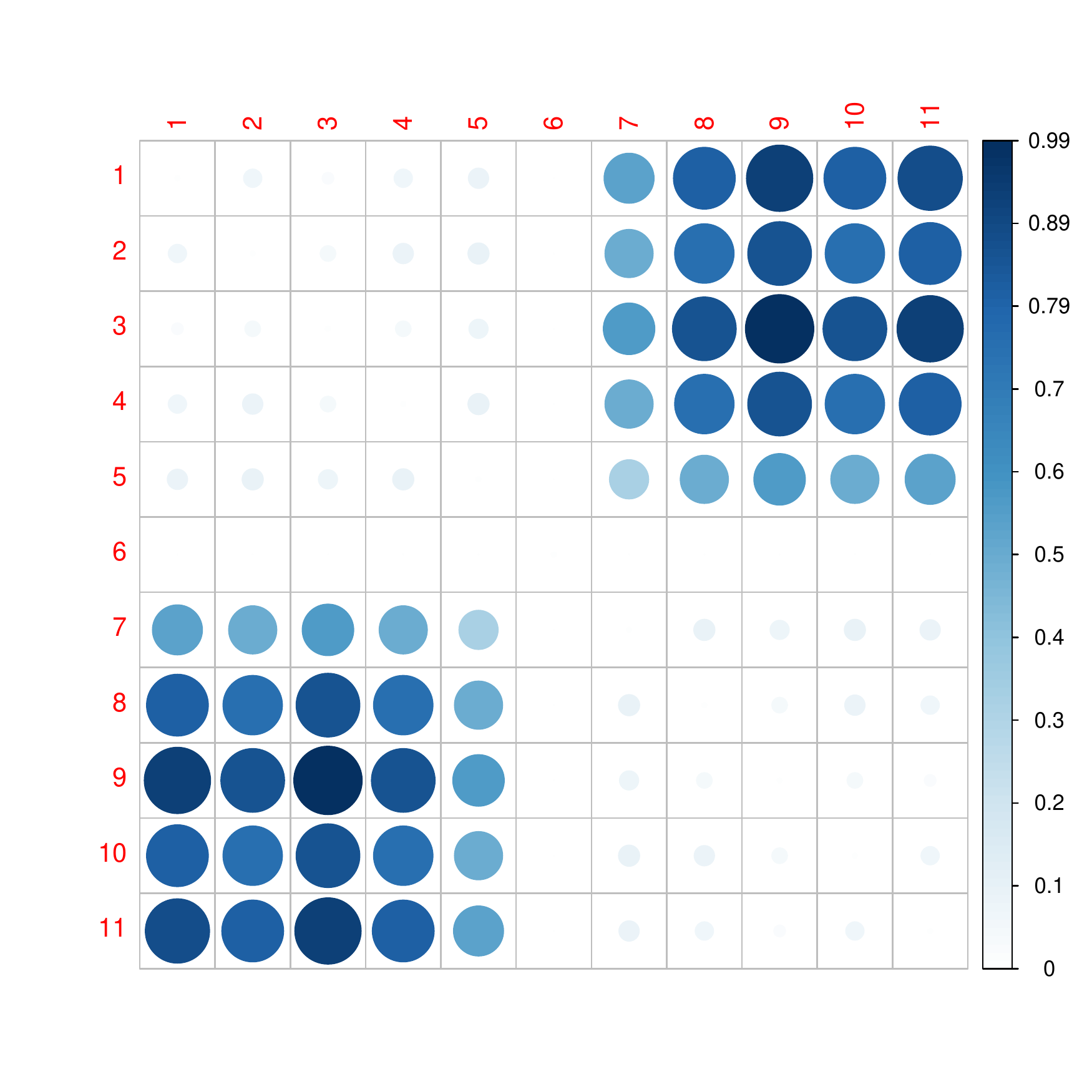}}\\
\subfloat[\label{fig:mo}]{\includegraphics[width=0.4\textwidth]{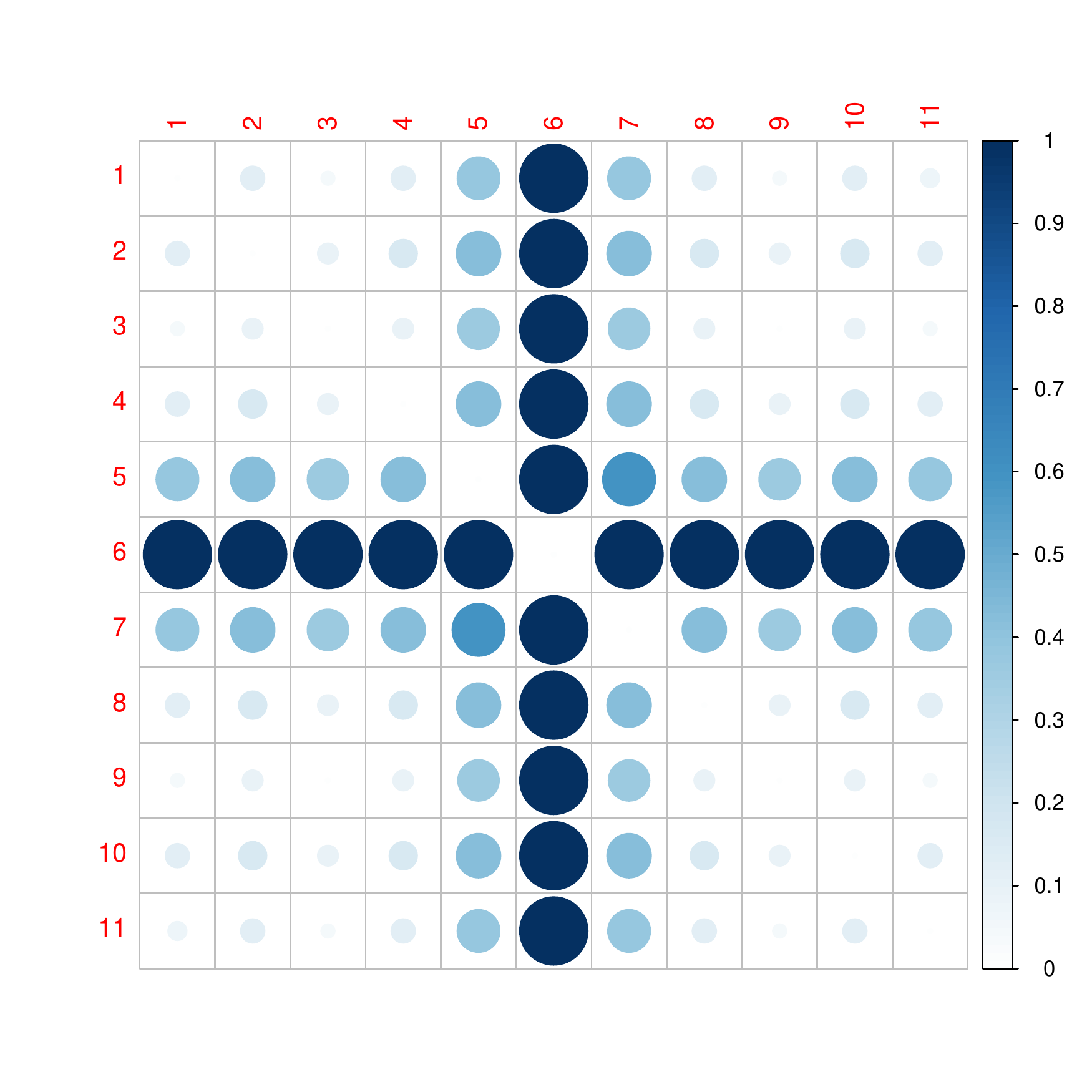}}
\subfloat[\label{fig:pl}]{\includegraphics[width=0.4\textwidth]{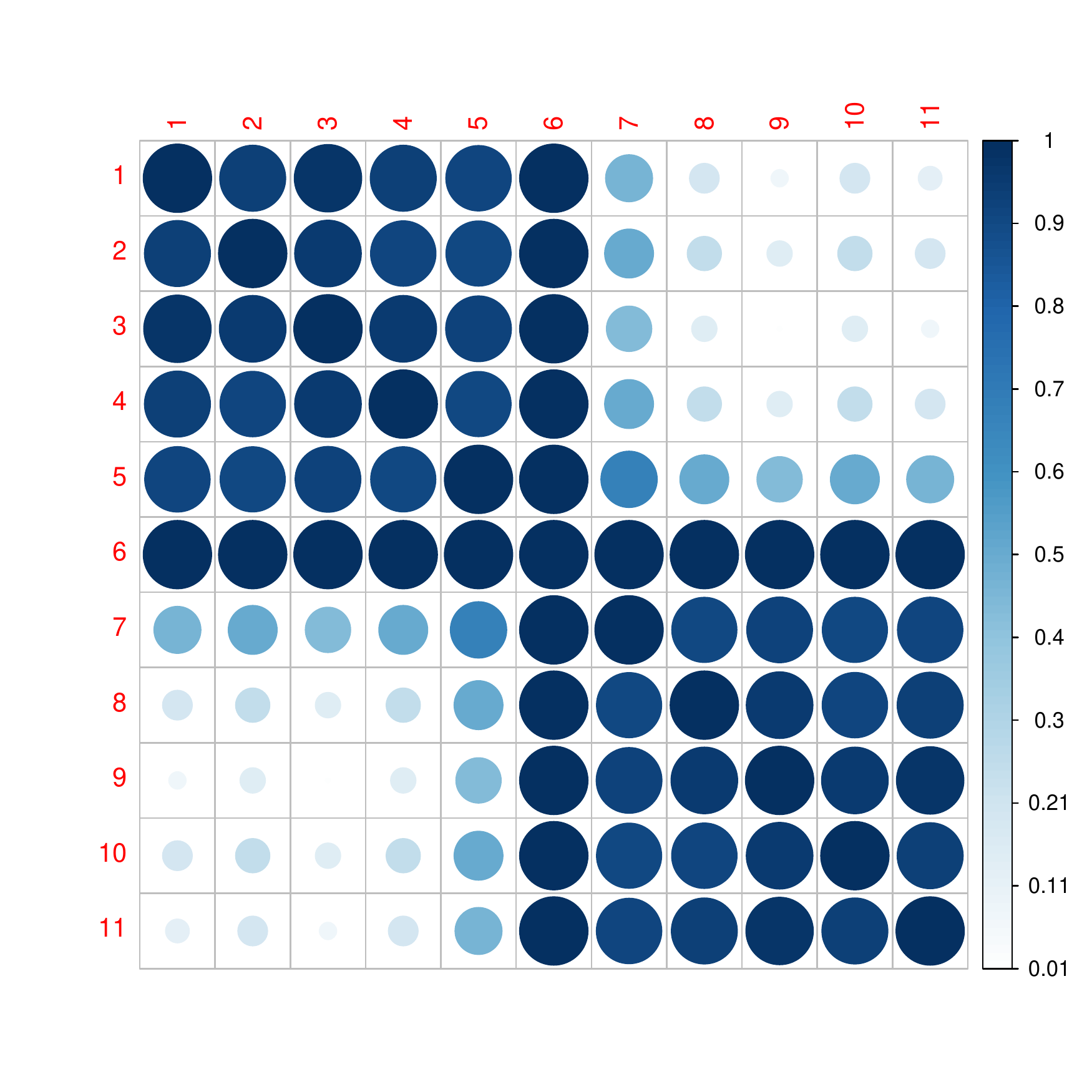}}
\caption{Relational representation of the evidential partition of the {\tt Butterfly} data shown in Figure \ref{fig:butterfly_credalpart}: masses $m_{ij}(\{s_{ij}\})$ (a), $m_{ij}(\{\neg s_{ij}\})$ (b), $m_{ij}(\Theta_{ij})$ (c) and pairwise plausibilities $Pl_{ij}(\{s_{ij}\})$ (d).  \label{fig:ex2}}
\end{figure}
\end{Ex}

\section{Computed calibrated evidential partitions}
\label{sec:method}

In this section, we describe our method for quantifying the uncertainty of model-based clustering using an evidential partition with well-defined properties with respect to the unknown true partition. The assumptions will first be stated in Section \ref{subsec:assumptions}. A method to compute bootstrap confidence intervals on pairwise probabilities $P_{ij}$ will then be described in Section \ref{subsec:IC}. Finally, an algorithm for computing an evidential partition from these confidence intervals will be introduced in Section \ref{subsec:algo}.

\subsection{Assumptions}
\label{subsec:assumptions}

We consider a population of objects, each one described by an attribute vector $\bX \in \reels^d$ and by a class variable $Y\in \Omega=\{1,\ldots,c\}$.  The conditional distribution of $\bX$ given $Y=k$ is described by a probability density function (pdf) $p_k(\bx;\btheta_k)$, where $\btheta_k$ is a vector of parameters. The marginal distribution of $\bX$ is, thus, a mixture distribution with pdf
\[
p(\bx;\btheta)= \sum_{k=1}^c \pi_k p_k(\bx;\btheta_k)
\]
where $\pi_k=\Pr(Y=k)$, $k=1,\ldots,c$ are the prior class densities, and $\btheta=(\btheta_1,\ldots,\btheta_c,\pi_1,\ldots,\pi_c)$ is the vector of all parameters in the model. The conditional probability $\pi_k(\bx;\btheta)$ that $Y=k$ given $\bX=\bx$ can be computed using Bayes' theorem as  
\[
\pi_k(\bx;\btheta)=\frac{p_k(\bx;\btheta_k) \pi_k}{\sum_{\ell=1}^c p_\ell(\bx;\btheta_\ell) \pi_\ell}.
\]

Let $\calD=\{\bx_1,\ldots,\bx_n\}$ be a dataset composed of $n$ attribute vectors describing  $n$ objects. We assume that $\calD$ is a realization of an i.i.d. sample from $\bX$, and we want to quantify the uncertainty about the classes $y_1,\ldots,y_n$ of the $n$ objects. If parameter $\btheta$ was known, then the uncertainty about $y_i$ could be described by the conditional class probabilities $\pi_k(x_i;\btheta)$, $k=1,\ldots,c$, and the probability $P_{ij}(\btheta)$ that  objects $i$ and $j$ belong to the same class could be computed as
\begin{equation}
P_{ij}(\btheta):= \Pr(Y_i=Y_j \mid \bx_i,\bx_j)
=\sum_{k=1}^c \pi_k(\bx_i;\btheta)\, \pi_k(\bx_j;\btheta).
\end{equation}
In usual situations, parameter $\btheta$ is  unknown and it needs to be estimated from the data. Let $\bthetah$ be the maximum likelihood estimate (MLE) of $\btheta$ obtained, e.g., using the EM algorithm \cite{dempster77,mclachlan97}. The estimated conditional class probabilities are $\pih_{ik}:=\pi_k(x_i;\bthetah)$, $k=1,\ldots,c$; they constitute a fuzzy partition of the dataset. The MLE of $P_{ij}(\btheta)$ is $P_{ij}(\bthetah):=\sum_{k=1}^c \pih_{ik}\pih_{jk}$ for all $(i,j)\in \{1,\ldots,n\}^2$. However, these point probability estimates do not adequately reflect group-membership uncertainty, because they do not account for the uncertainty on $\btheta$. In the next section, we propose a method to compute approximate confidence intervals on pairwise probabilities $P_{ij}(\btheta)$.

\subsection{Confidence intervals on pairwise probabilities}
\label{subsec:IC}

Let us consider two \emph{fixed} vectors $\bx_i$ and $\bx_j$ from dataset $\calD$, and an i.i.d. random sample $\bX'_1,\ldots,\bX'_n$ from $p(\bx;\btheta)$. A \emph{confidence interval} on $P_{ij}(\btheta)$ at level $1-\alpha$ is a random interval $\left[P^l_{ij}(\bX'_1,\ldots,\bX'_n),P^u_{ij}(\bX'_1,\ldots,\bX'_n)\right]$ such that, for all $\btheta$,
\[
\Pr\left(P^l_{ij}(\bX'_1,\ldots,\bX'_n) \le P_{ij}(\btheta) \le P^u_{ij}(\bX'_1,\ldots,\bX'_n)\right) \ge 1-\alpha.
\]

Approximate confidence intervals on $P_{ij}(\btheta)$ can be obtained in several different ways. One approach is to use the asymptotic normality of the MLE and to estimate the covariance matrix of $\bthetah$ by the observed information matrix. MacLachlan and Krishnan \cite[Chapter 4]{mclachlan97} review different methods for computing or approximating the observed information matrix, and MacLachlan and Basford \cite[Chapter 2]{mclachlan88} give an approximate analytical expression for the case of a Gaussian mixture. Estimates $\widehat{v}_{ij}$ of the variance $v_{ij}$ of $P_{ij}(\bthetah)$ could then be obtained by the delta method, leading to the following standard confidence interval:
\begin{equation}
\label{eq:CInorm}
P_{ij}(\bthetah) \pm u_{1-\alpha/2} \sqrt{\widehat{v}_{ij}},
\end{equation}
where $u_{1-\alpha/2}$ denotes the $1-\alpha/2$ quantile of the standard normal  distribution. Standard confidence intervals are consistent, but they are based on  asymptotic approximations that can be quite inaccurate in practice \cite{DiCiccio96}. As noted in \cite{ohagan19}, ``in the case of mixture models large sample sizes are required for the asymptotics to give a reasonable approximation''. In our case, the estimates $P_{ij}(\bthetah)$ take values in $[0,1]$, and their distribution can be very asymmetric for small $n$, as will be shown experimentally in Example \ref{ex:ex3} below (Figure \ref{fig:ex3_histos}) and in Section \ref{subsec:simul} (Figure  \ref{fig:confint_histo}). 

Bootstrap confidence intervals can be seen as algorithms for improving  standard intervals such as \eqref{eq:CInorm} without  human intervention \cite{DiCiccio96}.  Given a realization $\bx'_1,\ldots,\bx'_n$ of the random sample, a nonparameteric bootstrap ``pseudo-sample'' is generated by drawing $n$ observations randomly from $\bx'_1,\ldots,\bx'_n$ with replacement. Repeating this operation $B$ times, we obtain $B$  pseudo-samples $\{\bx'_{b1},\ldots,\bx'_{bn}\}_{b=1}^B$ and the corresponding estimates $\bthetah_1,\ldots,\bthetah_B$ of $\btheta$ (computed using the EM algorithm). The simplest technique for computing approximate confidence intervals using this approach is the \emph{bootstrap percentile (BP)} method \citep{efron93}. The BP confidence interval for $P_{ij}(\btheta)$ is defined by the $\alpha/2$ and $1-\alpha/2$ quantiles of $P_{ij}(\bthetah_1),\ldots,P_{ij}(\bthetah_B)$, which will be denoted as $P_{ij}^l$ and $P_{ij}^l$.  Because the original dataset  $\bx_1,\ldots,\bx_n$ was generated from the same distribution as $\bx'_1,\ldots,\bx'_n$, we can use it to compute bootstrap confidence intervals for any pair $(i,j)$ of objects. The procedure is summarized in Algorithm \ref{alg:CI}. For previous applications of the bootstrap approach to model-based clustering, see \cite{ohagan19} and references therein.

Under general conditions stated in \cite[Theorem 4.1]{shao95}, BP confidence intervals are consistent, i.e., we have
\begin{equation}
\label{eq:confint}
\Pr\left(P_{ij}^l \le P_{ij}(\btheta) \le P_{ij}^u\right) \rightarrow 1-\alpha
\end{equation}
as $n\rightarrow \infty$. As shown by Davison and Hinkley \cite[page 213]{davison97}, equi-tailed BP confidence intervals such as \eqref{eq:confint} are superior to standard confidence intervals such as \eqref{eq:CInorm}, in the sense that they are second-order accurate, i.e., we have
\begin{equation}
\label{eq:confint1}
\Pr\left(P_{ij}^l \le P_{ij}(\btheta) \le P_{ij}^u\right)=1-\alpha+ O(n^{-1}),
\end{equation}
whereas the coverage probability of normal approximation confidence intervals is $1-\alpha+ O(n^{-1/2})$. More sophisticated procedures  such as the bootstrap accelerated bias corrected ($BC_a$) method have also been developed to further improve the performance of BP confidence intervals \cite{efron93}, but these methods depend on additional coefficients that are not easy to determine. As confidence intervals on $P_{ij}(\btheta)$ need to be computed for each of the $n(n-1)/2$ pairs of objects, we will stick to the simple BP method. As will be shown in Section \ref{subsec:simul}, the confidence intervals computed by this method have coverage probabilities close to their nominal values, provided the model is correctly specified.

As a final argument in favor of the bootstrap  as compared to the normal approximation method, we can observe that the latter approach relies on the calculation of the information matrix, which can very cumbersome and has to be carried out for each new model. Even if we limit ourselves to  the family of Gaussian mixture models, we usually impose various restrictions on the parameters (as will be shown in Section \ref{subsec:simul}), resulting in different expressions for the information matrix. Furthermore, with some covariance structures,  we  use non-differentiable orthogonal matrices, which prohibits  the information matrix-based approach  \cite{ohagan19}. For non-normal models, the calculations often become intractable. In contrast, the bootstrap method can be  applied without modification to any model. This advantage does come at the cost of heavier computation but, as we will see in Section \ref{sec:results}, the computing time remains manageable on a personal computer with moderate size datasets, for which the method is useful (with large datasets, the second-order uncertainty on membership probabilities can often be neglected anyway).


\begin{algorithm}
\caption{Generation of BP confidence intervals on pairwise probabilities. \label{alg:CI}}
\begin{algorithmic}[1]
\REQUIRE Dataset $\bx_1,\ldots,\bx_n$, model $p(\cdot;\btheta)$, number of bootstrap samples $B$, confidence level $1-\alpha$
\FOR{$b=1$ \TO $B$}
\STATE Draw $\bx_{b1},\ldots,\bx_{bn}$ from $\bx_1,\ldots,\bx_n$ with replacement
\STATE Compute the MLE $\bthetah_b$ from $\bx_{b1},\ldots,\bx_{bn}$ using the EM algorithm
\FOR{all $i<j$}
\STATE Compute $P_{ij}(\bthetah_b)$
\ENDFOR
\ENDFOR
\FOR{all $i<j$}
\STATE $P_{ij}^l := \textsf{Quantile}\left(\left\{P_{ij}(\bthetah_b)\right\}_{b=1}^B;\frac{\alpha}2\right)$
\STATE $P_{ij}^u := \textsf{Quantile}\left(\left\{P_{ij}(\bthetah_b)\right\}_{b=1}^B;1-\frac{\alpha}2\right)$
\ENDFOR
\end{algorithmic}
\end{algorithm}

\begin{Ex}
\label{ex:ex3}
As an example, we consider the dataset shown in Figure \ref{fig:ex3}, consisting of $n=30$ two-dimensional vectors drawn from a mixture of $c=3$ components with the following parameters:
\[
\bmu_1:=(0,1)^T, \quad \bmu_2:=(1,0)^T, \quad \bmu_3:=(1,1)^T,
\]
\[
\bSigma_1=\bSigma_2 =\bSigma_3:=\begin{pmatrix} 0.1 & 0\\ 0 & 0.1 \end{pmatrix}, \quad \pi_1=\pi_2=\pi_3:=1/3.
\]
We applied the above method with $B=1000$, assuming the true model (spherical classes with equal volume). Figure \ref{fig:ex3_histos} shows histograms of the bootstrap estimates  $P_{ij}(\bthetah_b)$ and the bounds of the percentile 90\% confidence interval $P_{ij}^l,P_{ij}^u$ for four pairs of points. We can see that points 11 and 29  have a low probability $P_{11,29}$ of belonging to the same class, and the probability is well estimated with a narrow confidence interval. Point pairs (24,19) and (26,30) have a high  probability of belonging to the same class, and the corresponding confidence interval is also narrow. In contrast, the true probability that points 22 and 23 belong to the same class is approximately equal to 0.7, and the corresponding confidence interval is quite large.

\begin{figure}
\centering
\includegraphics[width=0.6\textwidth]{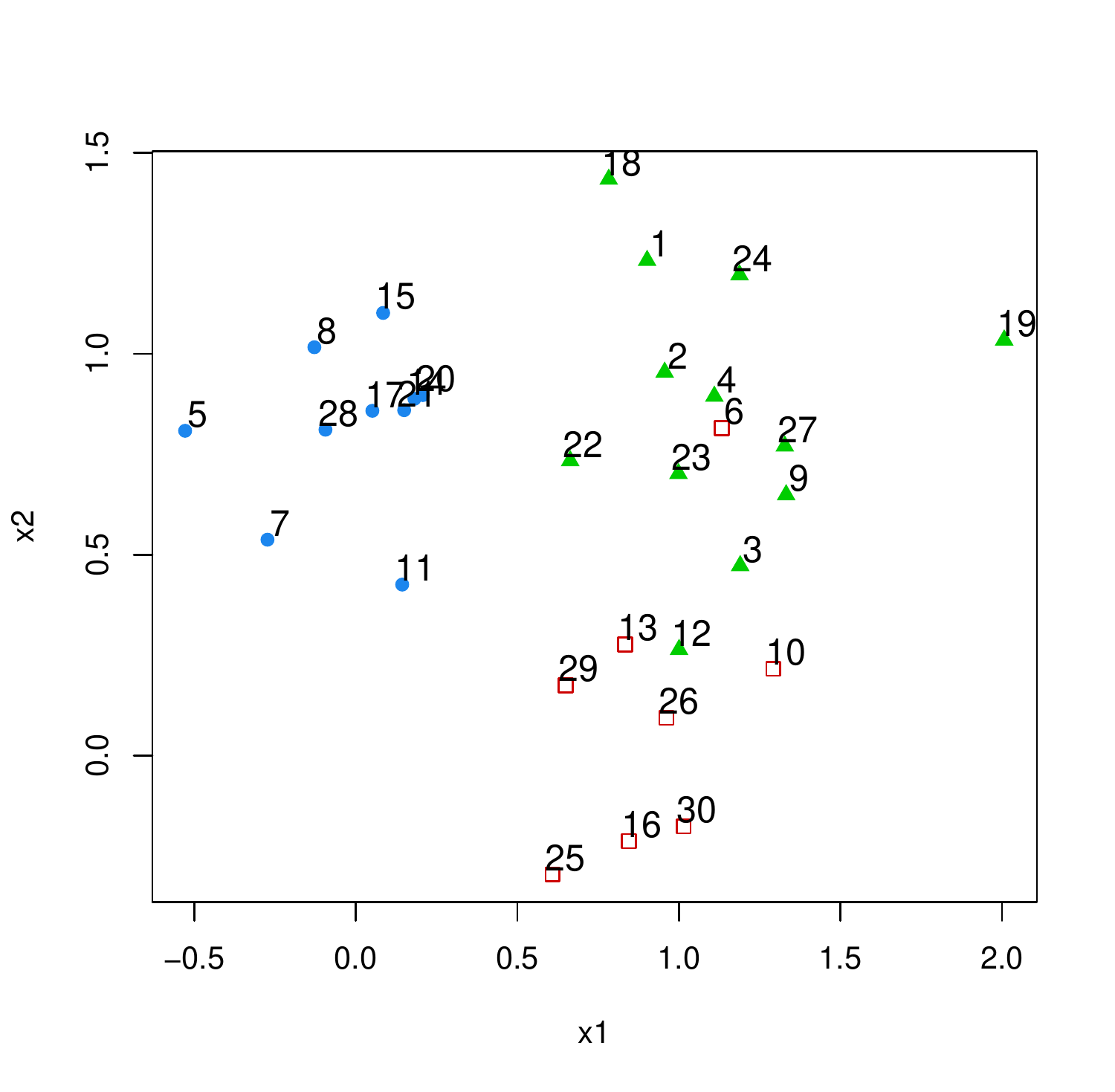}
\caption{Dataset of Example \ref{ex:ex3}. \label{fig:ex3}}
\end{figure}

\begin{figure}
\centering  
\subfloat{\includegraphics[width=0.4\textwidth]{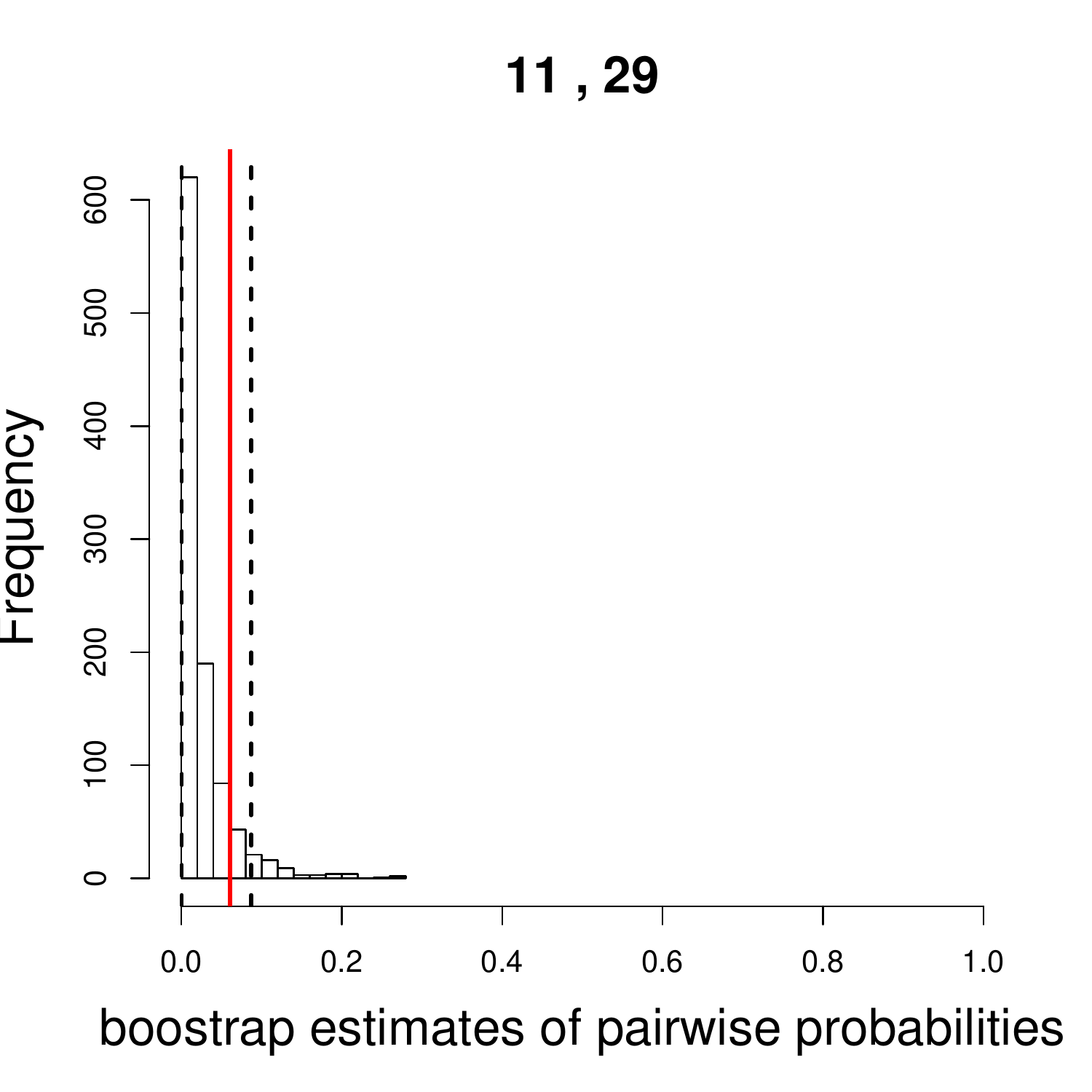}}
\subfloat{\includegraphics[width=0.4\textwidth]{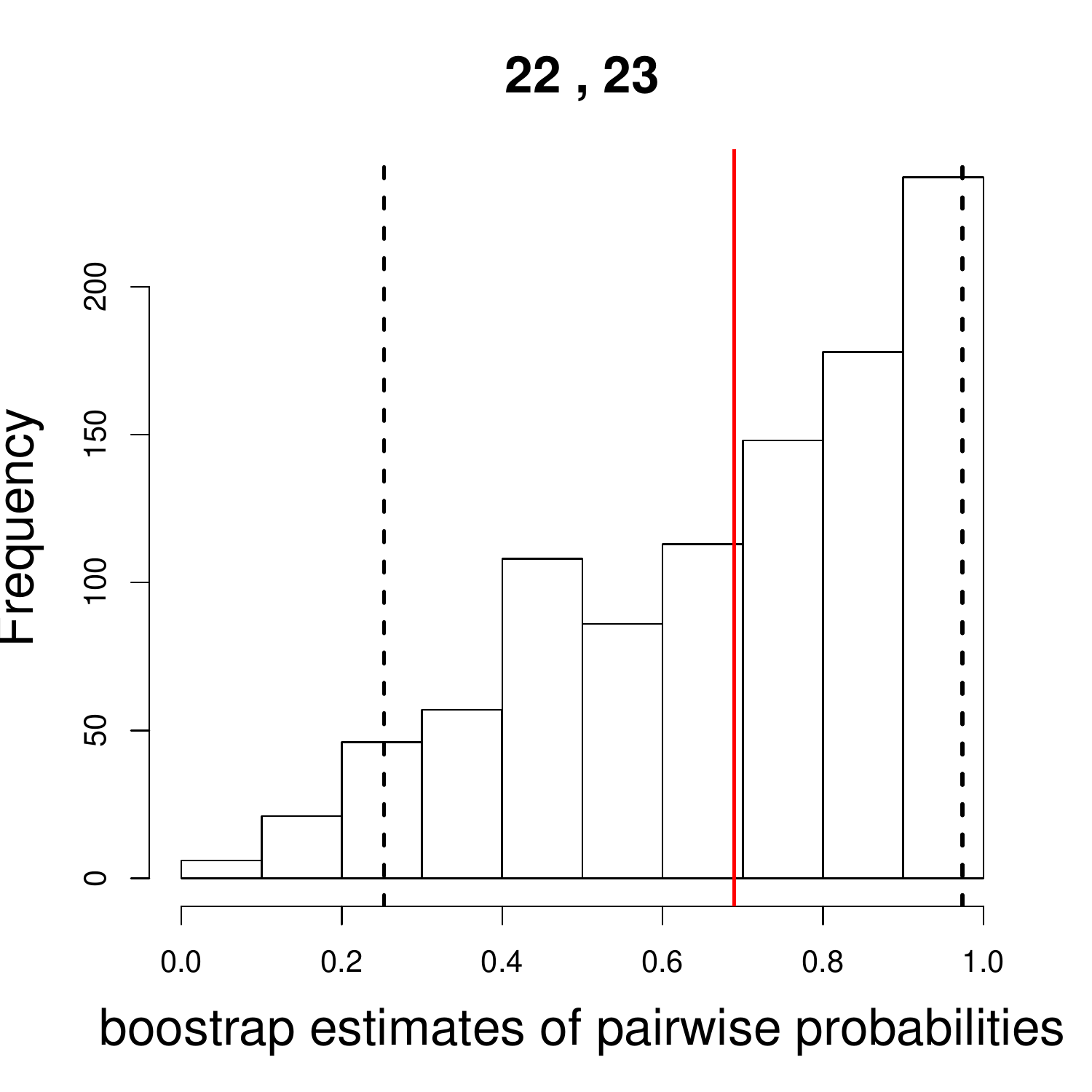}}\\
\subfloat{\includegraphics[width=0.4\textwidth]{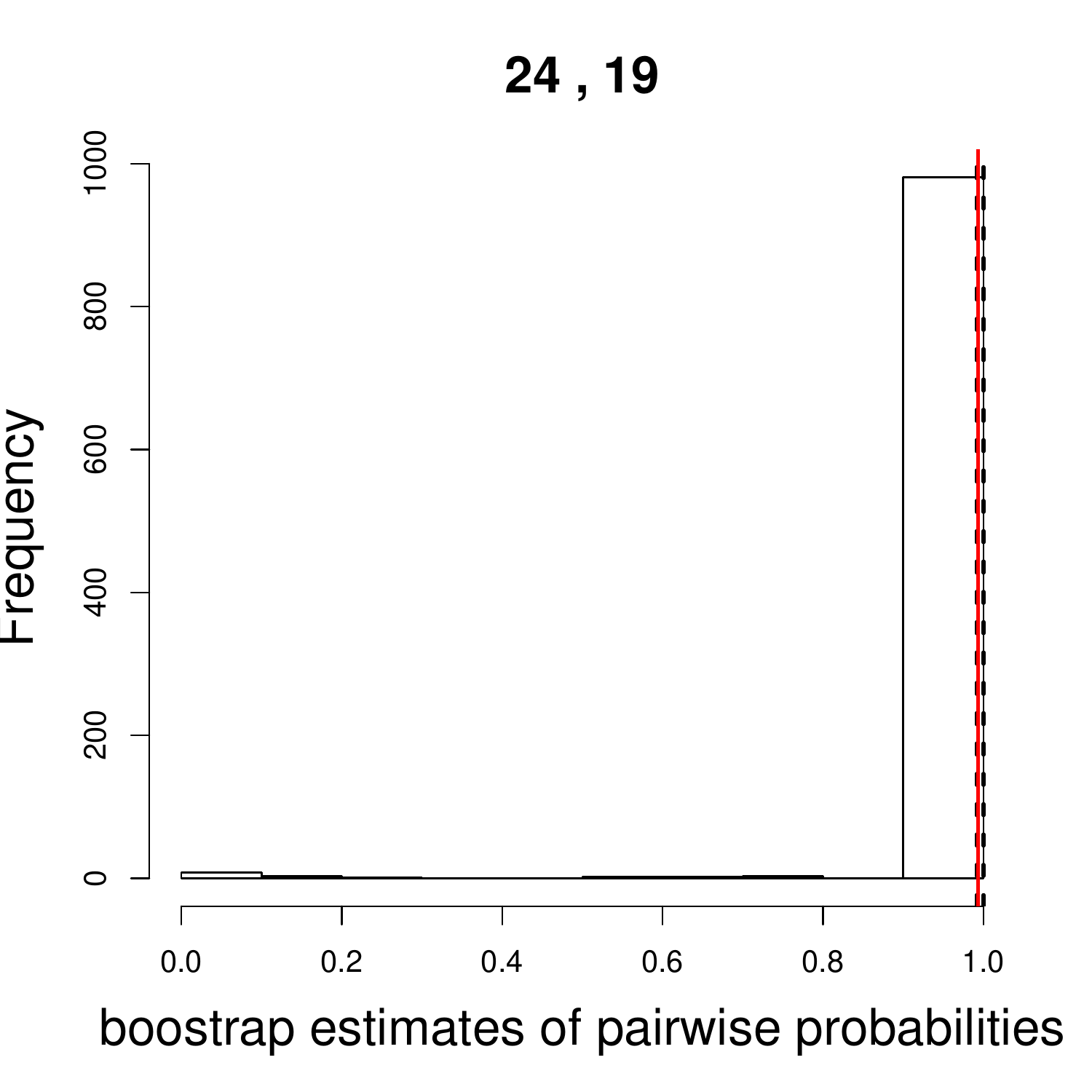}}
\subfloat{\includegraphics[width=0.4\textwidth]{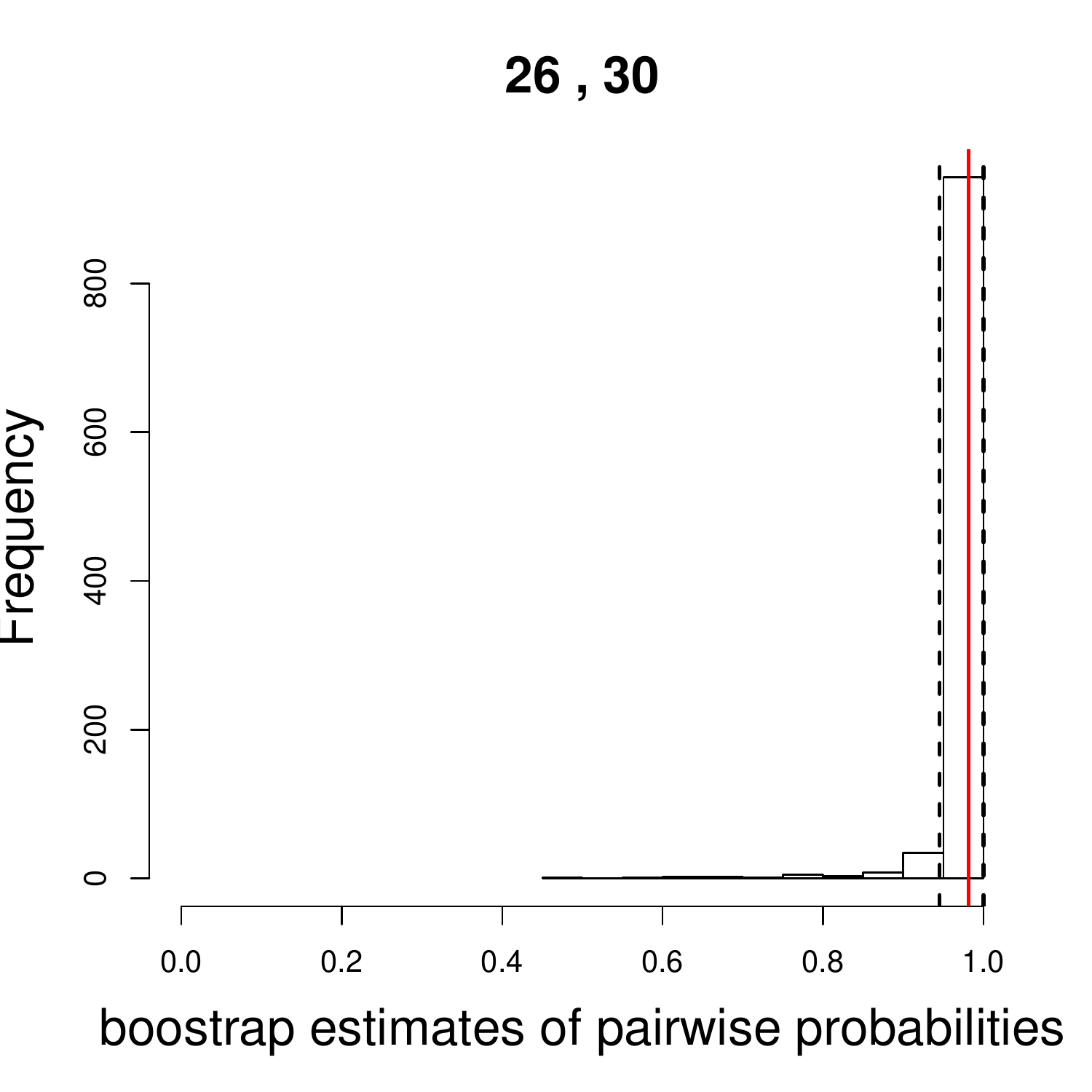}}
\caption{Histograms of bootstrap estimates $P_{ij}(\bthetah_b)$, $b=1,\ldots,1000$ for four pairs of objects $(i,j)$ in the dataset of Example \ref{ex:ex3} (see Figure \ref{fig:ex3}).  The  black broken vertical lines are the 0.025 and 0.975 quantiles $P_{ij}^l$ and  $P_{ij}^u$. The red solid vertical line is the true probability $P_{ij}(\btheta)$.   (This figure is better viewed in color).\label{fig:ex3_histos}}
\end{figure}
\end{Ex}

\subsection{Construction of an evidential partition}
\label{subsec:algo}

The $n(n-1)/2$ confidence intervals computed as described in the previous section are not easily interpretable. To obtain a simple and more user-friendly representation, we propose to construction an evidential partition $M=(m_1,\ldots,m_n)$ such that, for all pairs $(i,j)$ of objects, $Bel_{ij}(\{s_{ij}\})$ and $Pl_{ij}(\{s_{ij}\})$ as computed by \eqref{eq:BelPlij} approximate, respectively, the confidence bounds $P_{ij}^l$ and $P_{ij}^u$. More precisely, we want to find $M$ that minimizes the error function
\begin{equation}
J(M):=\sum_{i<j} \left(Bel_{ij}(\{s_{ij}\})-P_{ij}^l\right)^2 + \left(Pl_{ij}(\{s_{ij}\})-P_{ij}^u\right)^2.
\end{equation}
Using the equalities $Bel_{ij}(\{s_{ij}\})=m_{ij}(\{s_{ij}\})$ and $Pl_{ij}(\{s_{ij}\})=1-Bel_{ij}(\{\neg s_{ij}\})=1-m_{ij}(\{\neg s_{ij}\})$, we get
\begin{equation}
\label{eq:J}
J(M)=\sum_{i<j} \left(m_{ij}(\{s_{ij}\})-P_{ij}^l\right)^2 + \left(m_{ij}(\{\neg s_{ij}\})-(1-P_{ij}^u)\right)^2.
\end{equation}
Assuming that $Bel_{ij}(\{s_{ij}\})\approx P_{ij}^l$ and $Pl_{ij}(\{s_{ij}\})\approx P_{ij}^u$, we will have, from \eqref{eq:confint},
\begin{equation}
\label{eq:calib}
\Pr\left(Bel_{ij}(\{s_{ij}\}) \le P_{ij}(\btheta) \le Pl_{ij}(\{s_{ij}\})\right) \approx 1-\alpha.
\end{equation}
Eq. \eqref{eq:calib} corresponds to the definition of a \emph{predictive belief function} at confidence  level $1-\alpha$ as introduced in \cite{denoeux06b}. It is a particular kind of frequency-calibrated belief function as reviewed in \cite{denoeux18b}.

To find an evidential partition $M$ minimizing \eqref{eq:J}, let us assume that  each mass function $m_i$ has at most $f$ nonempty focal sets in $\calF=\{F_1,\ldots,F_f\} \subseteq 2^\Omega$. If $c$ is small, we can take $\calF=2^\Omega \setminus \{\emptyset\}$. Otherwise, we can restrict the focal sets to have a cardinality less than some value (typically, 2).  Each mass function $m_i$ can then be represented by the $f$-vector $\bm_i=(m_i(F_1),\ldots,m_i(F_f))^T$. Let $\bS=(S_{k\ell})$ and $\bC=(C_{k\ell})$ be the $f\times f$ matrices with general terms
\begin{equation}
\label{eq:S}
S_{k\ell}:=\begin{cases}
                   1 & \text{if } k=\ell \text{ and } |F_k|=1, \\
                   0 & \text{otherwise}.
                 \end{cases}
 \end{equation}
and 
\begin{equation}
\label{eq:C}
  C_{kl}:=\begin{cases}
                 1 & \text{if } F_k\cap F_l=\emptyset, \\
                 0 & \text{otherwise.}
               \end{cases}
\end{equation}
Furthermore, let $\bB$ be the $2f\times f$  block matrix 
\[
\bB:=\begin{pmatrix}
\bS\\
\bC
\end{pmatrix},
\]
and let  $\bA_j$ be the $2\times2f$ matrix defined by
\begin{equation}
\label{eq:Aj}
\bA_j:=\begin{pmatrix}
1 & 0\\
0 & 1
\end{pmatrix} \otimes \bm_j^T,
\end{equation}
where  $\otimes$ is the Kronecker product. 

With these notations, from \eqref{eq:mij}, we have $m_{ij}(\{s_{ij}\})=\bm_j^T\bS\bm_i$, $m_{ij}(\{\neg s_{ij}\})=\bm_i^T\bC\bm_i$, and
\begin{equation}
\label{eq:mij1}
\bm_{ij}=\bA_j\bB\bm_i,
\end{equation}
with $\bm_{ij}=(m_{ij}(\{s_{ij}\}),m_{ij}(\{\neg s_{ij}\}))^T$. Eq.  \eqref{eq:J} can thus be rewritten as
\begin{subequations}
\label{eq:Jvec}
\begin{align}
J(M) &= \sum_{i<j} (\bm_{ij}-\bm^*_{ij})^T(\bm_{ij}-\bm^*_{ij})\\
&=\sum_{i<j} (\bA_j\bB\bm_i-\bm^*_{ij})^T(\bA_j\bB\bm_i-\bm^*_{ij}), \label{eq:Jvec2}
\end{align}
\end{subequations}
with $\bm^*_{ij}=(P_{ij}^l,1-P_{ij}^u)^T$. From  \eqref{eq:Jvec2},  we can see that $J(M)$ is a quadratic function of $\bm_i$. We can then use the Iterative Row-wise Quadratic Programming (IRQP) algorithm introduced by \cite{terbraak09}. The IRQP is a block cyclic  coordinate descent procedure \cite[Section 2.7]{bertsekas99} minimizing $J(M)$ with respect to each vector $\bm_i$ one at a time, while keeping the other vectors $\bm_j$ for $j\neq i$ fixed. At each iteration, we thus minimize
\begin{subequations}
\label{eq:QP}
\begin{equation}
\label{eq:g}
J_i(\bm_i):=\sum_{\substack{j=1\\j\neq i}}^n (\bA_j\bB\bm_i-\bm^*_{ij})^T(\bA_j\bB\bm_i-\bm^*_{ij})
\end{equation}
subject to 
\begin{equation}
\label{eq:constr}
\bone^T \bm_i=1 \quad \textrm{and}  \quad \bm_i \ge \bzero.
\end{equation} 
\end{subequations}
Developing the expression in the right-hand side of \eqref{eq:g}, we get
\begin{equation}
\label{eq:g1}
J_i(\bm_i)= \bm_i^T \bQ_i \bm_i + \bu_i^T \bm_i + a_i
\end{equation}
with
\begin{subequations}
\label{eq:Qu}
\begin{align}
\label{eq:Qu1}
\bQ_i&:= \bB^T \left(\sum_{j\neq i} \bA_j^T\bA_j\right) \bB\\
\bu_i&:= -2 \left(\sum_{j\neq i} (\bm^*_{ij})^T \bA_j\right) \bB \\
a_i &:= \sum_{j\neq i} (\bm^*_{ij})^T \bm^*_{ij}.
\end{align}
\end{subequations}
The minimization of function $J_i$ in \eqref{eq:g1} subject to constraints \eqref{eq:constr} can be performed using a standard quadratic programming solver. To define a stopping criterion, we compute a running mean of the relative error as follows:  $e^{(0)} = 1$ and
\begin{equation}
e^{(t)} := \rho \; e^{(t-1)} + (1-\rho) \frac{|J^{(t)}-J^{(t-1)}|}{J^{(t-1)}}, \quad t=1,2,\ldots,
\end{equation}
where $t$ is the iteration counter, $J^{(t)}$ is the value of the cost function at iteration $t$,  and $\rho=0.5$. The algorithm is then stopped when $e^{(t)}< \epsilon$, for some threshold $\epsilon$. The whole procedure is summarized in Algorithm \ref{alg:IRQP}.

As matrix $\bQ_i$ in \eqref{eq:Qu1} is positive definitive, the quadratic programming problem \eqref{eq:QP} is convex \cite{Vavasis2001} and has a unique solution. Consequently, the whole block coordinate descent procedure is guaranteed to converge to a local minimum \cite[Proposition 2.7.1]{bertsekas99}.

\begin{algorithm}
\caption{IRQP algorithm. \label{alg:IRQP}}
\begin{algorithmic}[1]
\REQUIRE Confidence intervals $\bm_{ij}^*$ for $1\le i \le j \le n$, number of clusters $c$, focal sets $\calF=\{F_1,\ldots,F_f\}$, stopping threshold $\epsilon$
\STATE Initialize the evidential partition $M$ randomly
\STATE $t := 0$, $e^{(0)} := 1$
\STATE Compute $J^{(0)}$ using (\ref{eq:Jvec})
\WHILE{$e^{(t)} \ge \epsilon$}
\STATE $t := t+1$ 
\STATE $J^{(t)}:= 0$
\FOR{$i = 1$ \TO $n$}
\STATE Compute $\bQ_i$ and $\bu_i$ in \eqref{eq:Qu}
\STATE Find $\bm_i^{(t)}$ by minimizing \eqref{eq:g1} subject to \eqref{eq:constr} 
\STATE Update $M$ with $\bm_i^{(t)}$
\STATE $J^{(t)} := J^{(t)}+  J_i(\bm_i^{(t)})$
\ENDFOR
\STATE $e^{(t)} := 0.5 \; e^{(t-1)} + 0.5 \vert J^{(t)}-J^{(t-1)} \vert /J^{(t-1)}$
\ENDWHILE
\ENSURE Evidential partition $M$
\end{algorithmic}
\end{algorithm}

\begin{Ex}
\label{ex:ex4}
The procedure described in this section was applied to the data and bootstrap confidence intervals of Example \ref{ex:ex3}. The set $\calF$ of focal sets was defined to contain the singletons and the pairs, i.e., 
\[
\calF=\left\{\{\omega_1\},\{\omega_2\}, \{\omega_3\},\{\omega_1,\omega_2\}, \{\omega_1,\omega_3\},\{\omega_2,\omega_3\}\right\},
\]
and $f=6$. Figure \ref{fig:ex3_approx} shows the pairwise degrees of belief $Bel_{ij}(\{s_i\})$ and plausibility $Pl_{ij}(\{s_i\})$ as functions of, respectively, the lower bounds $P^l_{ij}$ and the lower bounds $P^u_{ij}$ of the bootstrap percentile 90\% confidence intervals. Figure \ref{fig:ex3_credpart} presents a view of of the resulting evidential partition, showing  the maximum-plausibility hard partition as well as the convex hulls of the lower and upper approximations of each cluster \cite{masson08}. These approximations  are obtained by first assigning each object $i$ to the set of clusters $A_i\subseteq \Omega$ with the highest mass, and then computing the lower and upper approximation defined by \eqref{eq:lower_upper}. The lower approximation of cluster $k$ contains the objects that \emph{surely} belong to that cluster, while the upper approximation contain those objects that \emph{possibly} belong to cluster $k$. We can see that objects 3, 10 and 22 are ambiguous. Their mass functions, as well as those of three other objects are shown in Table \ref{tab:ex3_masses}.

\begin{figure}
\centering  
\subfloat{\includegraphics[width=0.5\textwidth]{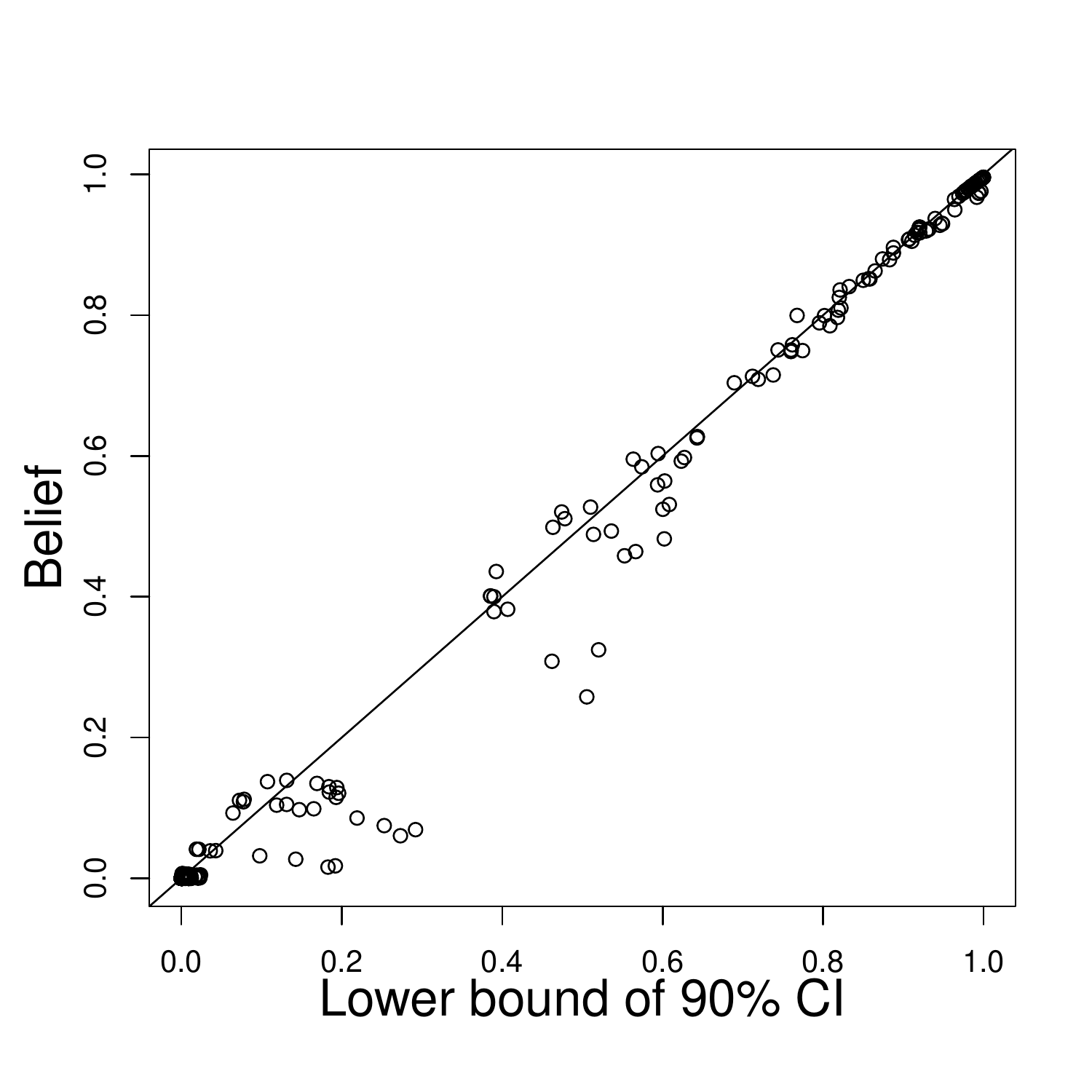}}
\subfloat{\includegraphics[width=0.5\textwidth]{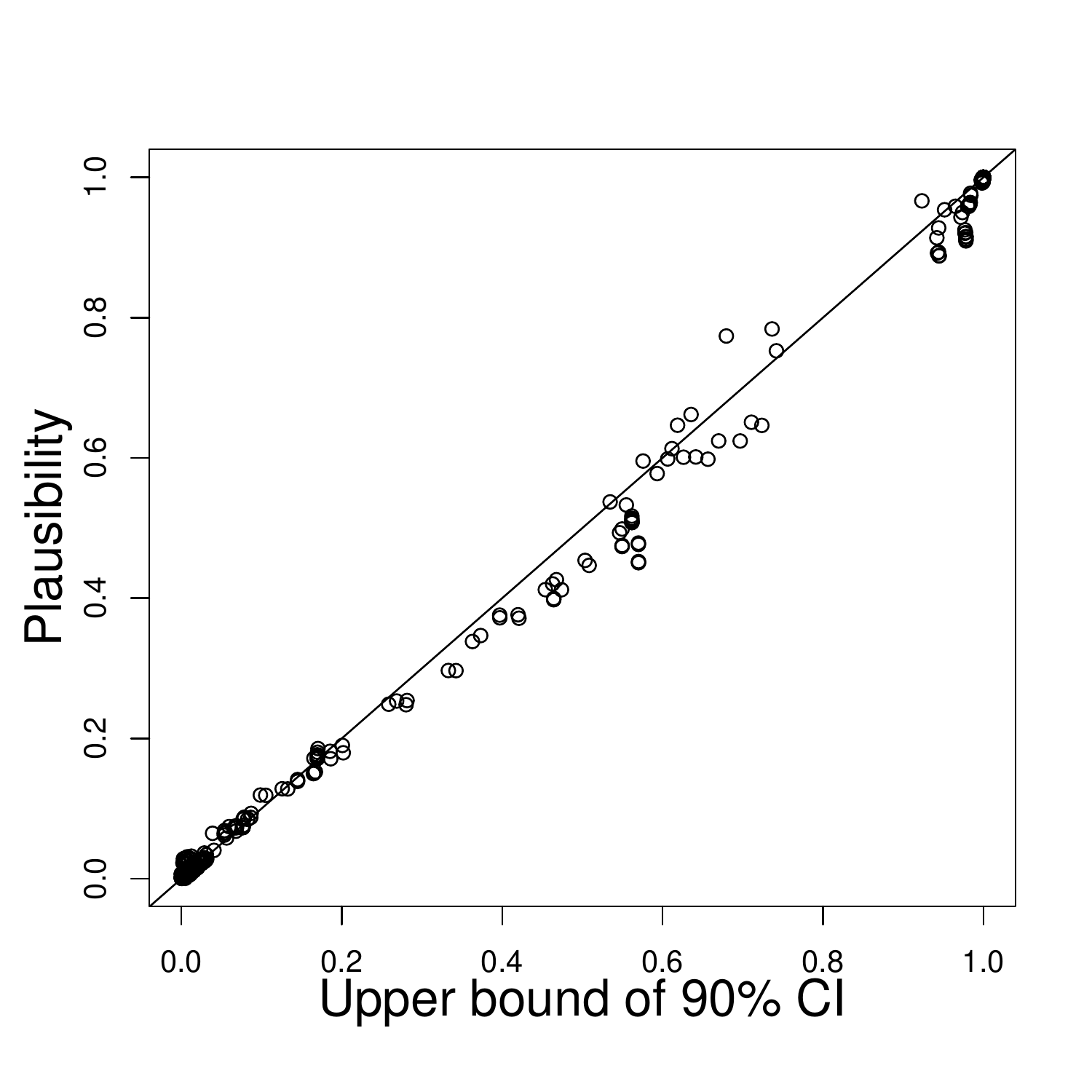}}
\caption{Pairwise degrees of belief $Bel_{ij}(\{s_i\})$ (left) and plausibility $Pl_{ij}(\{s_i\})$ (right) as functions of, respectively, the lower bounds $P^l_{ij}$ and the lower bounds $P^u_{ij}$ of  bootstrap percentile 90\% confidence intervals, for the data of Example \ref{ex:ex4}.   \label{fig:ex3_approx}}
\end{figure}

\begin{figure}
\centering  
\includegraphics[width=0.6\textwidth]{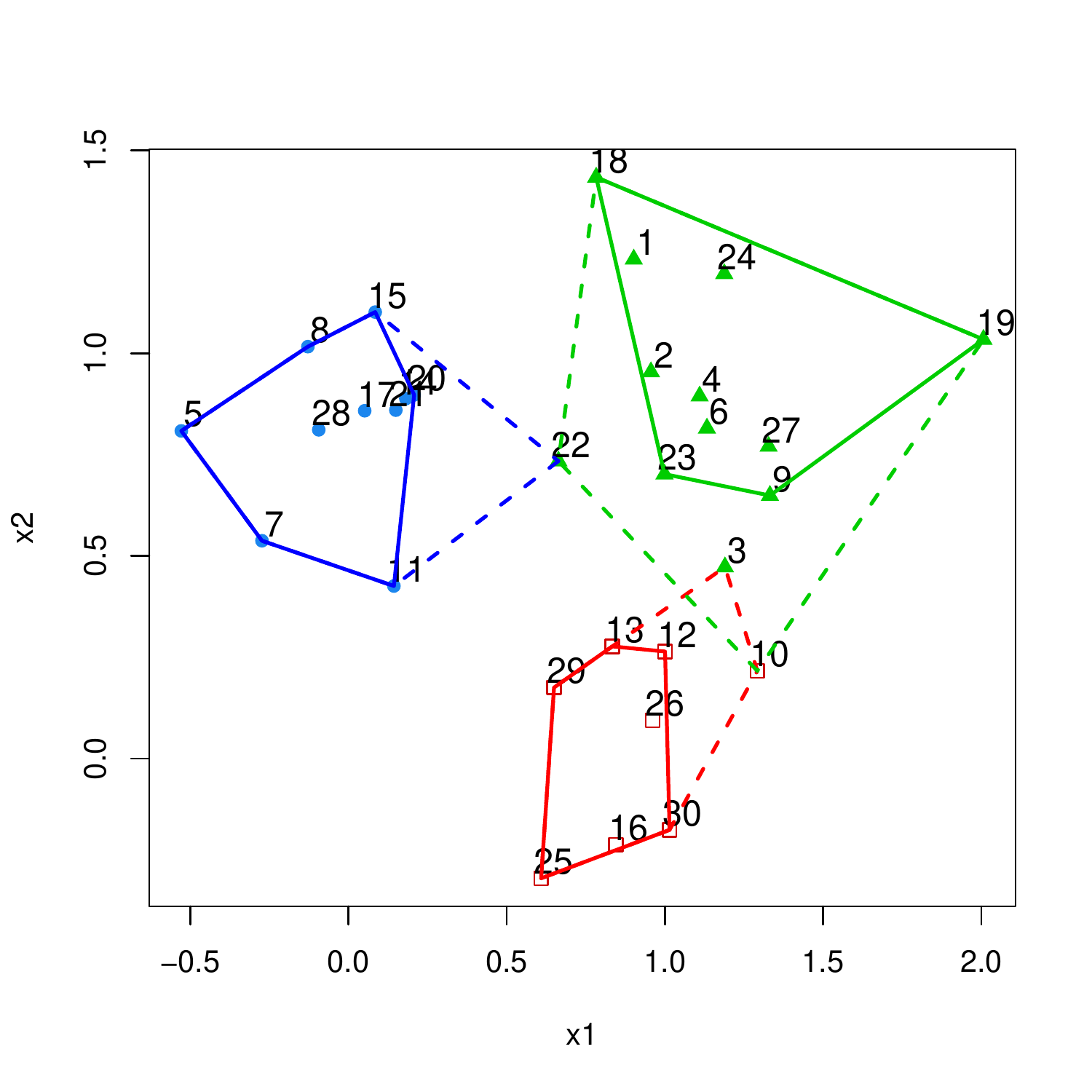}
\caption{Evidential partition of the data of Example  \ref{ex:ex4}.  The solid and broken lines represent, respectively, the convex hulls of the lower and upper approximation of each cluster, as defined in the text. The lower approximation of a cluster contains the objects that can confidently be assigned to that cluster, while the upper approximation contains objects that may belong to several clusters. For instance, Object 22 may belong to clusters $\omega_1$ (left)  or $\omega_3$ (top right), while object 3 may clusters $\omega_2$ (bottom right)  or $\omega_3$ (bottom right). \label{fig:ex3_credpart}}
\end{figure}
\end{Ex}

\begin{table}
\centering
\caption{Mass functions for six objects displayed in Figure  \ref{fig:ex3_credpart}. For each object, the largest mass is printed in bold. \label{tab:ex3_masses} }
\begin{tabular}{ccccccc}
\hline
Object & $m(\{\omega_1\})$  & $m(\{\omega_2\})$ & $m(\{\omega_3\})$ & $m(\{\omega_1,\omega_2\})$ & $m(\{\omega_1,\omega_3\})$ & $m(\{\omega_2,\omega_3\})$ \\
\hline
3   & 0 &  0.042&   0.113 &  0 & 0 & \textbf{0.845} \\
4   &0& 0 &  \textbf{0.926} &  0  & 0 &  0.074 \\
10 &0 &  0.406 &  0.007 &  0 & 0 &   \textbf{0.587} \\
11 & \textbf{0.927} &  0 &  0 &  0.073 &  0 & 0 \\
12 &0 &   \textbf{0.635} &  0.005 &  0 & 0 &  0.360 \\
22 &0 &  0 & 0.141 &  0.092 &   \textbf{0.415} &  0.352\\
\hline
\end{tabular}
\end{table}

\subsection{Complexity analysis}

The propose clustering methods consists in three steps:
\be
\item The computation of the estimates $\bthetah_b$ and $P_{ij}(\bthetah_b)$ for each of the $B$ bootstrap samples (lines 1-7 in Algorithm \ref{alg:CI})
\item The computation of the quantiles $P_{ij}^l$ and $P_{ij}^u$ (lines 8-11  in Algorithm \ref{alg:CI});
\item The construction of the evidential partition  (Algorithm \ref{alg:IRQP}).
\ee
In Step 1, each iteration of the EM has complexity $O(cn)$. Assuming that the number of iterations is roughly constant and does not depend on $n$, the computation of each estimate $\btheta_b$ has complexity $O(cn)$, and the computation of  $P_{ij}(\bthetah_b)$ for all $i<j$ involves $O(cn^2)$ operations. So, the complexity of Step 1 is $O(Bcn^2)$. In Step 2, each quantile can be computed in $O(n)$ operations \cite{blum73}, so the complexity of Step 2 is $O(Bn^2)$.  Finally, solving each quadratic programming problem in Step 3 has worst-case complexity $O(f^3)$, where $f$ is the number of focal sets, so that each iteration of Algorithm \ref{alg:IRQP} has $O(nf^3)$ complexity. Assuming the number of iterations of the IRQP algorithm to be roughly constant, the complexity of Step 3 is $O(nf^3)$. Overall, the time complexity of the global procedure is $O(Bcn^2+nf^3)$. As far as storage space is concerned, we need to store the confidence intervals, which has $O(n^2)$ space complexity, and the evidential partition, which takes $O(nf)$ space, so that the overall complexity is $O(n^2+nf)$.

In the worst case, the number of nonempty focal sets is $2^c-1$. It is thus crucial to limit the number of focals sets when $c$ is large. A simple strategy is to restrict the focal sets of mass functions $m_i$ in the evidential partition to singletons and pairs, which bring their number down to $c(c+1)/2$. A more sophisticated strategy, proposed in \cite{denoeux16a} is to first identify the pairs of overlapping clusters, and to use only these pairs (as well as the singletons) as focal sets; this strategy will be illustrated in Section \ref{subsec:real} with the \textsf{GvHD} dataset.

Another limitation of our method is its $O(n^2)$ complexity, which precludes application to very large datasets. We can remark that our approach is especially useful with small and medium-size datasets (typically, containing a few hundred or thousand objects), for which the cluster-membership probabilities usually cannot be estimated accurately. Nevertheless, some preliminary ideas to make our approach applicable to large datasets will be mentioned in the last paragraph of Section \ref{sec:concl} as directions for future work.

\section{Experimental results}
\label{sec:results}

We first present results with simulated data in Section \ref{subsec:simul} to verify the calibration property experimentally. Some results with real datasets are then reported in Section \ref{subsec:real}. All the simulations reported in this section were performed using an  implementation of our algorithm in R publicly available  as function {\tt bootclus} in  package {\tt evclust} \cite{denoeux20}.

\subsection{Simulated data}
\label{subsec:simul}

\begin{figure}
\centering  
\subfloat{\includegraphics[width=0.33\textwidth]{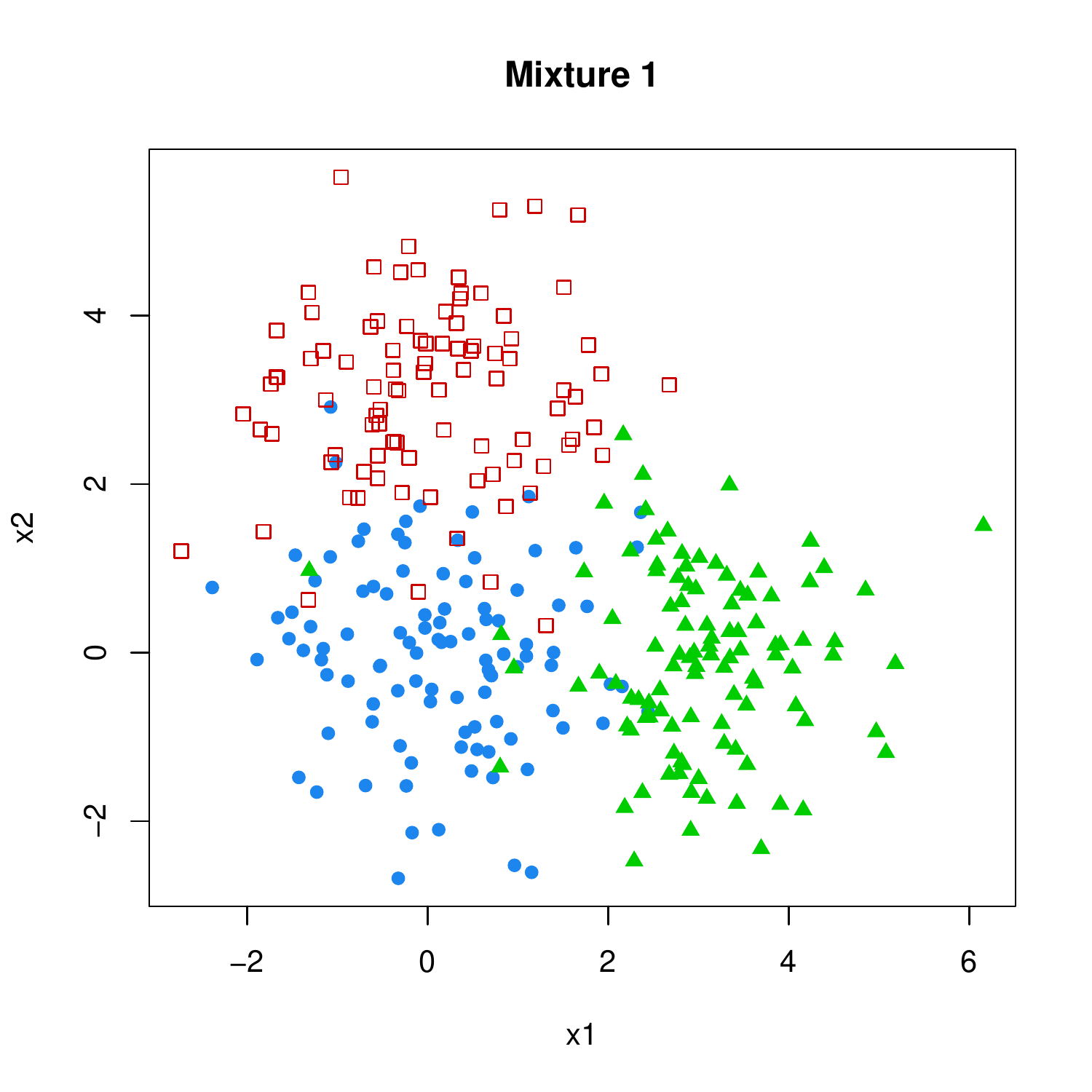}}
\subfloat{\includegraphics[width=0.33\textwidth]{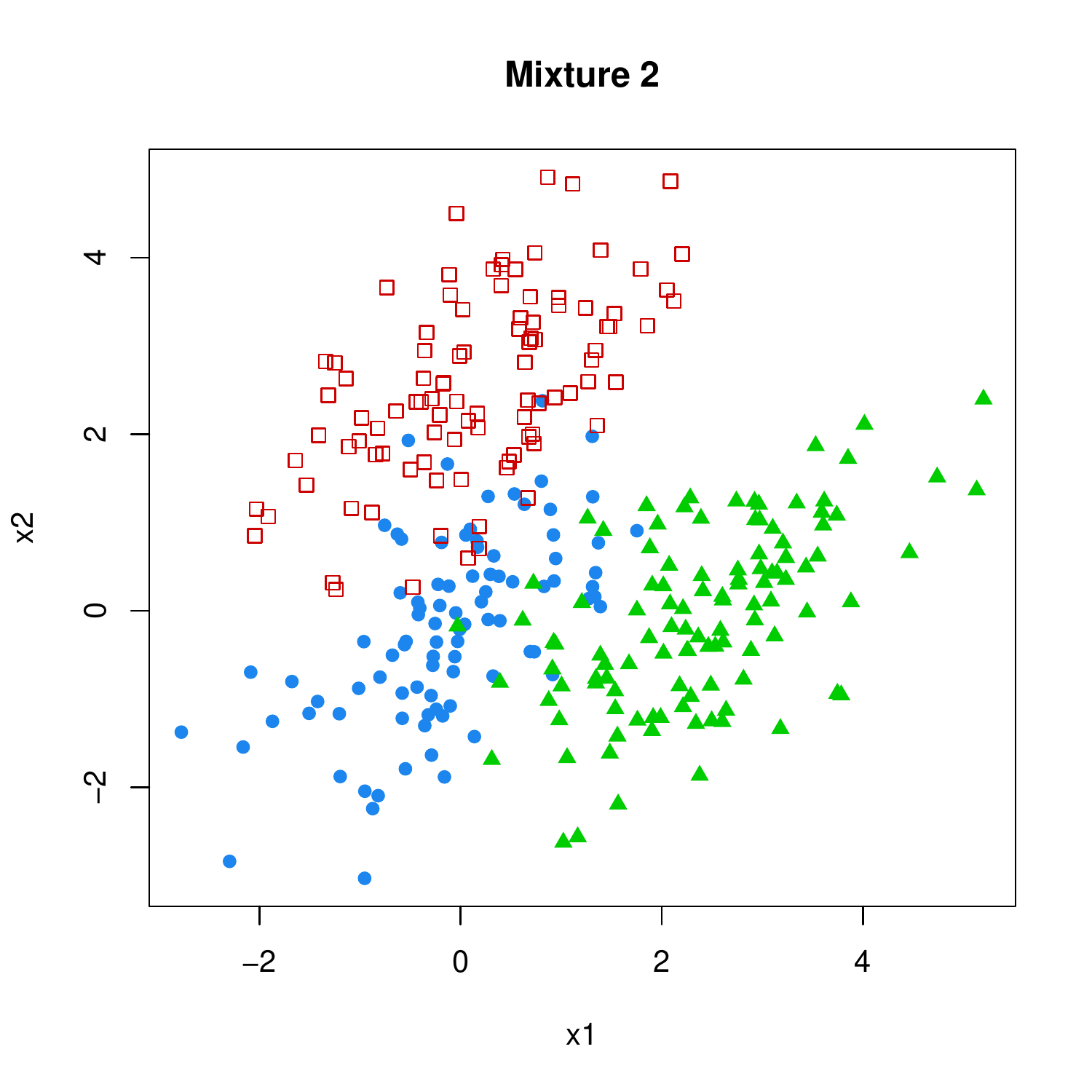}}
\subfloat{\includegraphics[width=0.33\textwidth]{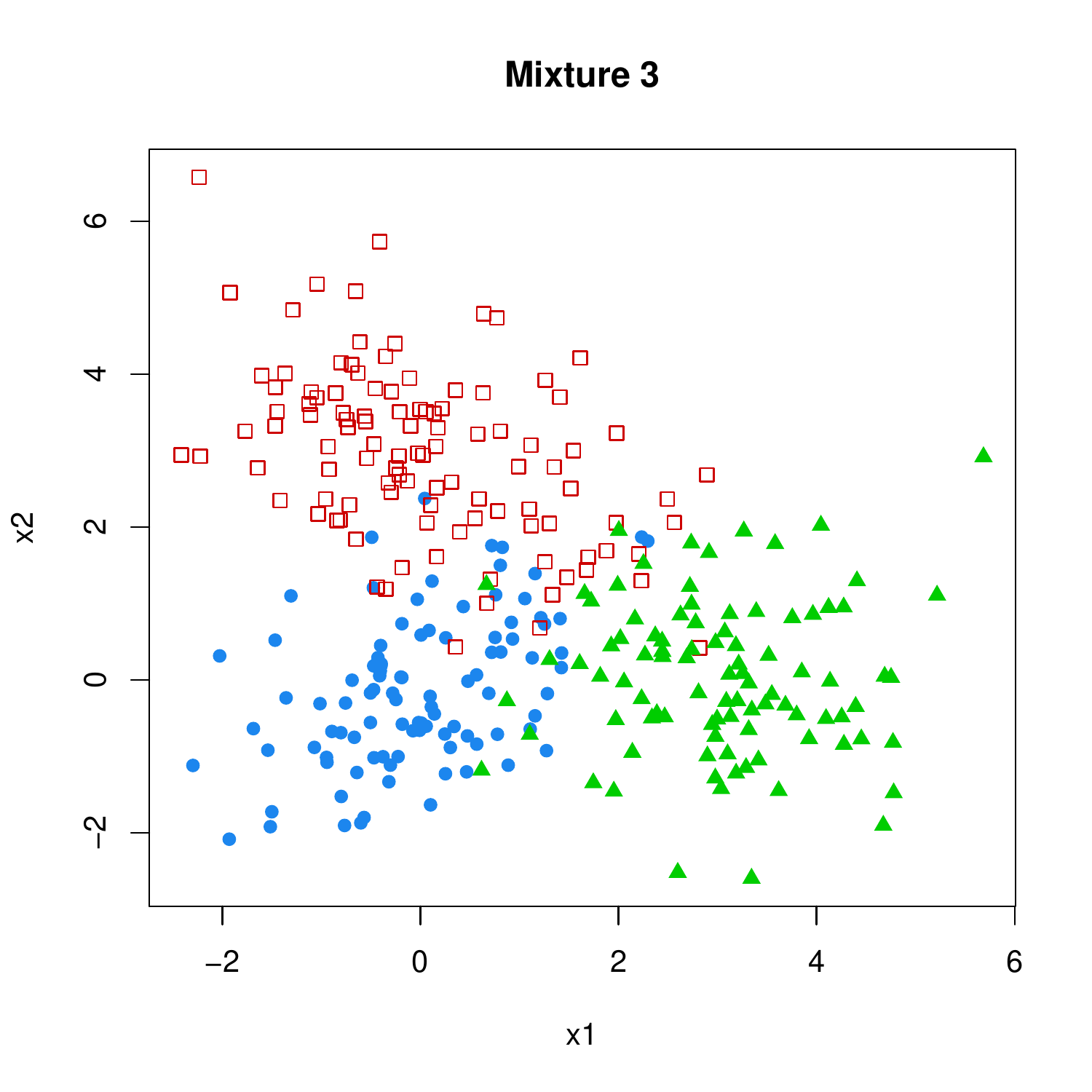}}\\
\caption{Three datasets drawn from three Gaussian mixtures with $c=3$ components.   \label{fig:datasets}}
\end{figure}

We first considered datasets with $n=300$ observations drawn from three different  two-dimensional Gaussian mixture models (GMM) with $c=3$ components and the following parameters:

\bd
\item[Mixture 1:]
\[
\bmu_1:=(0,0)^T, \quad \bmu_2:=(0,3)^T, \quad \bmu_3:=(3,0)^T,
\]
\[
\bSigma_1=\bSigma_2 =\bSigma_3:=\begin{pmatrix} 1 & 0\\ 0 & 1 \end{pmatrix}, \quad \pi_1=\pi_2=\pi_3:=1/3.
\]
\item[Mixture 2:]
\[
\bmu_1:=(0,0)^T, \quad \bmu_2:=(0,2.5)^T, \quad \bmu_3:=(2.5,0)^T,
\]
\[
\bSigma_1=\bSigma_2 =\bSigma_3:=\begin{pmatrix} 1 & 0.5\\ 0.5 & 1 \end{pmatrix}, \quad \pi_1=\pi_2=\pi_3:=1/3.
\]
\item[Mixture 3:]
\[
\bmu_1:=(0,0)^T, \quad \bmu_2:=(0,3)^T, \quad \bmu_3:=(3,0)^T,
\]
\[
\bSigma_1:=\begin{pmatrix} 1 & 0.5\\ 0.5 & 1 \end{pmatrix}, 
\bSigma_2:=1.5 \begin{pmatrix} 1 & -0.5\\ -0.5 & 1 \end{pmatrix},
\bSigma_3:=\begin{pmatrix} 1 & 0\\ 0 & 1 \end{pmatrix},
\]
\[
 \pi_1=\pi_2=\pi_3:=1/3.
\]
\ed
We generated 100 datasets from each distribution. Examples of datasets  are shown in Figure \ref{fig:datasets}.

For each dataset, we generated $B=1000$ nonparametric bootstrap samples and we estimated the parameters of three-component GMMs under four assumptions\footnote{We used the R  package {\tt mclust} \cite{scrucca16}.}:
\be
\item Spherical distributions, equal volume (EII);
\item Ellipsoidal distributions, equal volume, shape, and orientation (EEE);
\item Ellipsoidal distributions, varying volume, shape, and orientation (VVV);
\item Best model according to the BIC criterion (Auto).
\ee
Here, the terms ``volume'', ``shape'' and ``orientation'' refer to the eigenvalue decomposition of covariance matrices:
\[
\bSigma_k = \lambda_k \bD_k \bA_k \bD_k^T,
\]
where  parameter $\lambda_k=\vert \bSigma_k \vert^{1/d}$, $\bD_k$ is a matrix with eigenvectors, 
and $\bA_k$ is a diagonal matrix whose elements are proportional to the eigenvalues of $\bSigma_k$, scaled such that $\vert \bA_k\vert = 1$. With this parameterization, each of the three sets of parameters has a geometrical interpretation: $\lambda_k$ indicates  the volume of cluster $k$, $\bD_k$ its orientation, and $\bA_k$ its shape \cite{banfield93}.

It is clear that EII, EEE and VVV are the exact models for, respectively, Mixtures 1, 2 and 3. When the model selection strategy was employed, we selected the best model on the whole dataset, and we fitted the selected model on each bootstrap replicate. For each dataset and each model, we computed bootstrap confidence intervals $[P_{ij}^l,P_{ij}^u]$ on $P_{ij}(\btheta)$ for each pair of objects $(i,j)$ using Algorithm \ref{alg:CI}, at confidence levels $\alpha=0.1$ and $\alpha=0.05$. 

\begin{figure}
\centering  
\subfloat{\includegraphics[width=0.4\textwidth]{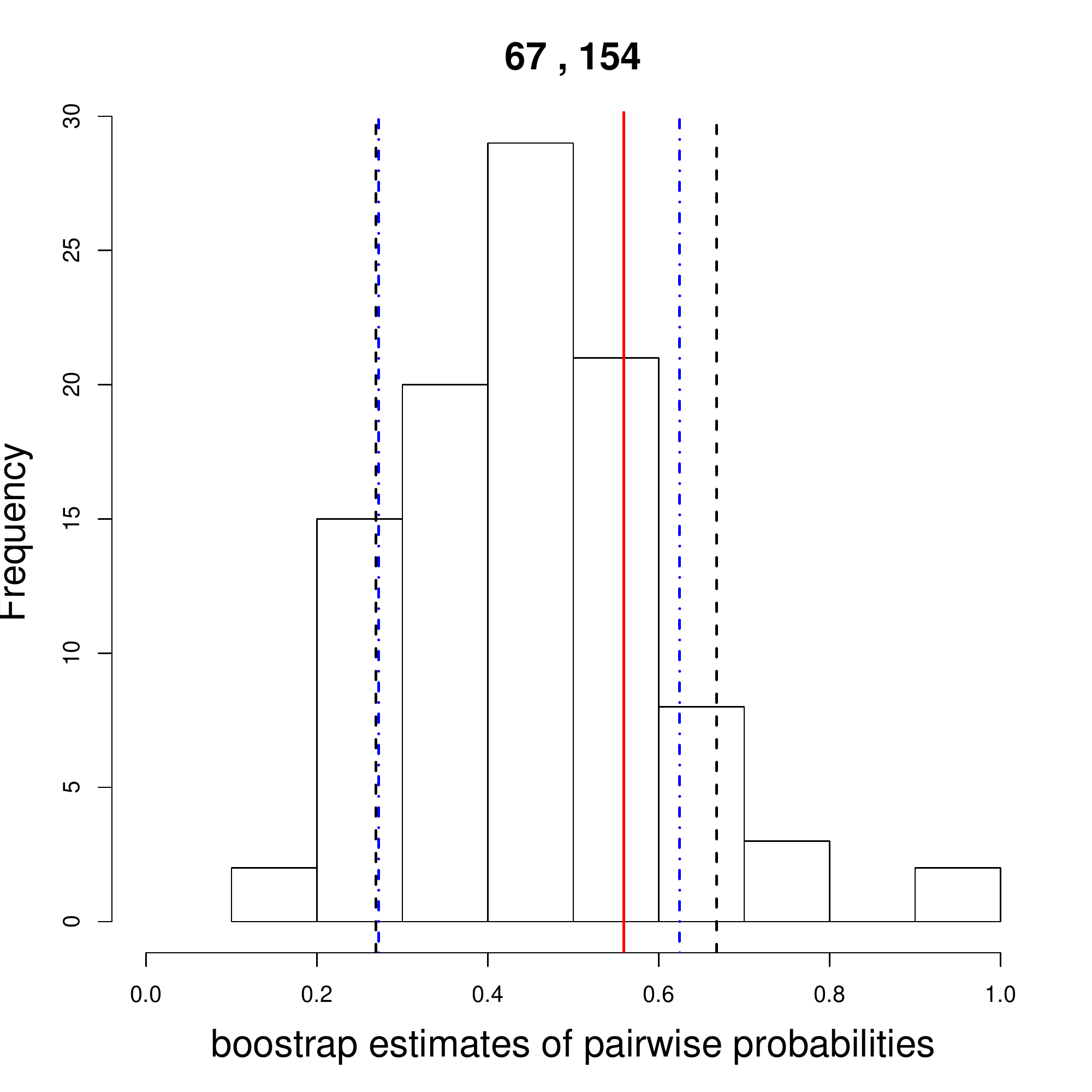}}
\subfloat{\includegraphics[width=0.4\textwidth]{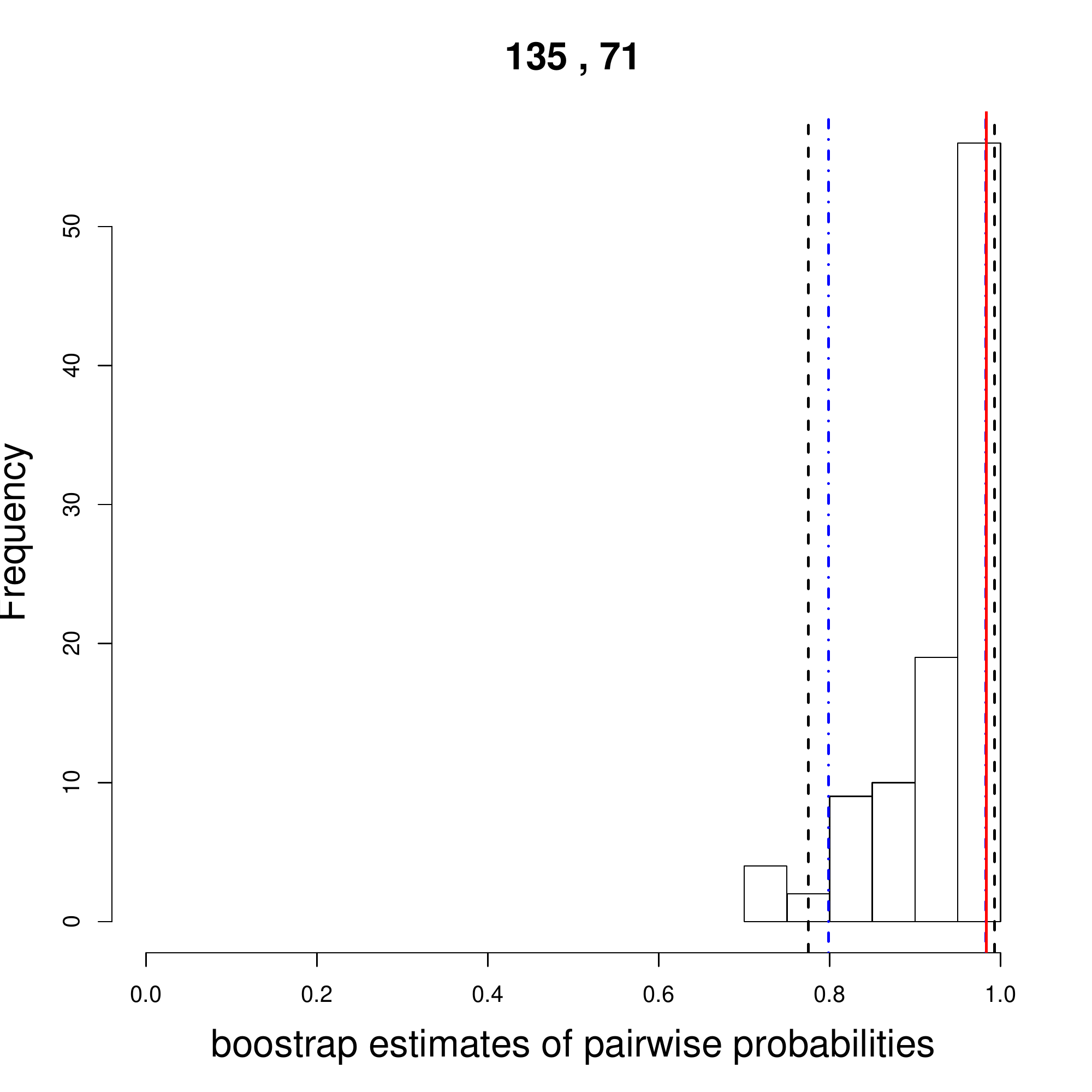}}\\
\subfloat{\includegraphics[width=0.4\textwidth]{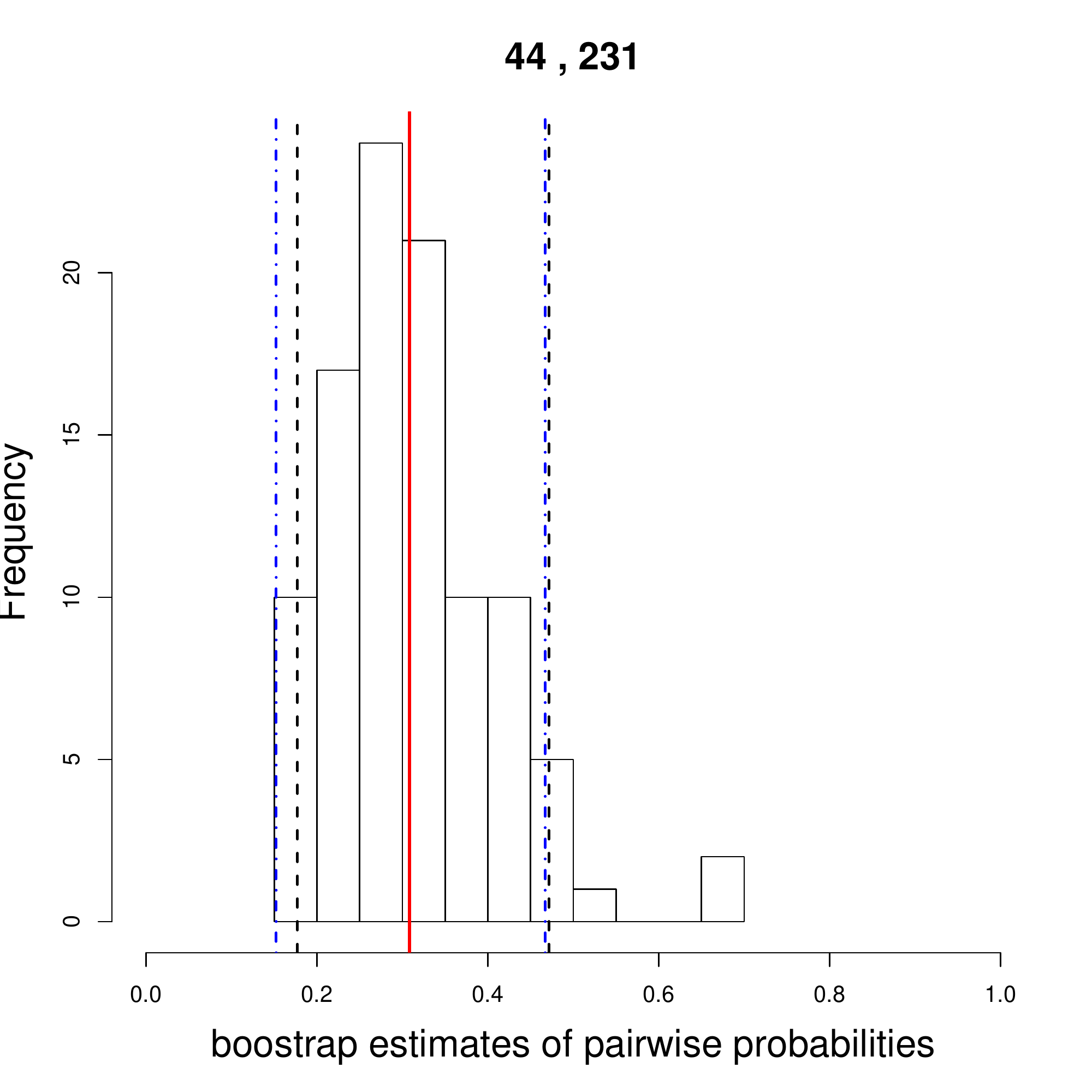}}
\subfloat{\includegraphics[width=0.4\textwidth]{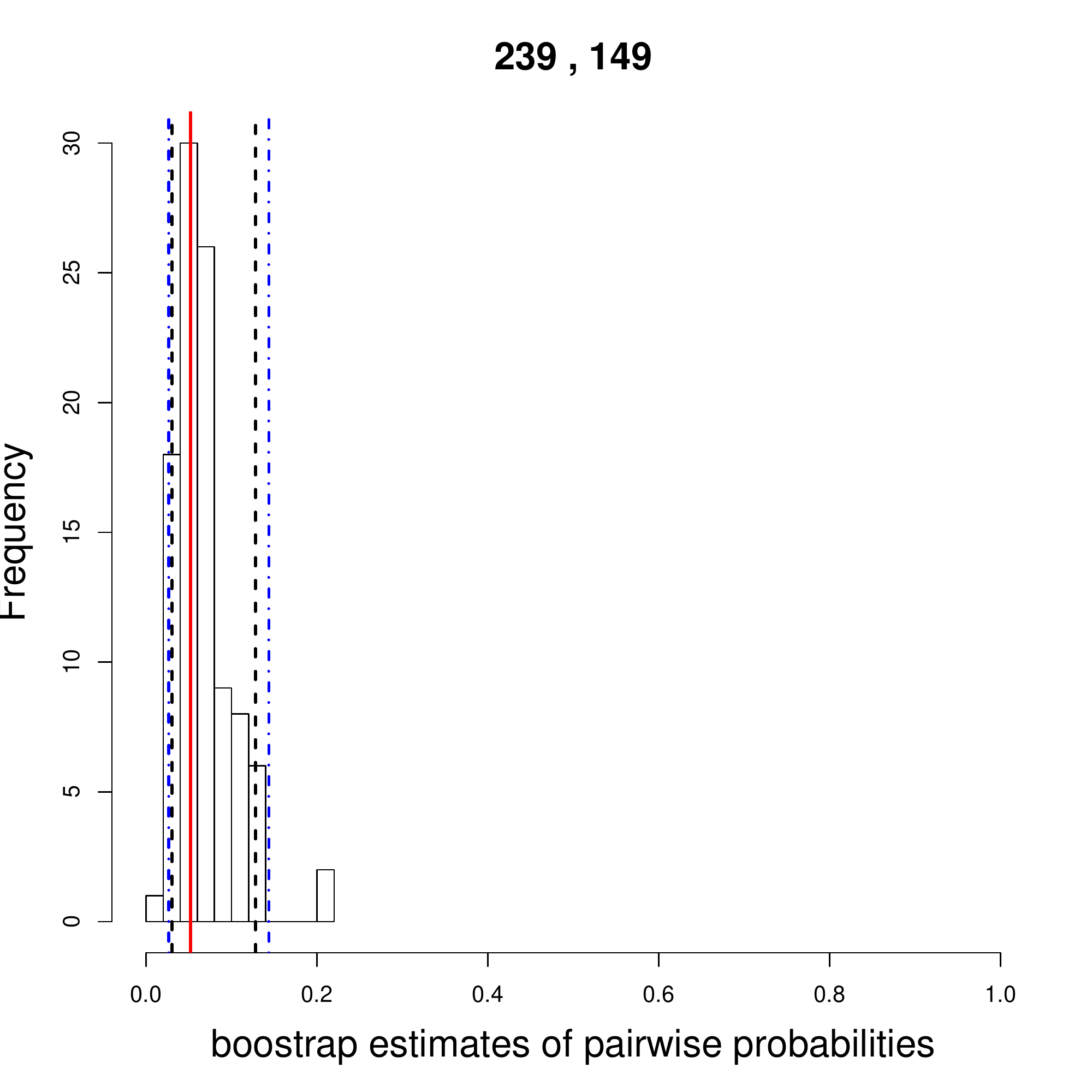}}
\caption{Histograms of bootstrap estimates $P_{ij}(\bthetah_b)$, $b=1,\ldots,1000$ for four pairs of objects $(i,j)$ in a particular dataset drawn from Mixture 2.  The  black broken vertical lines are the 0.05 and 0.95 quantiles $P_{ij}^l$ and  $P_{ij}^u$. The  blue dash-dot vertical lines are belief and plausibility degrees  $Bel_{ij}(\{s_{ij}\})$ and $Pl_{ij}(\{s_{ij}\})$. The red solid vertical line is the true probability $P_{ij}(\btheta)$. (This figure is better viewed in color).
\label{fig:confint_histo}}
\end{figure}

\begin{figure}
\centering  
\subfloat[]{\includegraphics[width=0.45\textwidth]{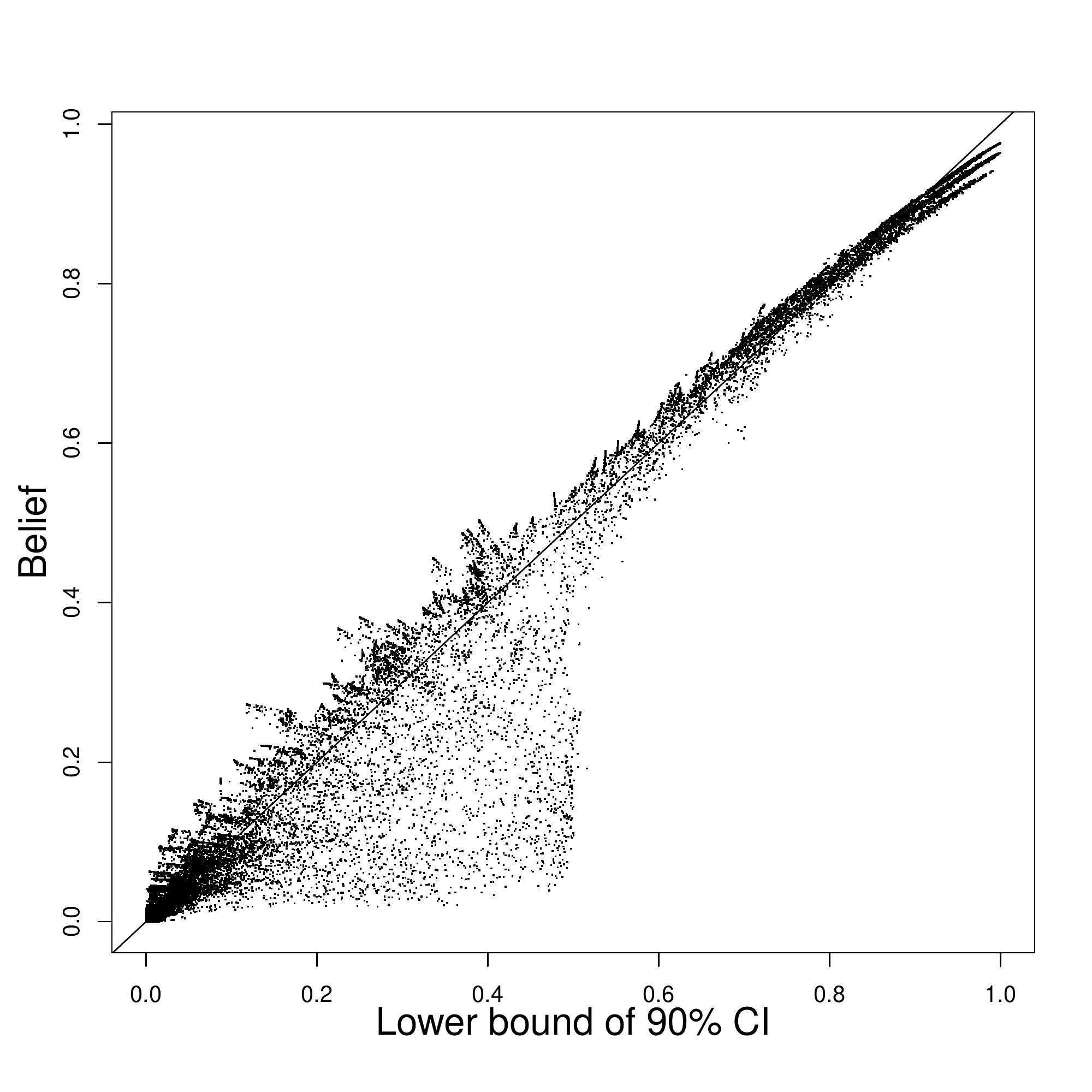}}
\subfloat[]{\includegraphics[width=0.45\textwidth]{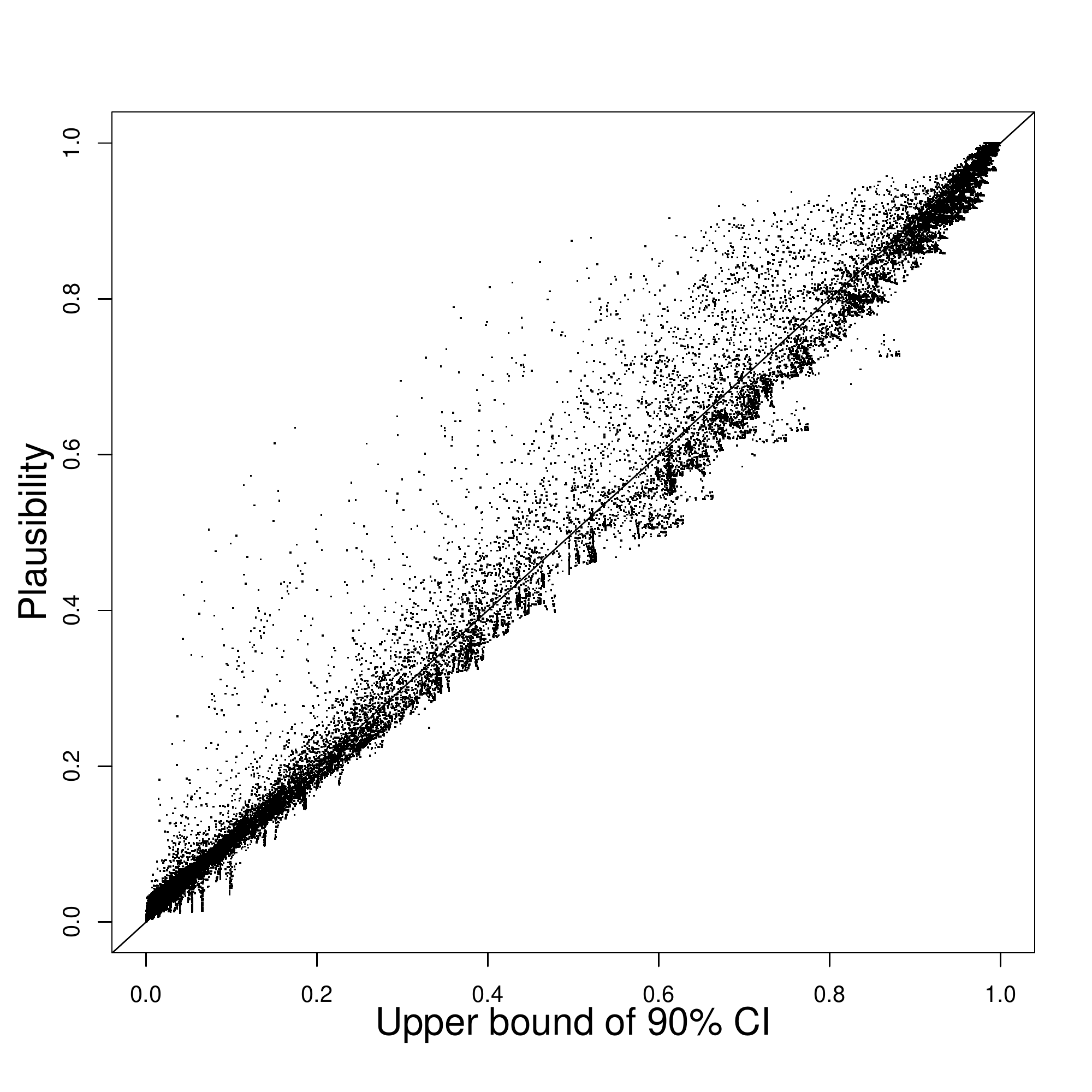}}
\caption{Dataset drawn from Mixture 2: (a) Lower bound $P_{ij}^l$ of the 90\% confidence interval on $P_{ij}(\btheta)$ ($x$-axis) vs. belief degree $Bel_{ij}(\{s_{ij}\})$ ($y$-axis); (b) Upper bound $P_{ij}^u$ of the 90\% confidence interval on $P_{ij}(\btheta)$ ($x$-axis) vs. plausibility degree $Pl_{ij}(\{s_{ij}\})$ ($y$-axis).
\label{fig:approx}}
\end{figure}

\begin{figure}
\centering  
\subfloat[]{\includegraphics[width=0.45\textwidth]{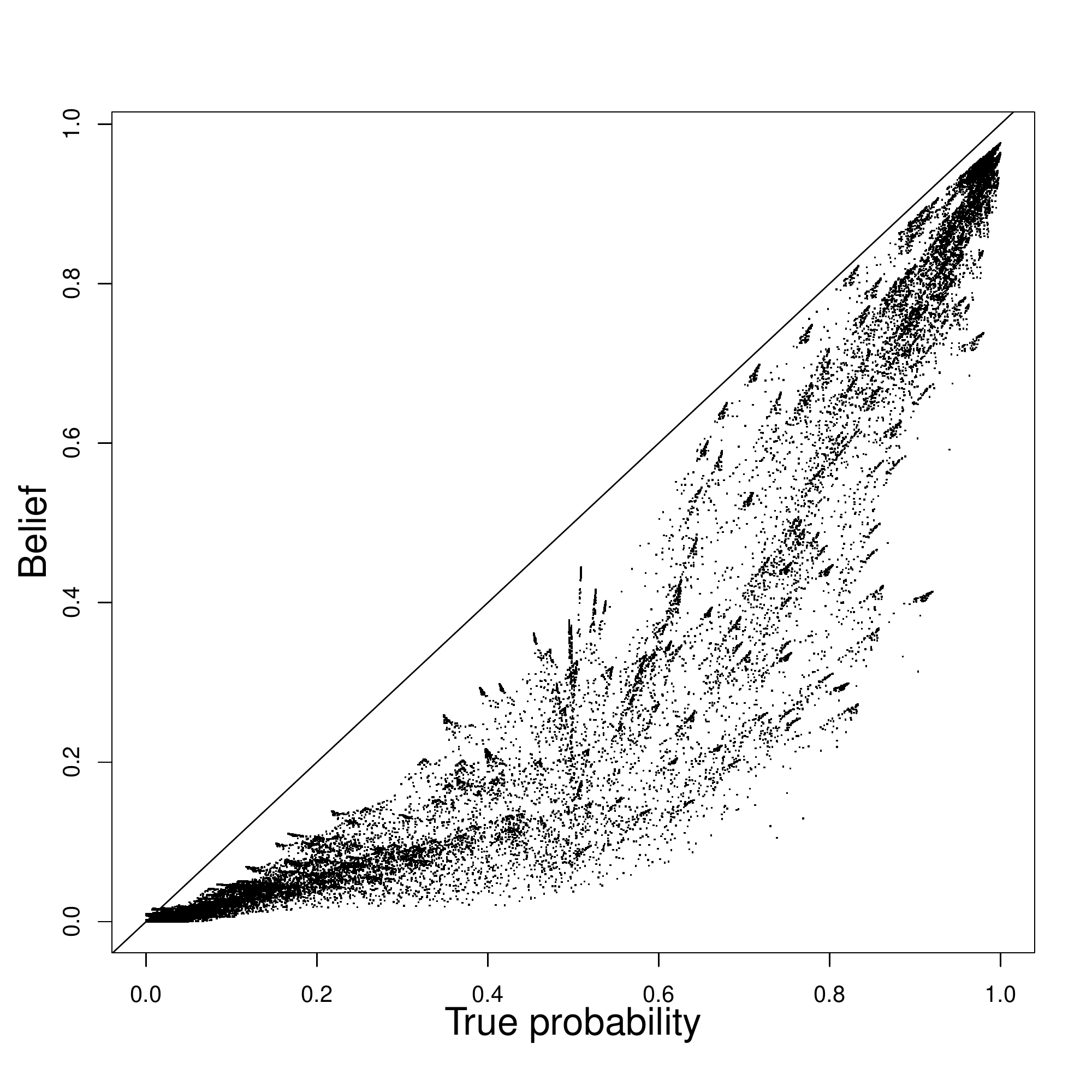}}
\subfloat[]{\includegraphics[width=0.45\textwidth]{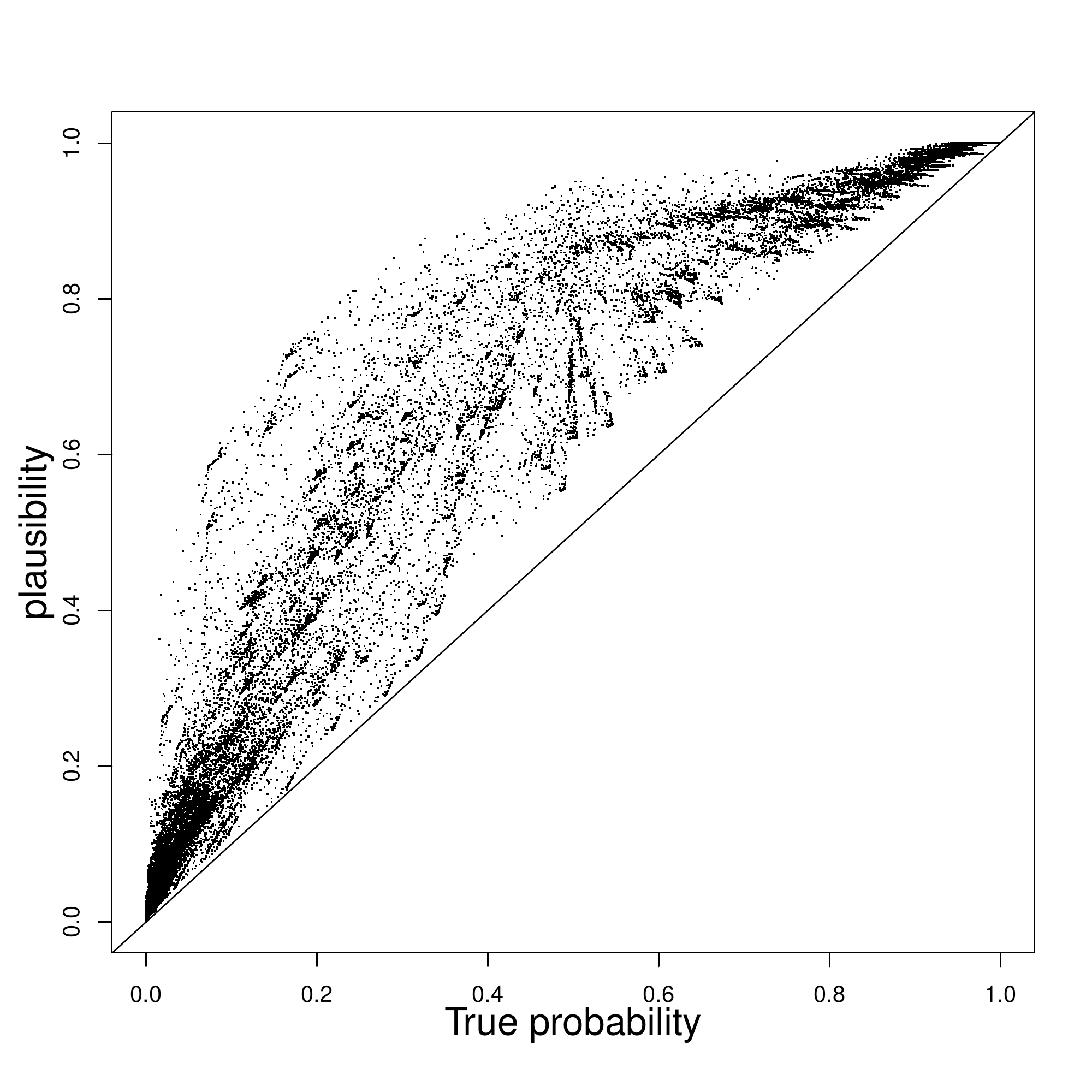}}
\caption{Dataset drawn from Mixture 2: (a) True probability $P_{ij}(\btheta)$ ($x$-axis) vs.  belief degree $Bel_{ij}(\{s_{ij}\})$ ($y$-axis); (b) True probability $P_{ij}(\btheta)$ ($x$-axis) vs.  plausibility degree $Pl_{ij}(\{s_{ij}\})$ ($y$-axis).
\label{fig:confint}}
\end{figure}

Examples of 90\% confidence intervals and approximating belief and plausibility degrees for four object pairs in one particular dataset drawn from Mixture 2 are shown in Figure \ref{fig:confint_histo}. In these four examples, both intervals $[P_{ij}^l,P_{ij}^u]$ and $[Bel_{ij}(\{s_{ij}\}),Pl_{ij}(\{s_{ij}\})]$ contain the true probability $P_{ij}(\btheta)$ that object $i$ and $j$ are in the same class. Figure \ref{fig:approx} plots the belief and plausibility degrees $Bel_{ij}(\{s_{ij}\})$ and $Pl_{ij}(\{s_{ij}\})$ vs.  the lower and upper bounds of 90\% confidence intervals on $P_{ij}(\btheta)$. We can see that there is a reasonably good fit between the belief-plausibility intervals and the bootstrap confidence intervals, thanks to the minimization of  criterion $J(M)$ in \eqref{eq:J}. The belief and plausibility degrees are plotted against the true probabilities $P_{ij}(\btheta)$ in Figure \ref{fig:confint}. For this dataset, almost all the belief-plausibility intervals contained the true probabilities.

Tables \ref{tab:res_bs1}-\ref{tab:res_bs3} show the estimated coverage probabilities (i.e., the proportion of intervals containing the true value $P_{ij}(\btheta)$ for 90\% and 95\% confidence intervals and their approximations by belief-plausibility intervals. We can see that confidence intervals and belief-plausibility intervals have similar coverage probabilities, and these probabilities are close to their nominal levels \emph{when the model is correctly specified}. For instance, in Table \ref{tab:res_bs1}, the true model is EII, which is a special case of models  EEE and VVV. Consequently, all three models are correct in this case, and they lead to intervals with coverage probability close to the specified value. However, assuming a more general model such as VVV results in wider intervals because of the larger standard error of the estimates. When the true model is EEE (Table \ref{tab:res_bs2}), assuming the incorrect model EII has a devastating effect in terms of coverage probabilities, which are then much smaller than the specified level. The same phenomenon is observed in Table \ref{tab:res_bs3}, where the correct model is VVV and models EII and EEE are both wrong. The automatic model determination method works well when the true model is EII or EEE (Tables \ref{tab:res_bs1} and \ref{tab:res_bs2}), but it does not work so well when the true model is VVV (Table \ref{tab:res_bs3}), because it sometimes select a simpler model than the true one.

From these experiments, we can conclude that the belief-plausibility intervals have coverage probabilities close to their nominal confidence levels when a correct model is assumed. Correct assumptions about parameter constraints (such as homoscedasticity) make it possible  to obtain shorter intervals when the assumptions are correct, but their can have a negative effect on coverage probabilities when the assumptions are wrong. Automatic model selection based, e.g., on the BIC criterion can be used, but the selection should be biased in favor of more complex models to avoid model misspecification.

\begin{sidewaystable}
\caption{Coverage rates and lengths of bootstrap confidence intervals (CI) and belief-plausibility intervals for 100 datasets generated from Mixture 1 (model EII), at nominal 90\% and 95\% confidence levels. The numbers in parentheses are the standard deviations over the 100 datasets. The coverage rates for correctly specified models are printed in bold.  \label{tab:res_bs1}}
\begin{center}
\begin{tabular}{ccccccccccc}
\cline{4-11}
 & &  & \multicolumn{8}{c}{Assumed model}\\
 \cline{4-11}
 & & & \multicolumn{2}{c}{EII} & \multicolumn{2}{c}{EEE} & \multicolumn{2}{c}{VVV} & \multicolumn{2}{c}{Auto}\\
 \hline
  True                              &  \multirow{2}{*}{$1-\alpha$}   & & \multirow{2}{*}{CI} & \multirow{2}{*}{[Bel,Pl]} & \multirow{2}{*}{CI} & \multirow{2}{*}{[Bel,Pl]} &\multirow{2}{*}{CI} & \multirow{2}{*}{[Bel,Pl]} &\multirow{2}{*}{CI} & \multirow{2}{*}{[Bel,Pl]}\\
  model & & & & & & & & && \\
\hline                    
 \multirow{8}{*}{EII} & \multirow{4}{*}{$0.90$} &cover. &\textbf{0.87}&\textbf{0.90} &\textbf{0.88} &\textbf{0.90} &\textbf{0.92} &\textbf{0.91}  &0.88 &0.89\\
& & & (0.159)& (0.101)&(0.125) & (0.091)& (0.102)& (0.088) & (0.15) & (0.12)\\
                               &        & length &   0.11 & 0.11 & 0.14 & 0.14  &  0.32  &0.32 &0.11 &0.11 \\
                                                              &        &  &    (0.017)&  (0.017)&  (0.028) &  (0.028) &   (0.085) & (0.088)&(0.018) &(0.018)\\
                               \cline{2-11}
 & \multirow{2}{*}{$0.95$} &cov. & \textbf{0.93} & \textbf{0.94} & \textbf{0.94} &\textbf{0.94}  &\textbf{0.96} &\textbf{0.94}  & 0.93 & 0.93 \\
 & & & (0.121)&  (0.079)&  (0.087)& (0.065) & (0.067)& (0.062) &  (0.126)& (0.097)\\
                               &        &length & 0.13 & 0.13 & 0.17 &0.17  &0.39  & 0.40  &0.133 &0.132 \\
                                                              &        & &  (0.021)&  (0.021) &  (0.033) & (0.033) & (0.097) &  (0.100) & (0.022) &  (0.022)\\
 \hline
  \end{tabular}
 \end{center}
\end{sidewaystable}

\begin{sidewaystable}
\caption{Coverage rates and lengths of bootstrap confidence intervals (CI) and belief-plausibility intervals for 100 datasets generated from Mixture 2 (model EEE), at nominal 90\% and 95\% confidence levels. The numbers in parentheses are the standard deviations over the 100 datasets. The coverage rates for correctly specified models are printed in bold. \label{tab:res_bs2}}
\begin{center}
\begin{tabular}{ccccccccccc}
\cline{4-11}
  & &  & \multicolumn{8}{c}{Assumed model}\\
 \cline{4-11}
 & & & \multicolumn{2}{c}{EII} & \multicolumn{2}{c}{EEE} & \multicolumn{2}{c}{VVV} & \multicolumn{2}{c}{Auto}\\
 \hline
  True                              &  \multirow{2}{*}{$1-\alpha$}   & & \multirow{2}{*}{CI} & \multirow{2}{*}{[Bel,Pl]} & \multirow{2}{*}{CI} & \multirow{2}{*}{[Bel,Pl]} &\multirow{2}{*}{CI} & \multirow{2}{*}{[Bel,Pl]} &\multirow{2}{*}{CI} & \multirow{2}{*}{[Bel,Pl]}\\
  model & & & & & & & & && \\
\hline   
  \multirow{8}{*}{EEE} & \multirow{4}{*}{$0.90$} &cover. &0.34 &0.50 & \textbf{0.89} & \textbf{0.91}&  \textbf{0.89} & \textbf{0.89} &0.88 &0.90\\
& & & (0.033) &(0.038) &(0.122) &(0.080)& (0.125)& (0.114)&( 0.155) &(0.107)\\
                               &        & length &   0.16 &0.16 &0.15 &0.15& 0.37& 0.37 &0.16 &0.16 \\
                                                              &        &  &   (0.032) &(0.032) &(0.031)& (0.031) &(0.082) &(0.084) &(0.036) &(0.036)\\
                               \cline{2-11}
 & \multirow{2}{*}{$0.95$} &cov. & 0.40 &0.56&  \textbf{0.95} & \textbf{0.95}&  \textbf{0.95} & \textbf{0.92} &0.94& 0.94 \\
 & & & (0.035) &(0.040) &(0.085) &(0.056) &(0.088) &(0.086) &(0.112) &(0.083)\\
                               &        &length & 0.19 &0.19 &0.18 &0.19 &0.45 &0.46 &0.19 &0.19 \\
                                                              &        & &  (0.038)& (0.039)& (0.037)& (0.037) &(0.088) &(0.091) &(0.043) &(0.044)\\
\hline
  \end{tabular}
 \end{center}
\end{sidewaystable}                         
                         
\begin{sidewaystable}
\caption{Coverage rates and lengths of bootstrap confidence intervals (CI) and belief-plausibility intervals for 100 datasets generated from Mixture 3 (model VVV), at nominal 90\% and 95\% confidence levels. The numbers in parentheses are the standard deviations over the 100 datasets. The coverage rates for correctly specified models are printed in bold. \label{tab:res_bs3}}
\begin{center}
\begin{tabular}{ccccccccccc}
\cline{4-11}
  & &  & \multicolumn{8}{c}{Assumed model}\\
 \cline{4-11}
 & & & \multicolumn{2}{c}{EII} & \multicolumn{2}{c}{EEE} & \multicolumn{2}{c}{VVV} & \multicolumn{2}{c}{Auto}\\
 \hline
  True                              &  \multirow{2}{*}{$1-\alpha$}   & & \multirow{2}{*}{CI} & \multirow{2}{*}{[Bel,Pl]} & \multirow{2}{*}{CI} & \multirow{2}{*}{[Bel,Pl]} &\multirow{2}{*}{CI} & \multirow{2}{*}{[Bel,Pl]} &\multirow{2}{*}{CI} & \multirow{2}{*}{[Bel,Pl]}\\
  model & & & & & & & & && \\
\hline   
  \multirow{8}{*}{VVV} & \multirow{4}{*}{$0.90$} &cover. &0.47 &0.58 &0.57 &0.64& \textbf{0.90} &  \textbf{0.89}& 0.65 &0.70\\
& & & (0.078) &(0.077) &(0.136)& (0.139) &(0.126) &(0.110) &(0.195) &(0.162)\\
                               &        & length &   0.16 &0.16 &0.24 &0.25 &0.31 &0.32 &0.18 &0.18 \\
                                                              &        &  &  (0.039) &(0.040) &(0.083) &(0.087)& (0.080) &(0.083)& (0.037) &(0.037)\\
                               \cline{2-11}
 & \multirow{2}{*}{$0.95$} &cov. & 0.55 &0.65 &0.67 &0.73 & \textbf{0.95}&  \textbf{0.93} &0.74 &0.78\\
 & & & (0.080) &(0.079) &(0.128) &(0.124) &(0.089) &(0.077) &(0.177) &(0.147)\\
                               &        &length & 0.19 &0.20 &0.30 &0.31& 0.39& 0.40 &0.22 &0.22 \\
                                                              &        & &  (0.046) &(0.047) &(0.093) &(0.097) &(0.097) &(0.099) &(0.046) &(0.047)\\
\hline
  \end{tabular}
 \end{center}
\end{sidewaystable}    


\paragraph{Experiment with non-normal data} Given the importance of correct model specification to ensure the frequency-calibration of belief-plausibility intervals, we can expect poor results when fitting a GMM to data generated by a mixture  whose components  are significantly non-normal. As a case study, we considered data from a mixture of three two-dimensional skew $t$ distributions \cite{wang09} with the following parameters:
\[
\bmu_1:=(3,-4)^T, \quad \bmu_2:=(3.5,4)^T, \quad \bmu_3:=(2, 2)^T,
\]
\[
\bSigma_1:=\begin{pmatrix} 1 & -0.1\\ -0.1 & 1 \end{pmatrix}, 
\bSigma_2=\bSigma_3:=\begin{pmatrix} 1 & 0\\ 0 & 1 \end{pmatrix},
\]
\[
 \pi_1=\pi_2:=0.4, \quad \pi_3:=0.4.
\]
\[
\nu_1:=3, \quad \nu_2=\nu_3:=5
\]
\[
\bdelta_1:=(3,3)^T, \quad \bdelta_2:=(1,5)^T, \quad \bdelta_3:=(-3,1)^T,
\]
where $\nu_k$ and $\bdelta_k$ denote, respectively, the degrees of freedom and the skewness parameters.

\begin{figure}
\centering  
\subfloat[\label{fig:data_t_GMM}]{\includegraphics[width=0.5\textwidth]{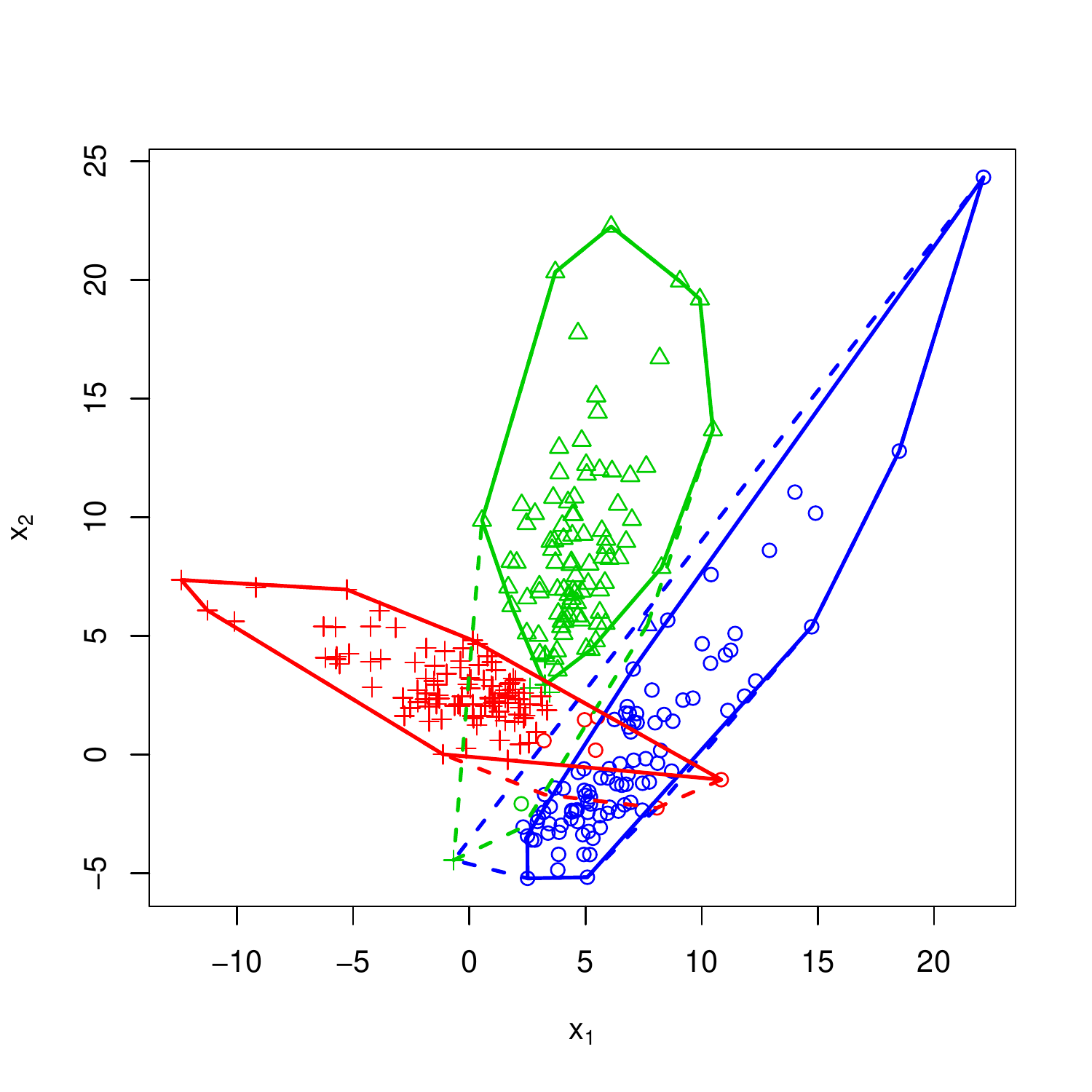}}
\subfloat[\label{fig:data_t_skewt}]{\includegraphics[width=0.5\textwidth]{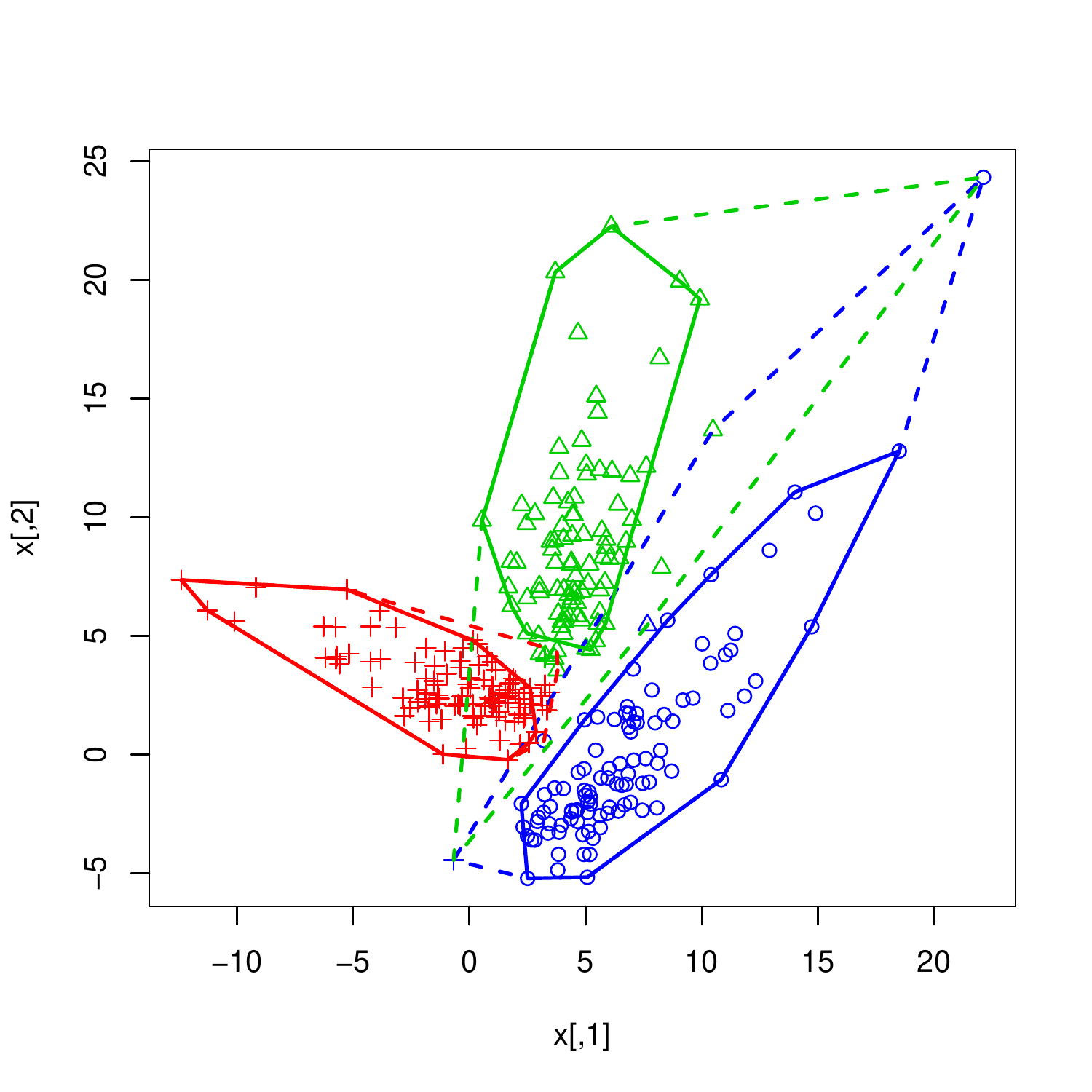}}
\caption{Evidential partitions of a dataset drawn from a mixture of skew $t$ distributions, fitted with a GMM (a) and with a mixture of skew $t$ distributions (b).  The true groups are represented by different symbols, and the maximum-plausibility  groups are represented by different colors. The solid and broken lines represent, respectively, the convex hulls of the lower and upper approximation of each cluster. (This figure is better viewed in color). \label{fig:data_t}}
\end{figure}

\begin{figure}
\centering  
\subfloat[]{\includegraphics[width=0.45\textwidth]{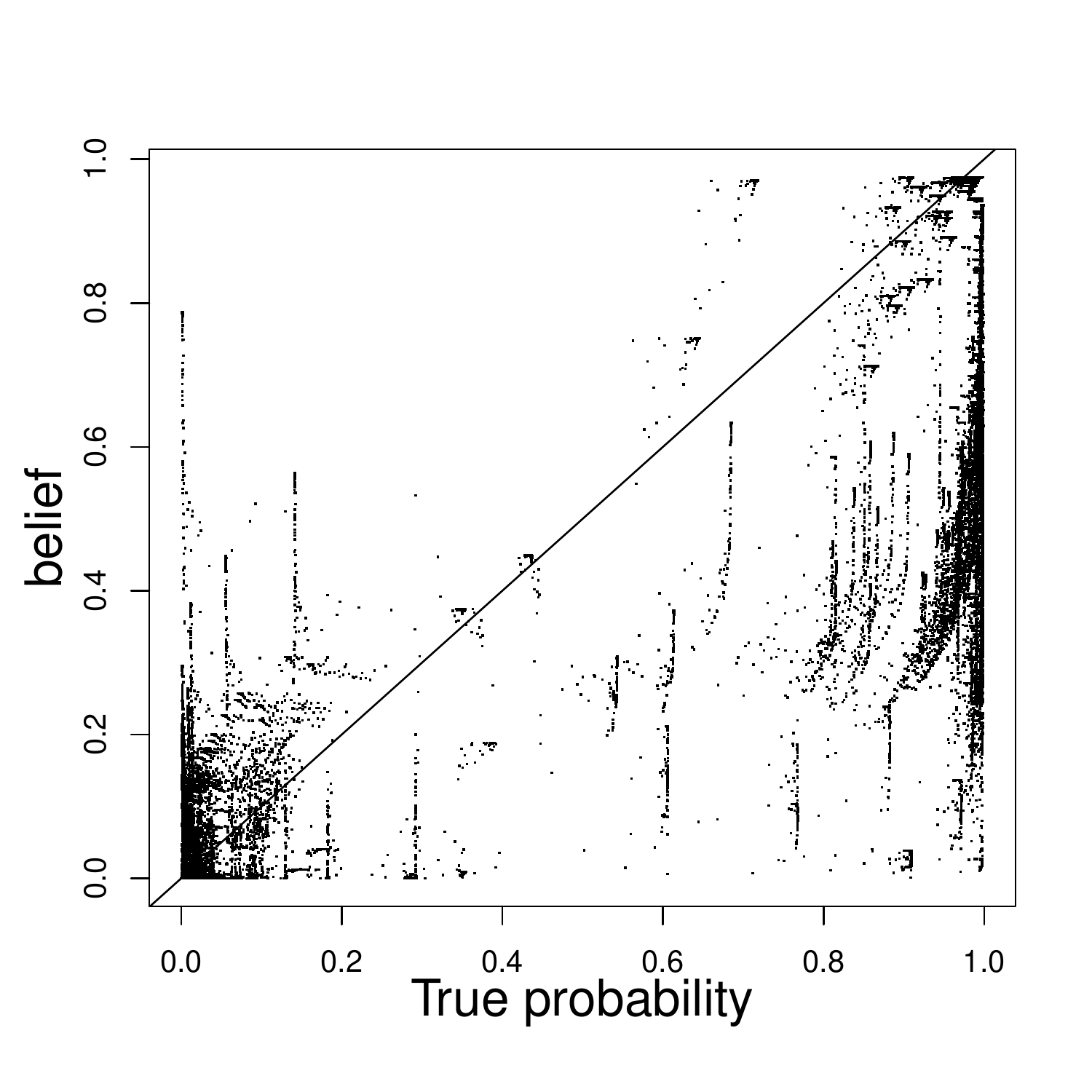}}
\subfloat[]{\includegraphics[width=0.45\textwidth]{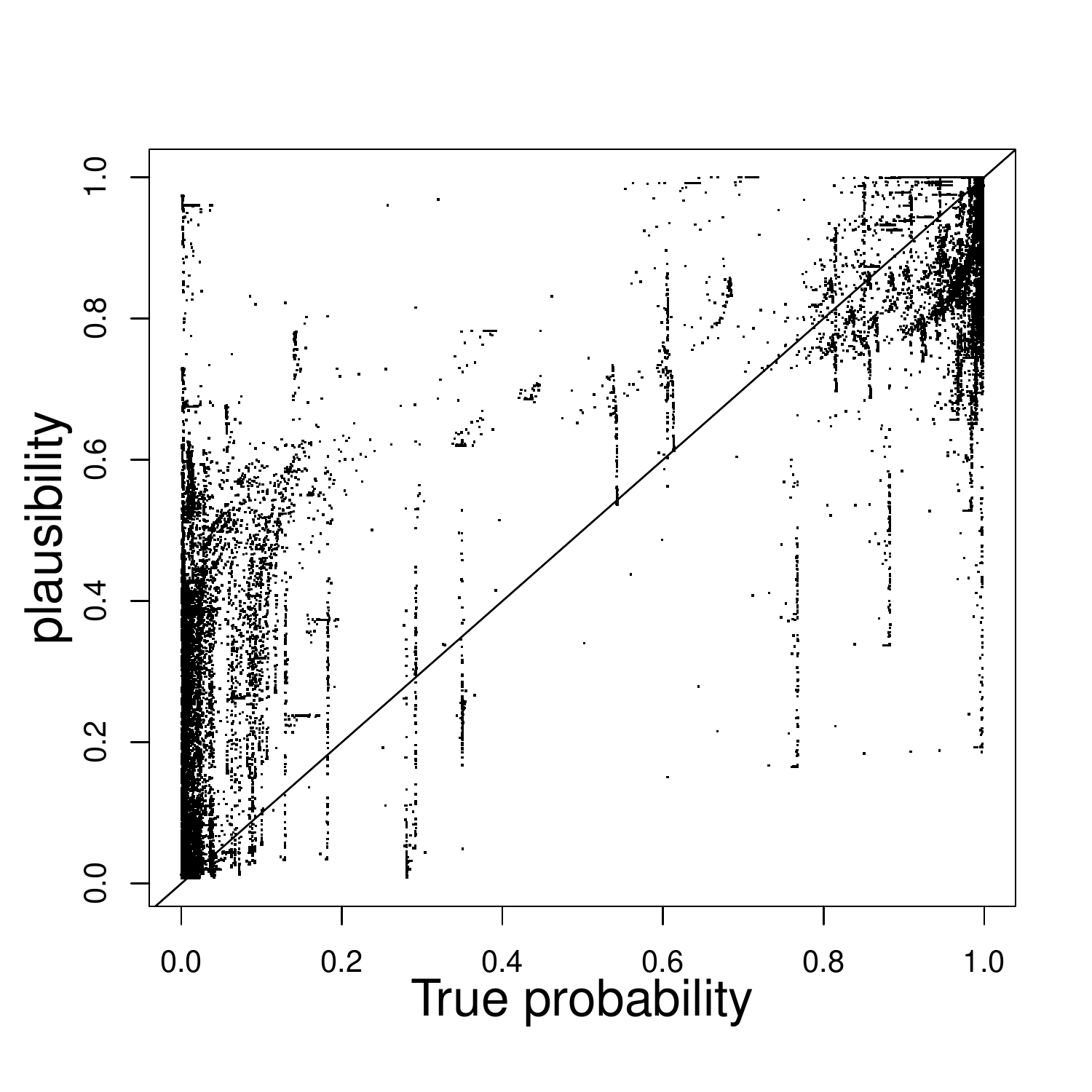}}
\caption{Dataset drawn from a mixture of skew $t$ distributions fitted with a GMM: (a) True probability $P_{ij}(\btheta)$ ($x$-axis) vs.  belief degree $Bel_{ij}(\{s_{ij}\})$ ($y$-axis); (b) True probability $P_{ij}(\btheta)$ ($x$-axis) vs.  plausibility degree $Pl_{ij}(\{s_{ij}\})$ ($y$-axis).
\label{fig:confint_t}}
\end{figure}

Figure \ref{fig:data_t_GMM} shows a  dataset of $n=300$ observations drawn from this distribution, together with the partition obtained by fitting a GMM with the assumption of equal volume of the three clusters (model EVV in package {\tt mclust}), as well as the lower and upper approximations of each cluster. We can see that the partition obtained with the normality assumption is close to the true partition (with only 12 misclassified points out of 300). However, only 50.4\% of the belief-plausibility intervals computed from 90\% bootstrap confidence intervals contain the true probabilities, which suggests that their true coverage probability is significantly smaller than the nominal one (see Figure \ref{fig:confint_t}).

\begin{figure}
\centering  
\subfloat[]{\includegraphics[width=0.45\textwidth]{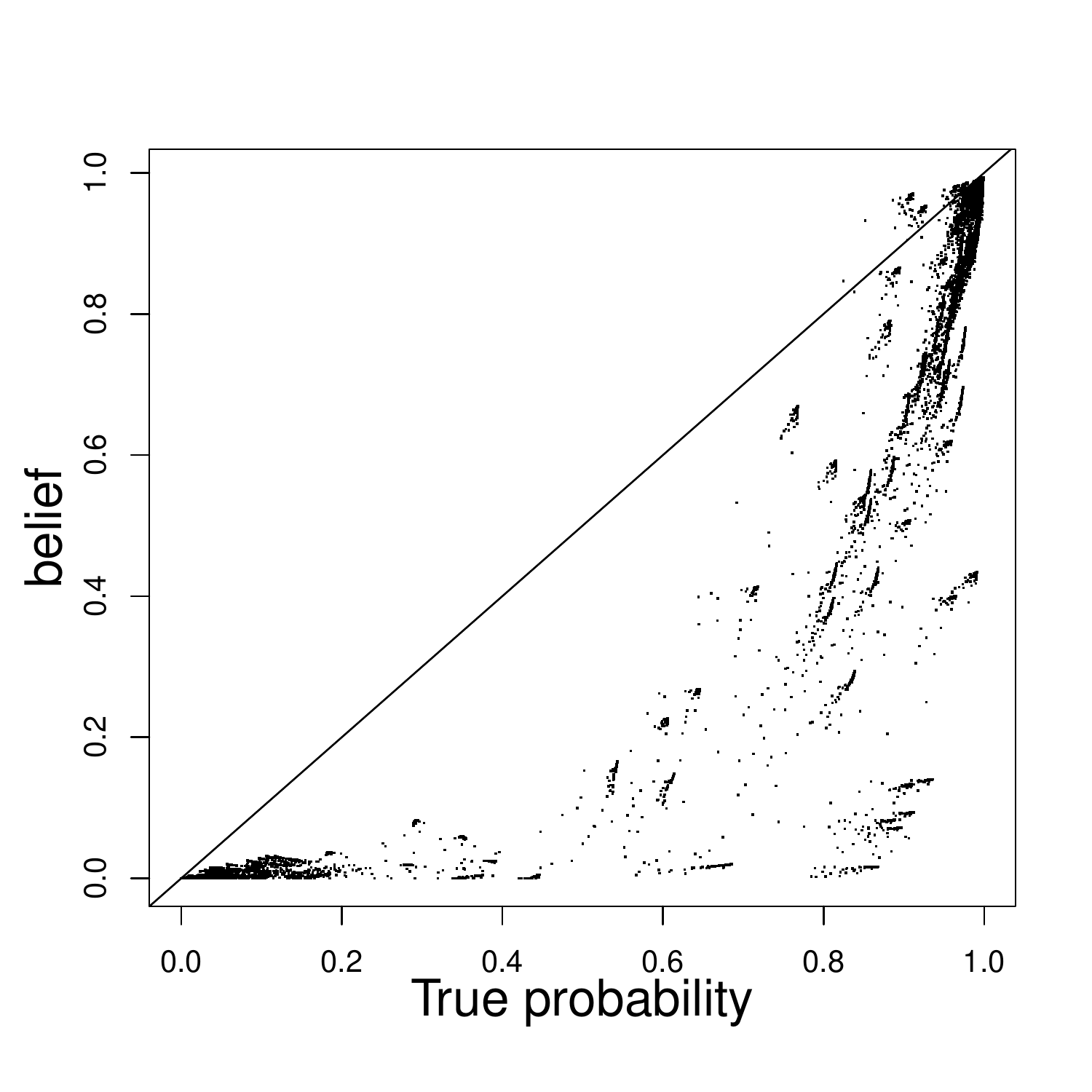}}
\subfloat[]{\includegraphics[width=0.45\textwidth]{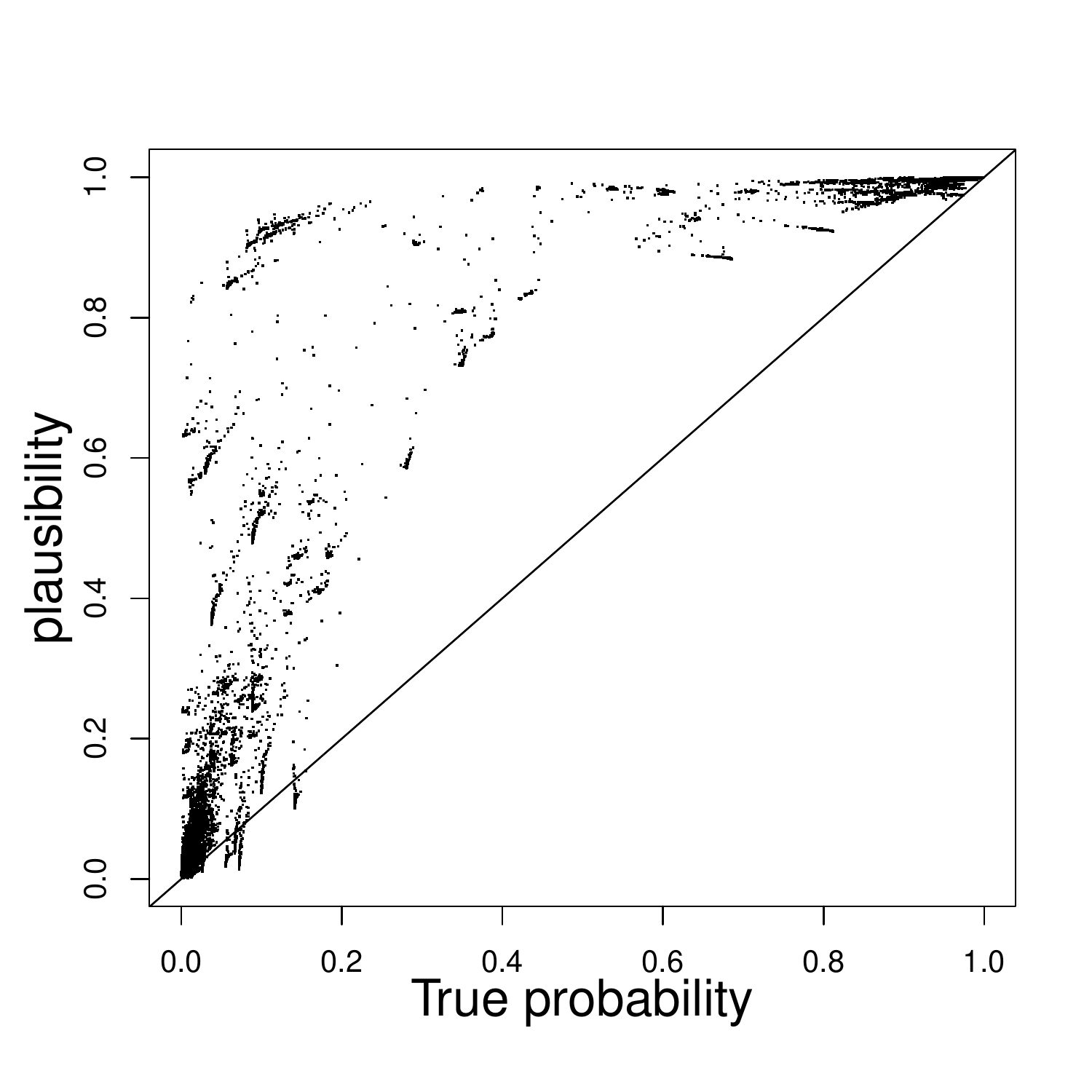}}
\caption{Dataset drawn from a mixture of skew $t$ distributions, fitted with the correct model: (a) True probability $P_{ij}(\btheta)$ ($x$-axis) vs.  belief degree $Bel_{ij}(\{s_{ij}\})$ ($y$-axis); (b) True probability $P_{ij}(\btheta)$ ($x$-axis) vs.  plausibility degree $Pl_{ij}(\{s_{ij}\})$ ($y$-axis).
\label{fig:confint_t_skew}}
\end{figure}

As noted by McLachlan and Basford \cite[Section 2.7]{mclachlan88}, ``In the situation where the sample is completely unclassified, as in the usual cluster analysis setting where there is no genuine group structure, it is a difficult task to assess the fit of a mixture model''. For assessing the fit of a GMM, a method that is not fully rigorous but that works reasonable well in practice is to fit a GMM first, and then to test the normality of the data in each cluster. Here, normality is rejected for all three components with high significance by, for instance, Henze-Zirkler's test of multivariate normality \cite{henze90}, with p-values equal to $3.2 \times 10^{-5}$, $4.0 \times 10^{-7}$ and $3.9 \times 10^{-5}$. Figure \ref{fig:data_t_skewt} displays the obtained partition as well as the lower and upper approximations of each cluster obtained by fitting a mixture of skew t distributions to the data (using the R package {\tt EMMIXskew} \cite{wang18}). As shown by Figure \ref{fig:confint_t_skew}, 93.2\% of the belief-plausibility intervals now contain the true probabilities, which is close to the nominal value of 90\%.

These results suggest that the  belief-plausibility intervals computed by our method may not be well calibrated when there is a severe lack of fit of the mixture model to the data, even though the obtained credal partition can still reveal the clustering structure of the data. In most cases, however, the data distribution can be reasonably well approximated by a GMM. This model will be assumed for the analysis of real datasets carried out in the next section. 

\subsection{Real data}
\label{subsec:real}

In this section, we apply our approach with GMMs to three real datasets, and we compare it to two evidential clustering algorithms: ECM \cite{masson08} and EVCLUS \cite{denoeux04b,denoeux16a}, both implemented in the R package {\tt evclust} \cite{denoeux20}.

\subsubsection*{\textsf{Iris} data}

\begin{figure}
\centering  
\includegraphics[width=0.8\textwidth]{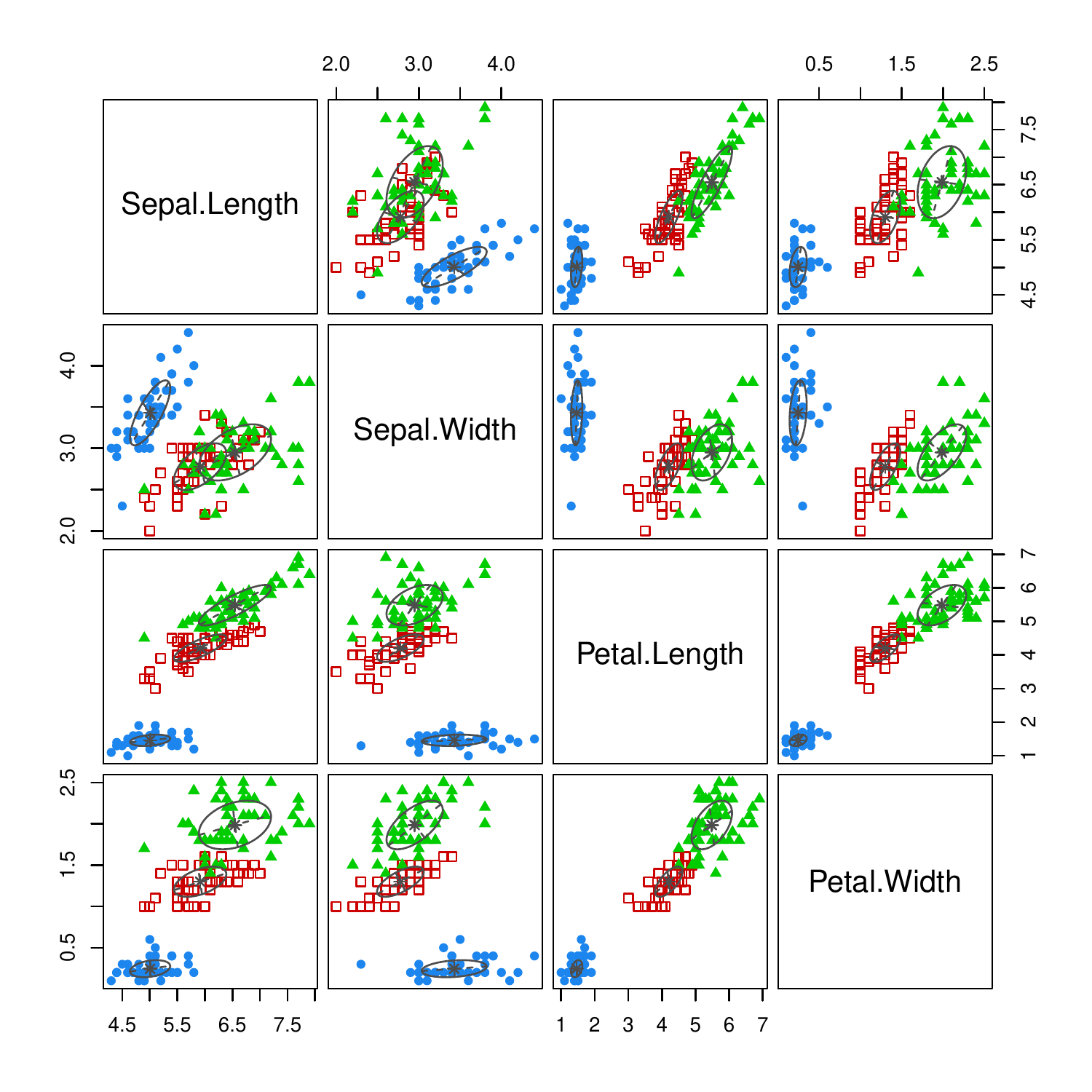}
\caption{\textsf{Iris} data with the partition obtained by fitting a GMM with $c=3$ components.  Covariances in each group are represented by isodensity ellipses. (This figure is better viewed in color). \label{fig:iris_clusters}}
\end{figure}

We first consider the well-known \textsf{Iris} dataset\footnote{Available in the R package {\tt datasets}.}, composed of 150 four-dimensional vectors partitioned in three groups corresponding to three species of Iris flowers (setosa, versicolor and virginica, abbreviated as \textsf{se}, \textsf{ve} and \textsf{vi}). For this dataset we fixed the number of clusters to $c=3$, and we searched for the best GMM model using function {\tt Mclust} in the {\tt mclust} package. The selected model was ``VEV'' corresponding to ellipsoidal clusters with equal shape. The result is represented graphically in Figure \ref{fig:iris_clusters}, showing the obtained partition as well as the cluster centers and cluster shapes represented by isodensity ellipses. The adjusted rand index (ARI) for the obtained partition is 0.90, with five objects from the \emph{versicolor} group incorrectly assigned to the \emph{virginica} group.

\begin{figure}
\centering  
\subfloat[\label{fig:iris_approx_lower}]{\includegraphics[width=0.45\textwidth]{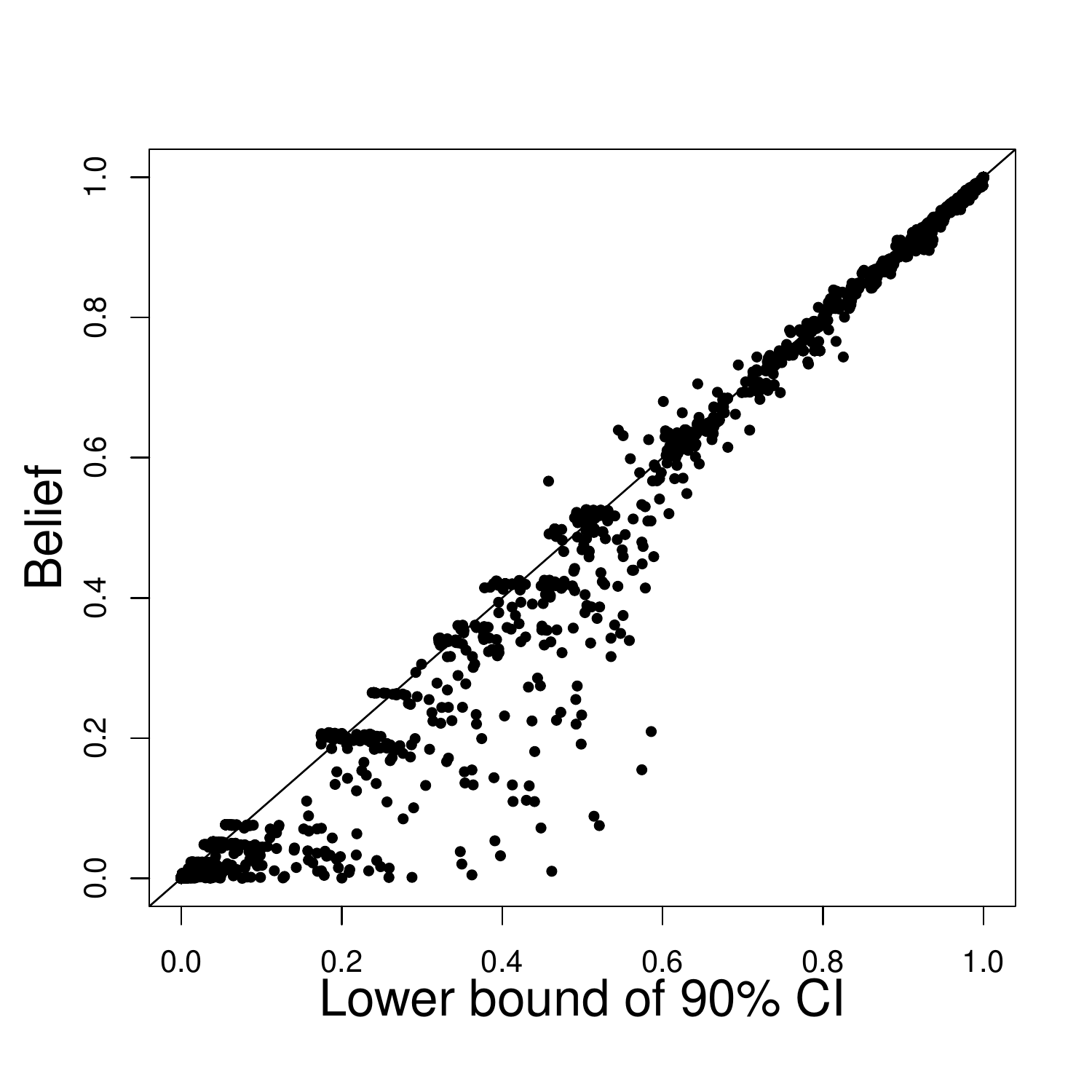}}
\subfloat[\label{fig:iris_approx_lower}]{\includegraphics[width=0.45\textwidth]{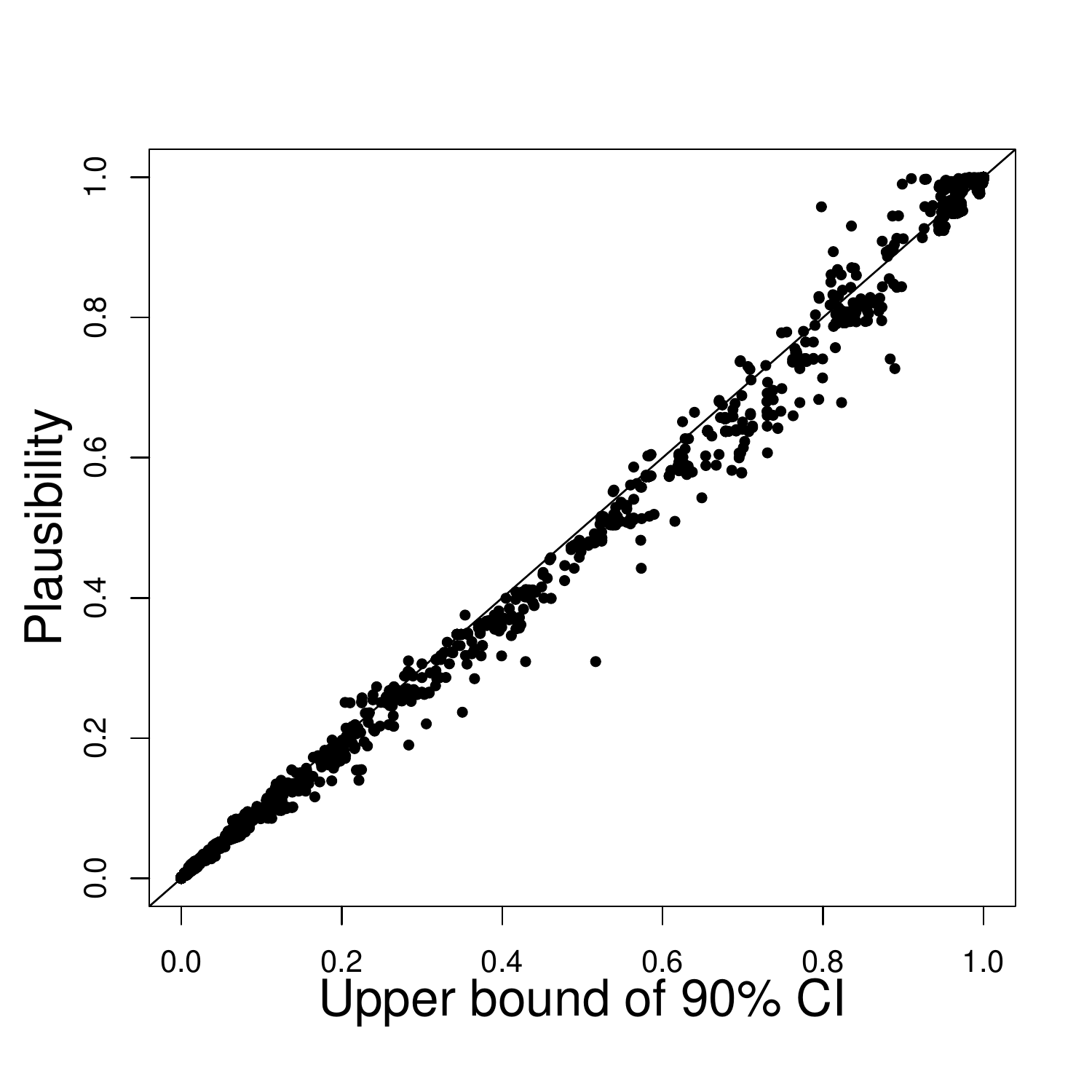}}
\caption{Approximation of confidence intervals by belief-plausibility intervals for the \textsf{Iris} data. (a) Lower bound $P_{ij}^l$ of the 90\% confidence interval on $P_{ij}(\btheta)$ ($x$-axis) vs. belief degree $Bel_{ij}(\{s_{ij}\})$ ($y$-axis); (b) Upper bound $P_{ij}^u$ of the 90\% confidence interval on $P_{ij}(\btheta)$ ($x$-axis) vs. plausibility degree $Pl_{ij}(\{s_{ij}\})$ ($y$-axis).
\label{fig:iris_approx}}
\end{figure}

We then computed 90\% bootstrap percentile confidence intervals using Algorithm \ref{alg:CI} with $B=1000$, and we constructed an evidential partition using Algorithm \ref{alg:IRQP}, with $f=6$ focal sets (the singletons and the pairs). As shown by Figure \ref{fig:iris_approx}, the confidence bounds are quite well approximated by the belief-plausibility intervals. Some belief values are smaller than the lower bounds of the confidence intervals (Figure \ref{fig:iris_approx_lower}), which suggests that the coverage probability of these intervals might be larger than the 90\% specified level.

\begin{figure}
\centering  
\includegraphics[width=0.8\textwidth]{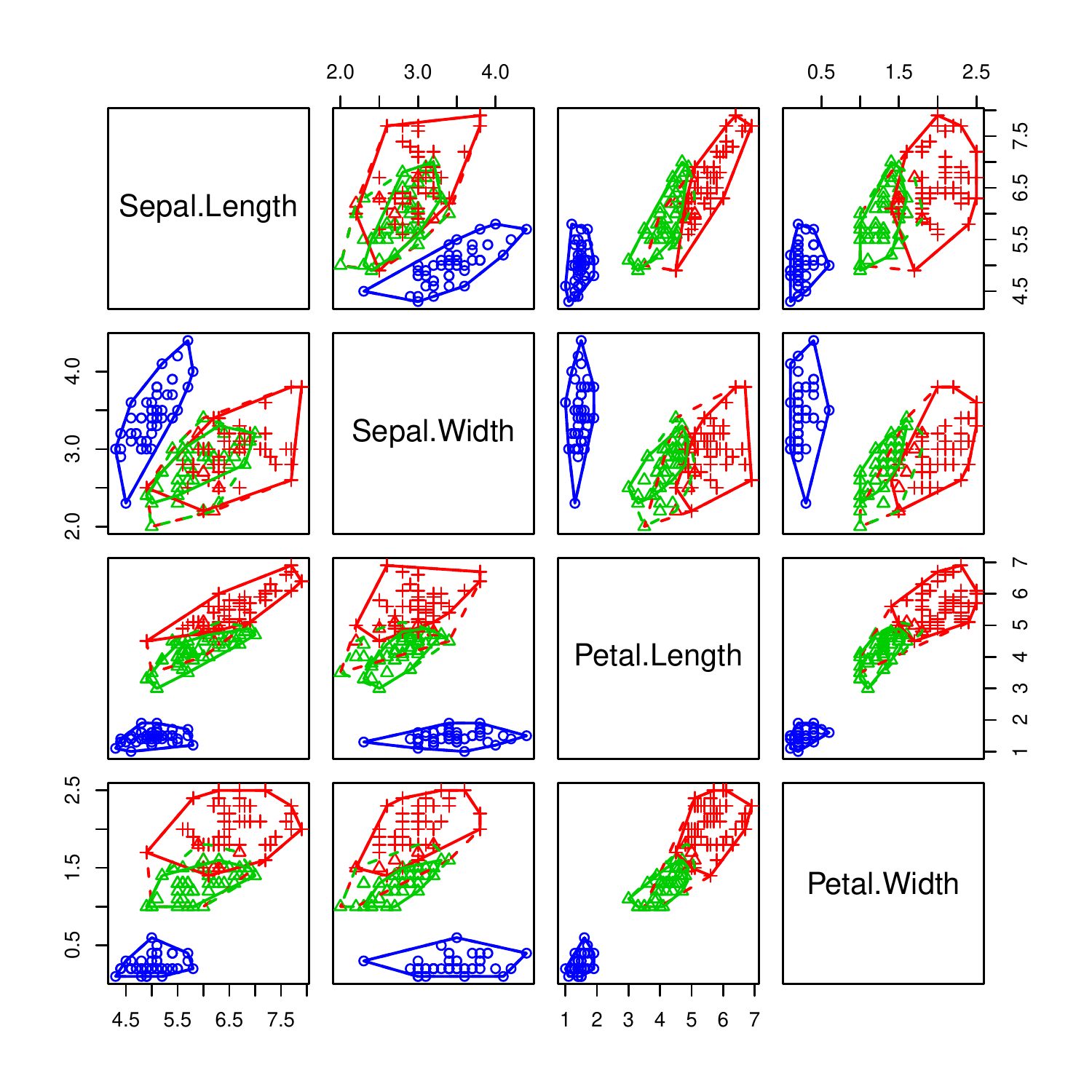}
\caption{Evidential partition of the \textsf{Iris} data using the model-based approach.  The true groups are represented by different symbols (o: setosa; triangle: versicolor; +: virginica), and the maximum-plausibility  groups are represented by different colors. The solid and broken lines represent, respectively, the convex hulls of the lower and upper approximation of each cluster. (This figure is better viewed in color). \label{fig:iris_credpart}}
\end{figure}

The lower and upper cluster approximations for the obtained evidential partition are represented in Figure  \ref{fig:iris_credpart}. We can see that the \emph{setosa} group, which is well separated from the other two, has a precise representation (for that cluster, the lower and upper approximations are equal). In contrast, the other two groups are overlapping, resulting in some objects being assigned to more than one group. Table \ref{tab:mis_iris} shows the mass functions for the five objects from the \emph{versicolor} group wrongly clustered with the \emph{virginica} in the model-based clustering. We can see that four of them (objects 69, 71, 73 and 78) have a large mass on the set $\{\textsf{ve},\textsf{vi}\}$ corresponding to the union of the \emph{versicolor} and \emph{virginica}, which indicates doubt in the assignment to any of these two clusters. Table \ref{tab:conf_iris} shows the confusion matrix, after each object has been assigned to the cluster subset with the highest mass. Clusters were labeled according to the majority group of objects they contained. We can see that 11 objects from the \emph{versicolor} group, and three from the \emph{virginica}, are assigned to the set $\{\textsf{ve},\textsf{vi}\}$. Only objet (\# 84) from the \emph{versicolor} group is misclassified as \emph{virginica}.

\begin{table}
\centering
\caption{Mass functions for the five misclassified instances in the \textsf{Iris} dataset.  The three clusters have been renamed as \textsf{se}, \textsf{ve} and \textsf{vi}. \label{tab:mis_iris} }
\begin{tabular}{ccccccc}
\hline
Object & $m(\{\textsf{se}\})$  & $m(\{\textsf{ve}\})$ & $m(\{\textsf{vi}\})$ & $m(\{\textsf{se},\textsf{ve}\})$ & $m(\{\textsf{se},\textsf{vi}\})$ & $m(\{\textsf{ve},\textsf{vi}\})$ \\
\hline
69 &0.012 &0 & 0.007 &0 &0 &0.991\\
71 &0 &  0.005 &  0.077  & 0 &  0 &  0.918 \\
73 & 0 &  0.003 &  0.202 &  0 & 0 & 0.795\\
78 &0 &  0.051&  0.052 &  0 &  0  &0.897 \\
84 & 0 & 0 & 0.882 &  0 &  0  & 0.117\\
\hline
\end{tabular}
\end{table}

\begin{table}
\centering
\caption{Confusion matrix for the \textsf{Iris} dataset.  \label{tab:conf_iris}}
\begin{tabular}{lcccc}
  &\multicolumn{4}{c}{Clustering}\\
\cline{2-5}
                 & $\{\textsf{se}\}$ &  $\{\textsf{ve}\}$ &  $\{\textsf{vi}\}$ & $\{\textsf{ve},\textsf{vi}\}$\\
\hline
\emph{setosa}      & 50  &0  &0 & 0\\
\emph{versicolor} & 0  &38 &1 &11\\
\emph{virginica}   & 0 &0  &47  &3\\
\hline
\end{tabular}
\end{table}

\begin{figure}
\centering  
\includegraphics[width=0.8\textwidth]{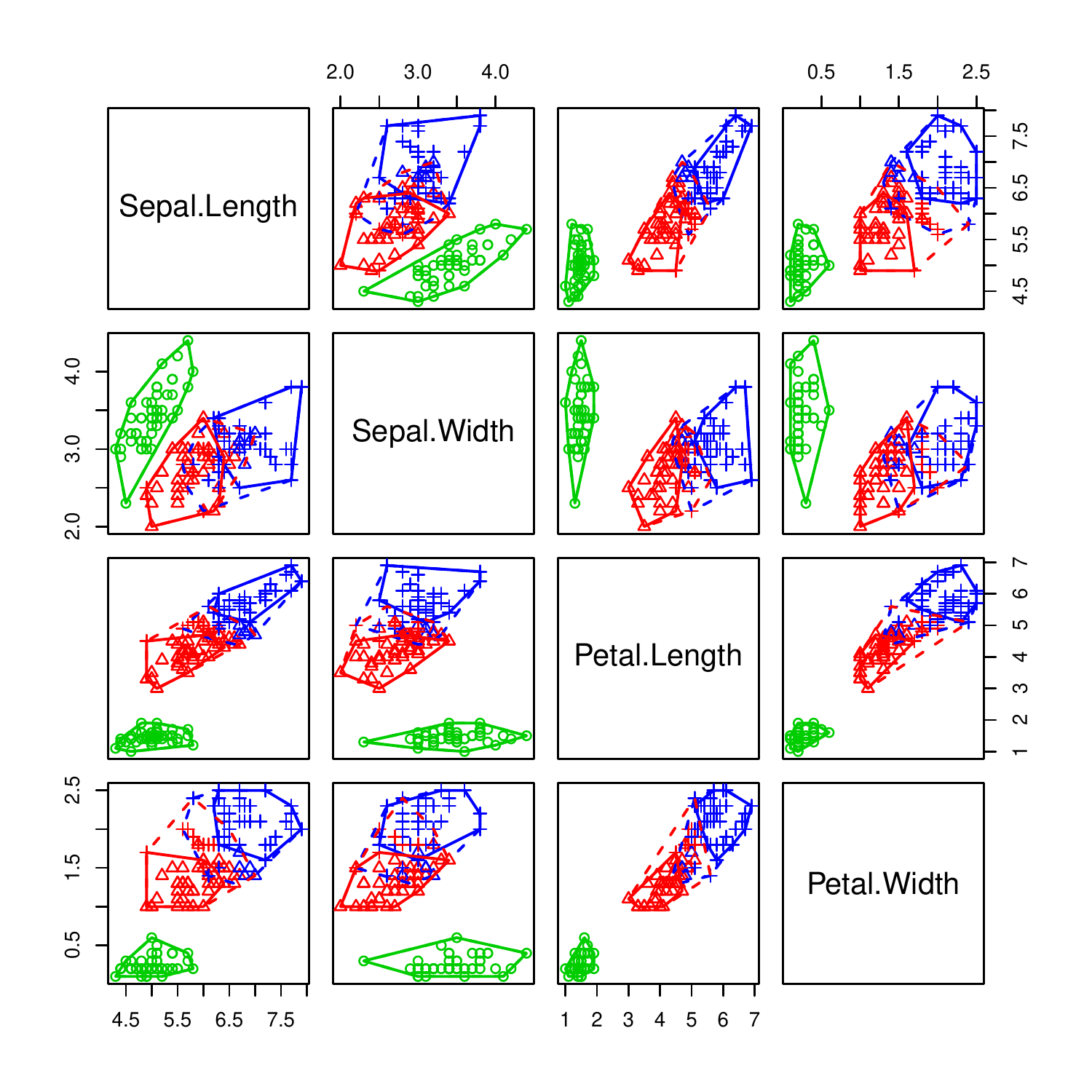}
\caption{Evidential partition obtained by ECM applied to the \textsf{Iris} data.  The true groups are represented by different symbols (o: setosa; triangle: versicolor; +: virginica), and the maximum-plausibility  groups are represented by different colors. The solid and broken lines represent, respectively, the convex hulls of the lower and upper approximation of each cluster. (This figure is better viewed in color). \label{fig:iris_credpart_ecm}}
\end{figure}

We also compared the above results to those obtained using ECM and EVCLUS. For ECM, we set the parameters $\alpha$ and $\beta$ to their default values ($\alpha = 1$ and $\beta = 2$), and we set $\delta = 100$ to avoid having any mass on the empty set. To select the focal sets, we used the method described in \cite{denoeux16a}: we first ran the algorithm using only the singletons as focal sets, and we found the pairs of classes with high similarity (see \cite{denoeux16a} for details). Here, the pair $\{\textsf{ve},\textsf{vi}\}$ was selected. Then, we ran the ECM algorithm again with focal sets $\{\textsf{se}\}$,  $\{\textsf{ve}\}$,  $\{\textsf{vi}\}$ and $\{\textsf{ve},\textsf{vi}\}$. The resulting evidential partition is shown in Figure \ref{fig:iris_credpart_ecm}, and the confusion matrix is shown in Table \ref{tab:conf_iris_ecm}. As we can see, ECM tends to extract spherical clusters, and thus fails to identify correctly the  \emph{versicolor} and \emph{virginica} groups. Comparing Tables \ref{tab:conf_iris} and \ref{tab:conf_iris_ecm}, we can see that ECM also misclassified one \emph{virginica} object  as \emph{versicolor}, but it provides a much more imprecise evidential partition, with 16 objects from the \emph{versicolor} group and 17 objects from the \emph{virginica} assigned to the compound cluster $\{\textsf{ve},\textsf{vi}\}$.
  
\begin{table}
\centering
\caption{Confusion matrix for the evidential partition obtained by ECM on the \textsf{Iris} dataset.  \label{tab:conf_iris_ecm}}
\begin{tabular}{lcccc}
  &\multicolumn{4}{c}{Clustering}\\
\cline{2-5}
                 & $\{\textsf{se}\}$ &  $\{\textsf{ve}\}$ &  $\{\textsf{vi}\}$ & $\{\textsf{ve},\textsf{vi}\}$\\
\hline
\emph{setosa}      & 50  &0  &0 & 0\\
\emph{versicolor} & 0  &34 &0 &16\\
\emph{virginica}   & 0 &1  &32  &17\\
\hline
\end{tabular}
\end{table}

\begin{figure}
\centering  
\includegraphics[width=0.8\textwidth]{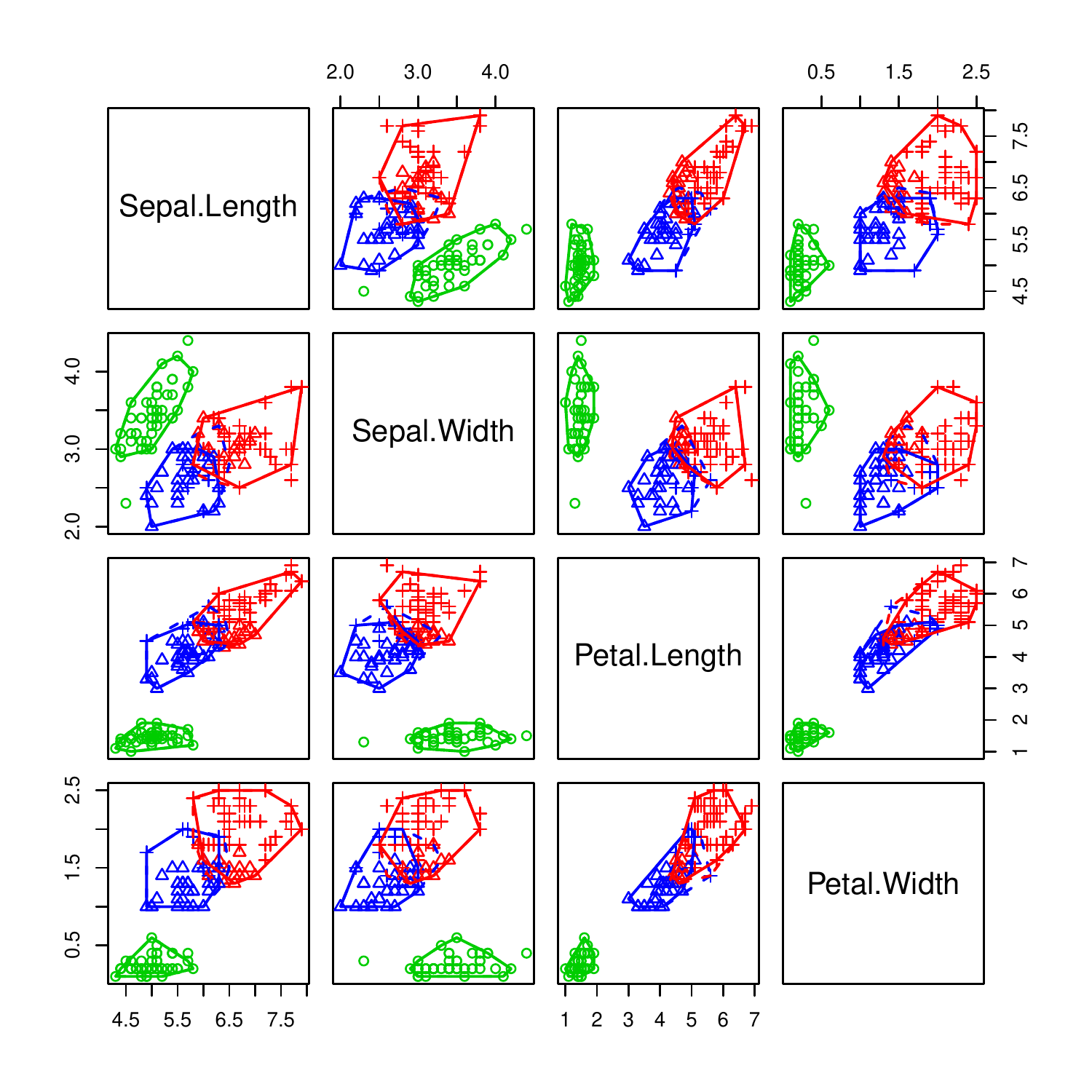}
\caption{Evidential partition obtained by $k$-EVCLUS applied to the \textsf{Iris} data.  The true groups are represented by different symbols (o: setosa; triangle: versicolor; +: virginica), and the maximum-plausibility  groups are represented by different colors. The solid and broken lines represent, respectively, the convex hulls of the lower and upper approximation of each cluster. (This figure is better viewed in color). \label{fig:iris_credpart_evclus}}
\end{figure}

Finally, we also applied the $k$-EVCLUS algorithm \cite{denoeux16a} to the same data, after normalizing the four attributes. We used the default settings and $k=50$ (see  \cite{denoeux16a} for details). We used the same procedure as with ECM to identify pairs of clusters to include as focal sets, and the whole set $\Omega=\{\textsf{se},\textsf{ve},\textsf{vi}\}$ was also included as a focal set. Again, the pair $\{\textsf{ve},\textsf{vi}\}$ was correctly identified and included as focal set. The resulting evidential partition is shown in Figure \ref{fig:iris_credpart_evclus}, and the confusion matrix is shown in Table \ref{tab:conf_iris_evclus}. EVCLUS is designed to assign some mass to the empty set, with a high mass on the empty set signaling an outlier. Here, four points were identified as outliers: they are the points outside the cluster upper approximations in Figure \ref{tab:conf_iris_evclus}. As shown by the confusion matrix in  Table \ref{tab:conf_iris_evclus}, EVCLUS does not perform very well on this dataset, with roughly the same number of correctly classified objects as ECM, but 16 misclassified objects. Overall, both ECM and EVCLUS performed significantly worse on this dataset than the model-based approach.

\begin{table}
\centering
\caption{Confusion matrix for the evidential partition obtained by $k$-EVCLUS on the \textsf{Iris} dataset.  \label{tab:conf_iris_evclus}}
\begin{tabular}{lccccc}
  &\multicolumn{5}{c}{Clustering}\\
\cline{2-6}
                 & $\emptyset$ & $\{\textsf{se}\}$ &  $\{\textsf{ve}\}$ &  $\{\textsf{vi}\}$ & $\{\textsf{ve},\textsf{vi}\}$\\
\hline
\emph{setosa}      & 2 & 48  &0  &0 & 0\\
\emph{versicolor} & 0 & 0  &32 &10 &8\\
\emph{virginica}   & 2 &0 & 6  &33  &9\\
\hline
\end{tabular}
\end{table}

\subsubsection*{\textsf{Diabetes} data}

\begin{figure}
\centering  
\includegraphics[width=0.8\textwidth]{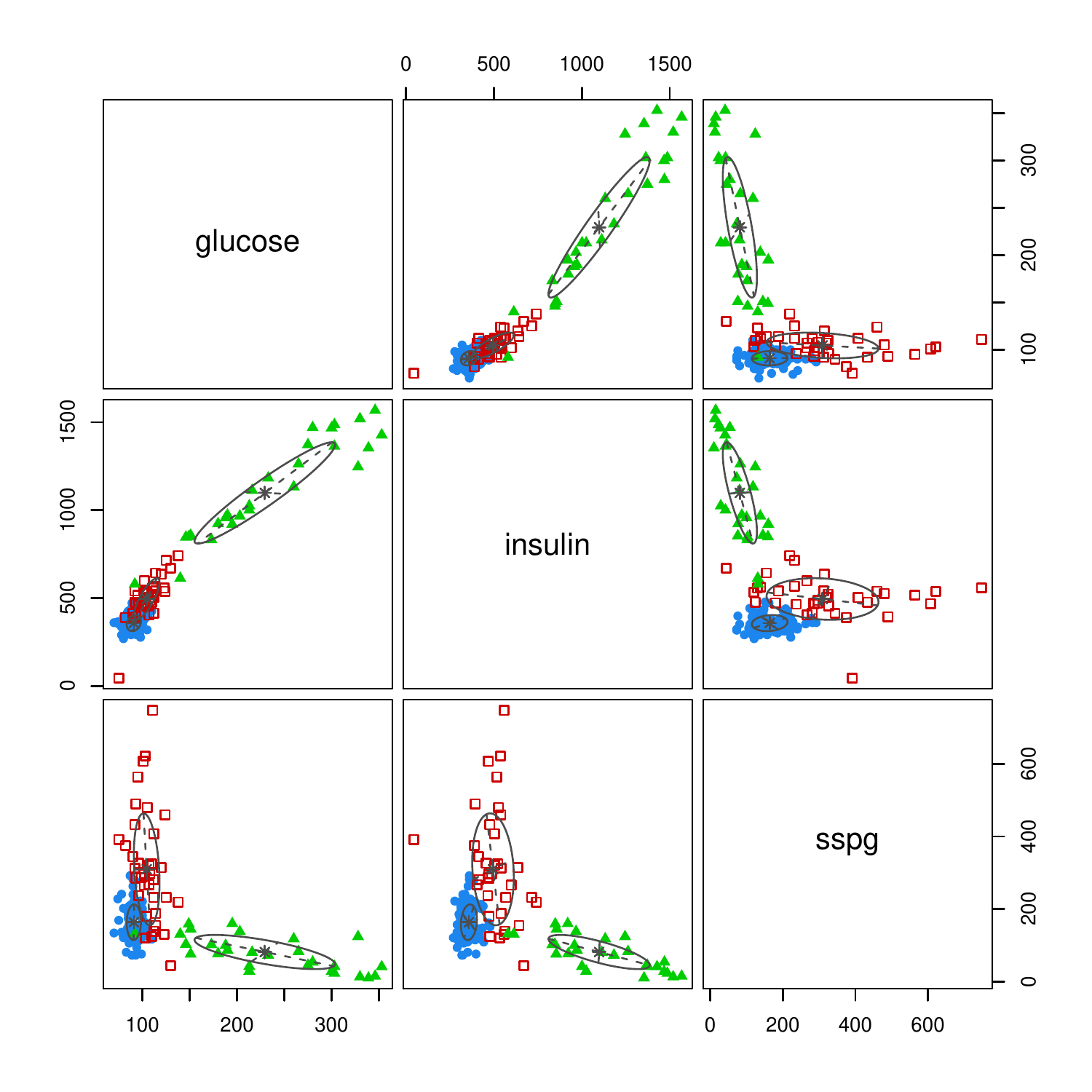}
\caption{\textsf{Diabetes} data with the partition obtained by fitting a GMM with $c=3$ components.  Covariances in each group are represented by isodensity ellipses. (This figure is better viewed in color). \label{fig:diabetes_data}}
\end{figure}

The \textsf{Diabetes} dataset\footnote{Available in the R package {\tt  mclust}.} \cite{reaven79,scrucca16} contains three measurements made on 145 non-obese adult patients classified into three groups (normal, overt, and chemical, abbreviated as \textsf{no}, \textsf{ov} and \textsf{ch}). The three attributes are \textsf{glucose} (area under plasma glucose curve after a three hour oral glucose tolerance test), \textsf{insulin} (area under plasma insulin curve after a three hour oral glucose tolerance test), and \textsf{sspg} (steady state plasma glucose). For this dataset, the best model according to BIC was found to be the full unconstrained model (``VVV'') with $c=3$ components. The data with the obtained partition as well as the estimated cluster centers and covariance ellipses are shown in  Figure \ref{fig:diabetes_data}. The ARI for the obtained partition is 0.66, and the confusion matrix is shown in Table \ref{tab:conf_diabetes_hard}. As before, clusters were labeled from the majority group among their elements. As we can see, there are 20 misclassified objects.

\begin{table}
\centering
\caption{Confusion matrix for the hard partition of the \textsf{Diabetes} dataset (model-based approach).  \label{tab:conf_diabetes_hard}}
\begin{tabular}{lccc}
  &\multicolumn{3}{c}{Clustering}\\
\cline{2-4}
                 & $\{\textsf{ch}\}$ &  $\{\textsf{no}\}$ &  $\{\textsf{ov}\}$ \\
\hline
chemical     & 26  &9  &1 \\
normal        & 4  &72 &0 \\
overt          & 6 &0  &27\\
\hline
\end{tabular}
\end{table}

\begin{figure}
\centering  
\subfloat[\label{fig:diabetes_approx_lower}]{\includegraphics[width=0.45\textwidth]{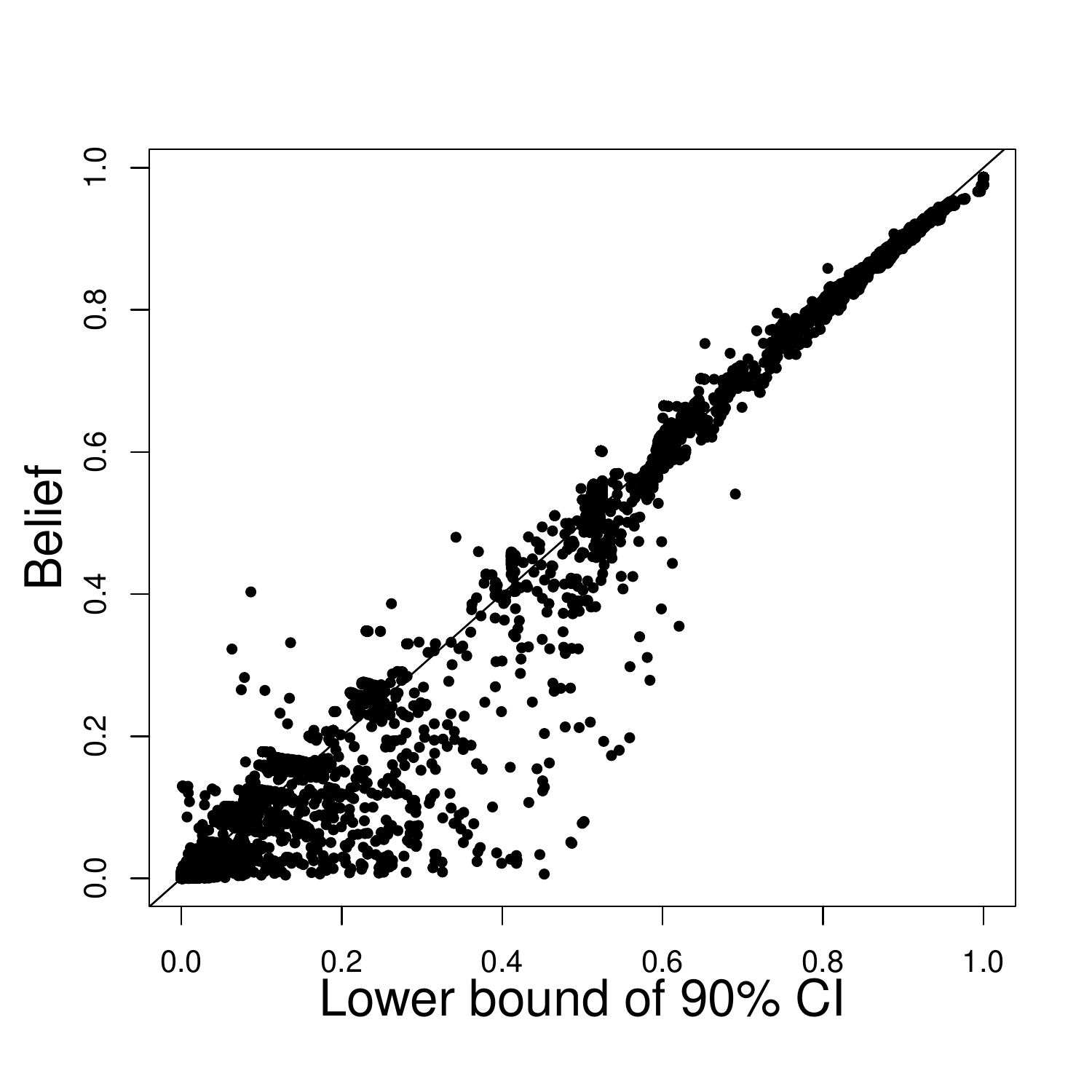}}
\subfloat[\label{fig:diabetes_approx_lower}]{\includegraphics[width=0.45\textwidth]{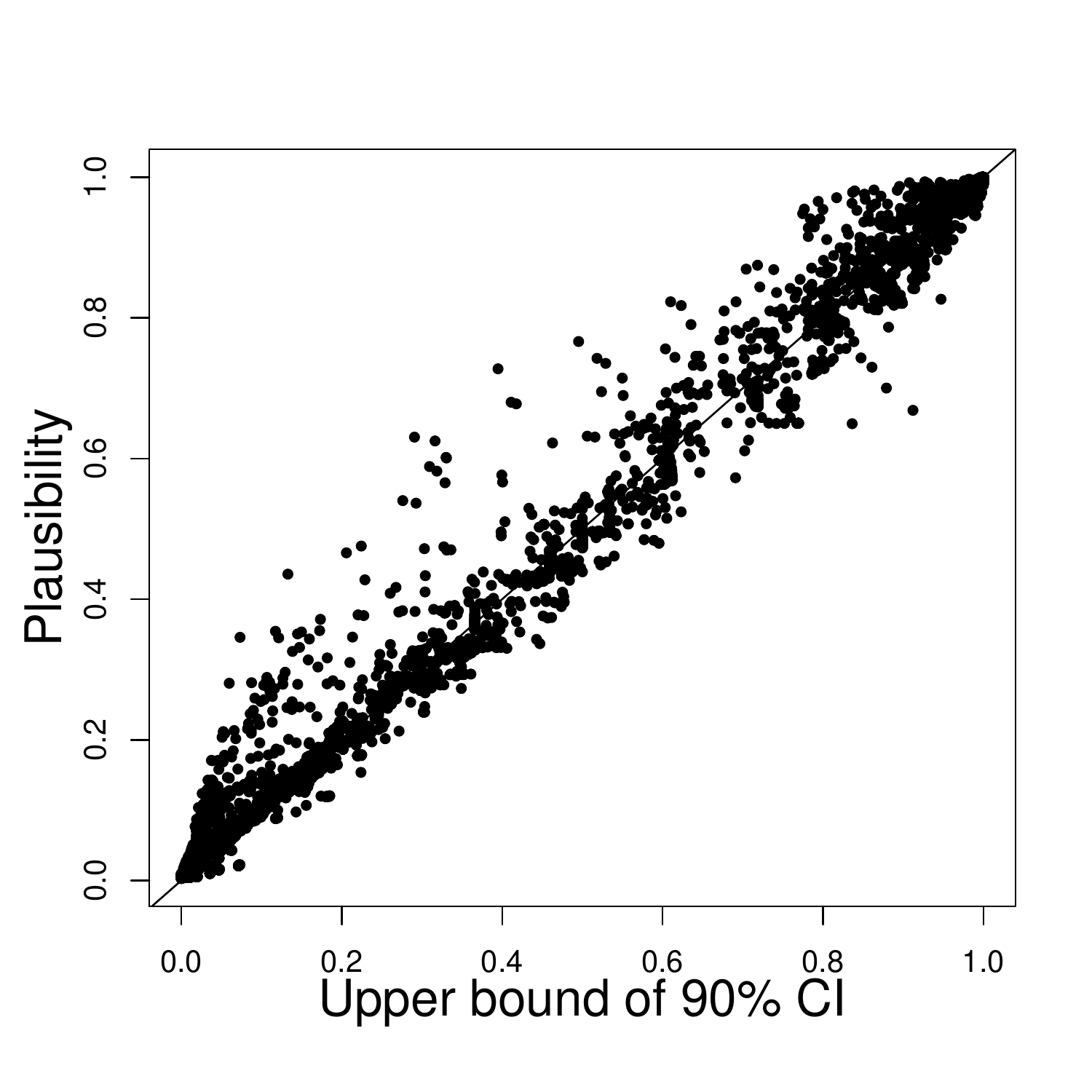}}
\caption{Approximation of confidence intervals by belief-plausibility intervals for the \textsf{Diabetes} data. (a) Lower bound $P_{ij}^l$ of the 90\% confidence interval on $P_{ij}(\btheta)$ ($x$-axis) vs. belief degree $Bel_{ij}(\{s_{ij}\})$ ($y$-axis); (b) Upper bound $P_{ij}^u$ of the 90\% confidence interval on $P_{ij}(\btheta)$ ($x$-axis) vs. plausibility degree $Pl_{ij}(\{s_{ij}\})$ ($y$-axis).
\label{fig:diabetes_approx}}
\end{figure}

\begin{figure}
\centering  
\includegraphics[width=0.8\textwidth]{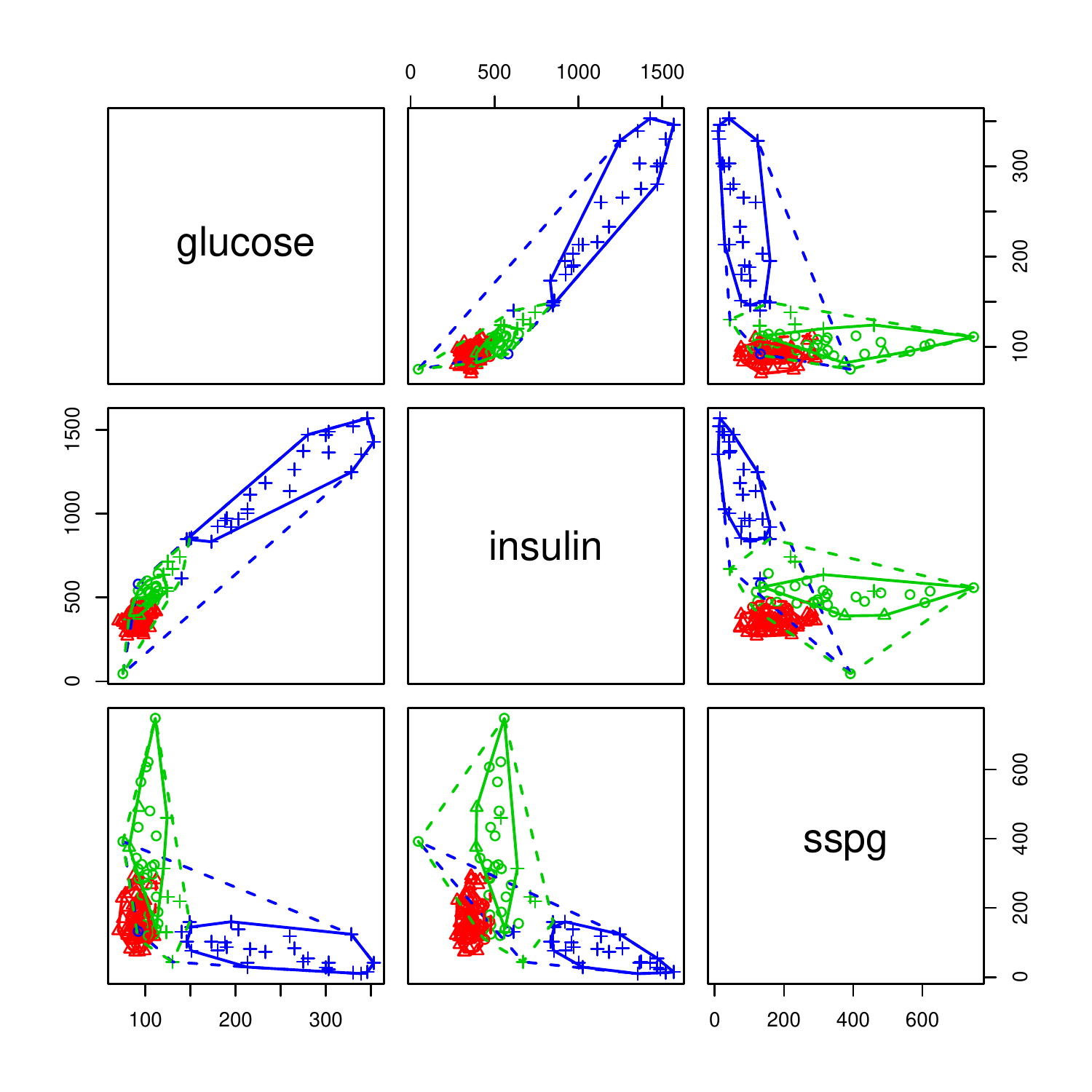}
\caption{Evidential partition of the \textsf{Diabetes} data obtained using the model-based approach.  The true groups are represented by different symbols (o: chemical; triangle: normal; +: overt), and the maximum-plausibility  groups are represented by different colors. The solid and broken lines represent, respectively, the convex hulls of the lower and upper approximation of each cluster. (This figure is better viewed in color).  \label{fig:diabetes_credpart}}
\end{figure}

As before, we computed 90\% bootstrap percentile confidence intervals with $B=1000$, and we used these intervals to constructed an evidential partition with $f=6$ focal sets (the singletons and the pairs). The resulting evidential partition is displayed in Figure \ref{fig:diabetes_credpart} and the quality of the approximation of confidence intervals by belief-plausibility intervals is illustrated in Figure \ref{fig:diabetes_approx}. The confusion matrix is shown in Table \ref{tab:conf_diabetes_model}. As we can see, the number of misclassifications is down to 14 (including 4 objects of class ``chemical'' wrongly assigned to $\{\textsf{no},\textsf{ov}\}$). As a comparison, we show the confusion matrices for ECM (Table \ref{tab:conf_diabetes_ecm}) and $k$-EVCLUS (Table \ref{tab:conf_diabetes_evclus}), which were used with the same parameter settings as for the \textsf{Iris} data. As we can see, these two methods fail to group the observations from the class ``chemical'' in a single cluster, and they perform significantly worse than the model-based approach.

\begin{table}
\centering
\caption{Confusion matrix for the evidential partition of the \textsf{Diabetes} dataset (model-based approach).  \label{tab:conf_diabetes_model}}
\begin{tabular}{lccccc}
  &\multicolumn{5}{c}{Clustering}\\
\cline{2-6}
                 & $\{\textsf{ch}\}$ &  $\{\textsf{no}\}$ &  $\{\textsf{ov}\}$ & $\{\textsf{ch},\textsf{no}\}$ & $\{\textsf{no},\textsf{ov}\}$\\
\hline
chemical     & 18  &6  &0  & 8 & 4\\
normal        & 2  &68 &0  & 6 & 0\\
overt          & 2 &0  &25   & 0 & 6\\
\hline
\end{tabular}
\end{table}

\begin{table}
\centering
\caption{Confusion matrix for the evidential partition of the \textsf{Diabetes} dataset obtained by ECM.  \label{tab:conf_diabetes_ecm}}
\begin{tabular}{lcccc}
  &\multicolumn{4}{c}{Clustering}\\
\cline{2-5}
                 & $\{\textsf{ch}\}$ &  $\{\textsf{no}\}$ &  $\{\textsf{ov}\}$ & $\{\textsf{ch},\textsf{no}\}$ \\
\hline
chemical     & 8  &17  &0  & 11 \\
normal        & 2  &68 &0  & 6 \\
overt          & 1 &10  &21   & 1 \\
\hline
\end{tabular}
\end{table}

\begin{table}
\centering
\caption{Confusion matrix for the evidential partition of the \textsf{Diabetes} dataset obtained by $k$-EVCLUS.  \label{tab:conf_diabetes_evclus}}
\begin{tabular}{lccccc}
  &\multicolumn{5}{c}{Clustering}\\
\cline{2-6}
                 & $\{\textsf{ch}\}$ &  $\{\textsf{no}\}$ &  $\{\textsf{ov}\}$ & $\{\textsf{no},\textsf{ov}\}$ & $\{\textsf{ch},\textsf{no},\textsf{ov}\}$\\
\hline
chemical     & 8  &27  &0  & 0 & 1 \\
normal        & 1  &75 &0  & 0 & 0 \\
overt          & 1 &9  &21   & 2 & 0 \\
\hline
\end{tabular}
\end{table}

\subsubsection*{\textsf{GvHD} data}

The GvHD (Graft-versus-Host Disease) data\footnote{Available in the R package {\tt  mclust}.} consist of four biomarker variables, namely, CD4, CD8b, CD3, and CD8, observed in flow cytometry data for two patients \cite{brinkman07,scrucca16}. We used the data from the GvHD positive patient, which originally contained  9083 observations. We randomly selected 1000 observations. The objective of the analysis is to identify  cell sub-populations present in the  sample. There are no ground truth labels for this dataset, but we use it to as an example of a dataset with a larger number of clusters than the two previous datasets.

\begin{figure}
\centering  
\includegraphics[width=0.6\textwidth]{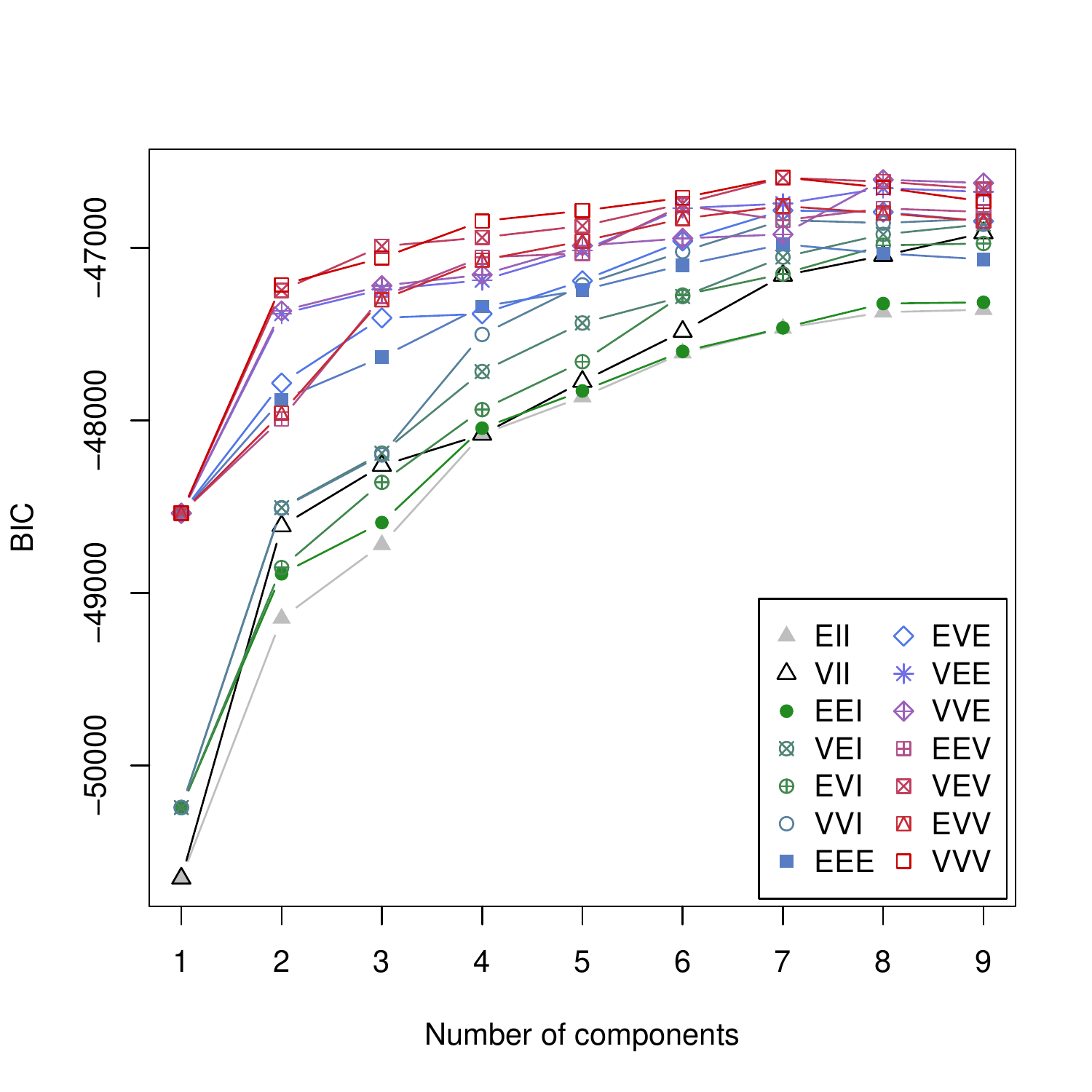}
\caption{Model selection for the \textsf{GvHD}: BIC vs. number of clusters for the 14 models defined in R package {\tt mclust}. \label{fig:GvHD_model_selection}}
\end{figure}

\begin{figure}
\centering  
\includegraphics[width=0.8\textwidth]{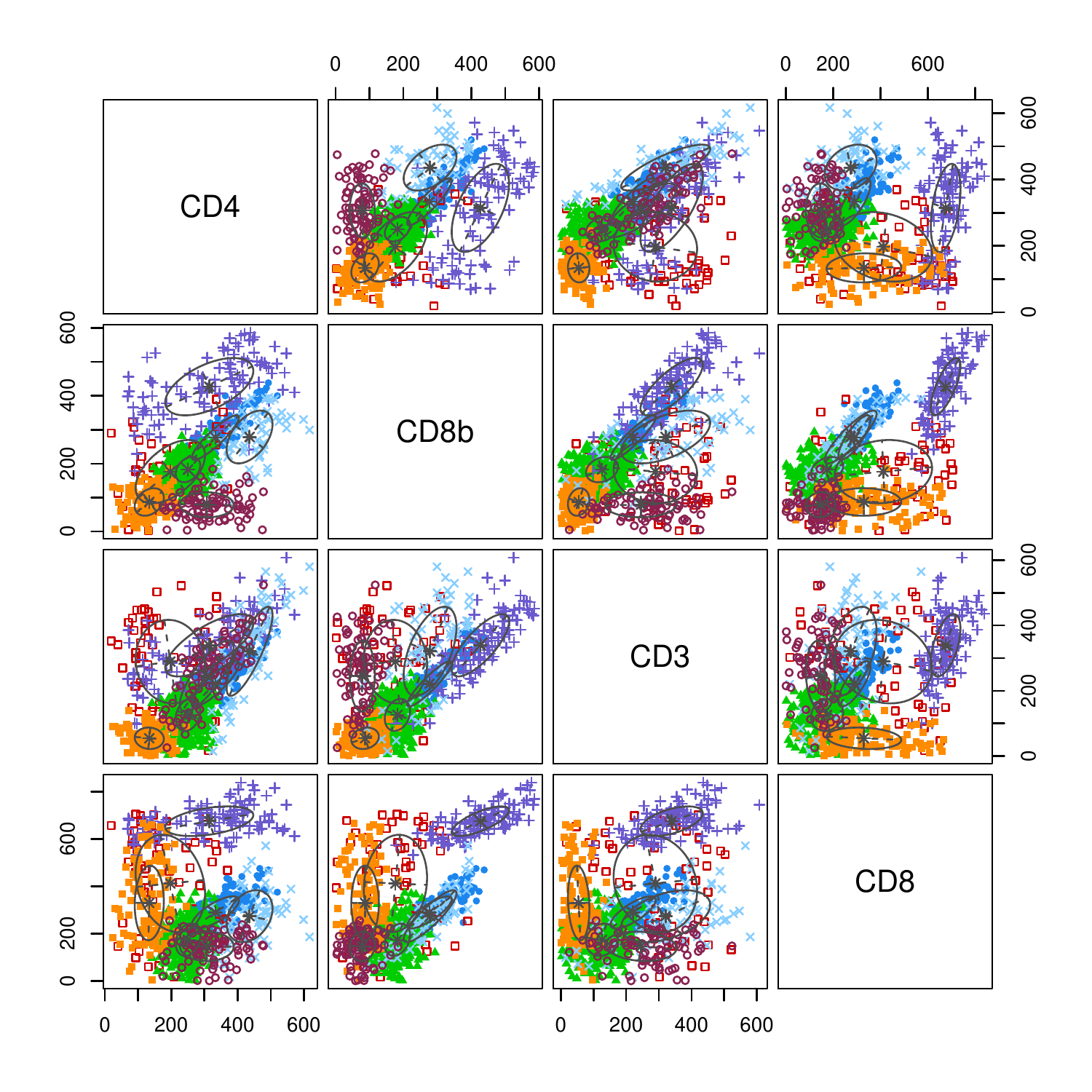}
\caption{\textsf{GvHD} data with the partition obtained by fitting a GMM with $c=7$ components.  Covariances in each group are represented by isodensity ellipses. (This figure is better viewed in color). \label{fig:GvHD_cluster}}
\end{figure}

As seen in Figure \ref{fig:GvHD_model_selection}, the best model according to BIC is the full (unconstrained) model with $c=7$ clusters. The corresponding partition as well as the cluster centers and covariance ellispses are shown in Figure \ref{fig:GvHD_cluster}.

\begin{figure}
\centering  
\includegraphics[width=0.8\textwidth]{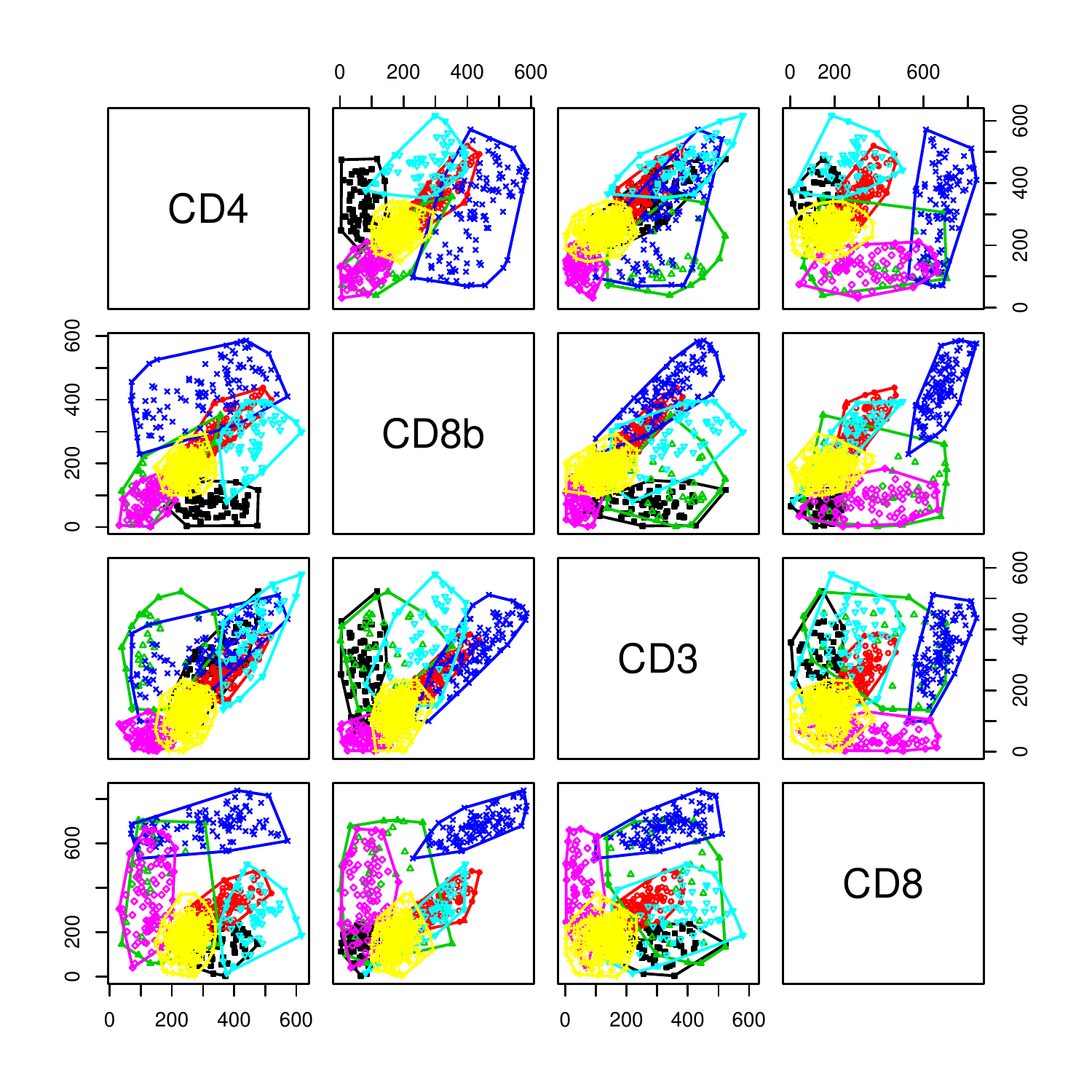}
\caption{Evidential partition of the \textsf{GvHD} data: lower approximations and convex hulls of the upper approximations. The solid and broken lines represent, respectively, the convex hulls of the lower and upper approximation of each cluster. (This figure is better viewed in color). \label{fig:GvHD_credpart}}
\end{figure}

With seven clusters, the maximum number of nonempty focal sets in the evidential partition is $2^7-1=127$. Restricting the focal sets to singletons and pairs leaves us with $7+ (6\times7)/2=28$ focal sets. However, not all pairs are needed, because some pairs of clusters actually do not overlap. To further reduce the number of focal sets, we can use a method similar to the one proposed in \cite{denoeux16a}. The similarity between two clusters $k$ and $l$ can be measured by
\[
s_{kl}=\sum_{i=1}^n \pi_k(\bx_i;\bthetah)\, \pi_l(\bx_i;\bthetah).
\]
Based on these similarities, we can identify clusters that are mutual $K$-nearest neighbors. With $K=2$, we obtained five pairs of mutual nearest neighbors: (1,3), (2,4), (1,6), (3,7) and (5,7). These five pairs and the seven singletons gave us $f=12$ focal sets. We used the same method as above to compute the bootstrap percentile confidence intervals and construct an evidential partition. The cluster lower approximations and the convex hulls of the upper approximations are shown in Figure \ref{fig:GvHD_credpart}.

Using this pair selection approach, the model can be used even with large numbers of clusters (several dozens or even several hundreds). The main limitation of the method is related to the number of objects. The necessity to compute and store the $n(n-1)/2$  belief-plausibility intervals results in a quadratic memory and time complexity, which precludes application of the method to datasets with more than a few thousand objects. However, it might be possible to use only pairwise belief-plausibility intervals for pairs of neighboring objects, as done in \cite{denoeux16a} to make the EVCLUS algorithm applicable to large datasets. This idea remains to be  investigated.

\section{Conclusions}
\label{sec:concl}

We have described a new model-based approach  to evidential clustering. The method starts by estimating the parameters of  a finite mixture model. In this paper, we used GMMs, but there is no restriction on the kinds of models that can be used. For instance, for  categorical data, latent class models would be more suitable. The model is first fitted using the EM algorithm, and bootstrap percentile confidence intervals on pairwise probabilities $P_{ij}$ at some confidence level $1-\alpha$ are computed. Here, $P_{ij}$ is the probability that objects $i$ and $j$ belong to the same cluster. Finally, an evidential partition is constructed in such a way that pairwise degrees of belief $Bel_{ij}(\{s_{ij}\})$ and plausibility $Pl_{ij}(\{s_{ij}\})$ approximate the bounds of the confidence intervals in the least squares sense. The evidential partitions constructed using this method are approximately calibrated, in the sense that the belief-plausibility intervals $[Bel_{ij}(\{s_{ij}\}),Pl_{ij}(\{s_{ij}\})]$ contain the true probabilities $P_{ij}$ with probability approximately equal to $1-\alpha$. This evidential partition provides a more complete description of the clustering structure than does  the fuzzy partition directly provided by the  EM algorithm, as  it also takes into account uncertainty in the estimation of class probabilities.

We have presented extensive experimental results showing that the coverage probabilities of the belief-plausibility intervals are  close to their nominal confidence level when the model is correctly specified. We have also demonstrated the applicability of this approach to several real datasets, and compared the evidential partitions obtained using this model-based approach to those obtained with ECM and EVCLUS, the two main evidential clustering algorithms  available so far. Model-based evidential clustering inherits the advantages of classical model-based clustering. In particular, various assumptions about cluster shapes can be formalized as assumptions about component probability distributions, and model selection criteria such as BIC make it possible to determine the number of clusters automatically.

As the method requires the construction of confidence intervals for each pair objects, it has quadratic complexity, which makes it unsuitable for the analysis of very large datasets containing more than a few thousands of objects. One remedy could be to use only the belief-plausibility intervals for pairs of neighboring objects, an idea exploited  in \cite{denoeux16a}  to apply  the EVCLUS algorithm to large datasets. This research direction will be explored in future work. 


\end{document}